\documentclass[runningheads]{llncs}

\usepackage{eccv}

\usepackage{eccvabbrv} 

\usepackage{graphicx}
\usepackage{amsmath}
\usepackage{amssymb}
\usepackage{booktabs}
\usepackage{multirow}
\usepackage{color}
\usepackage[accsupp]{axessibility}  

\usepackage{xr-hyper} 
\usepackage[pagebackref,breaklinks,colorlinks,citecolor=eccvblue]{hyperref}

\usepackage{orcidlink}

\definecolor{citecolor}{HTML}{0071BC}
\definecolor{linkcolor}{HTML}{ED1C24}

\usepackage{fancyvrb}
\usepackage{xparse}
\usepackage{xcolor}   
\usepackage{colortbl}        
\usepackage{array}
\usepackage{adjustbox}            
\usepackage{threeparttable}       
\usepackage{tabularx}
\usepackage{siunitx}
\usepackage{caption}
\usepackage{wrapfig}
\usepackage{enumitem}
\usepackage{tikz}
\usetikzlibrary{arrows.meta,positioning,calc}

\usepackage{amsmath,amsfonts,bm}
\usepackage{booktabs}
\usepackage{multirow}

\def\eqref#1{equation~\ref{#1}}

\def\1{\bm{1}}

\DeclareMathAlphabet{\mathsfit}{\encodingdefault}{\sfdefault}{m}{sl}
\SetMathAlphabet{\mathsfit}{bold}{\encodingdefault}{\sfdefault}{bx}{n}

\DefineVerbatimEnvironment{ColorVerbatim}{Verbatim}{
  baselinestretch=1,      
  commandchars=\\\{\}     
}

\definecolor{zhen}{rgb}{0.08, 0.38, 0.74}
\definecolor{Gray}{gray}{0.92}

\newcommand{\ov}{LLaVA-OneVision}

\newcommand{\ColorMin}{-10}
\newcommand{\ColorMax}{30}
\newcommand{\setcolorrange}[2]{\renewcommand{\ColorMin}{#1}\renewcommand{\ColorMax}{#2}}
\newcommand{\ColorGammaPos}{0.8}
\newcommand{\ColorGammaNeg}{0.9}
\newcommand{\ColorNegYellowGamma}{1.2}
\newcommand{\setcolorgammas}[2]{%
  \renewcommand{\ColorGammaNeg}{#1}%
  \renewcommand{\ColorGammaPos}{#2}%
}
\newcommand{\setnegyellow}[1]{\renewcommand{\ColorNegYellowGamma}{#1}}

\newcommand{\applycellcolor}[1]{%
  \pgfmathsetmacro{\val}{#1}%
  \ifdim \val pt < 0pt
    \pgfmathsetmacro{\den}{abs(\ColorMin)}%
    \ifdim \den pt = 0pt
      \def\mix{0}\def\hue{0}%
    \else
      \pgfmathsetmacro{\ratio}{min(1,max(0,(-\val)/\den))}%
      \pgfmathsetmacro{\rpow}{pow(\ratio,\ColorGammaNeg)}%
      \pgfmathsetmacro{\invpow}{pow(1-\ratio,\ColorGammaNeg)}%
      \pgfmathsetmacro{\mixraw}{100*\rpow/(\rpow+\invpow)}%
      \pgfmathtruncatemacro{\mix}{max(0,min(100,round(\mixraw)))}%
      \pgfmathsetmacro{\hueraw}{100*pow(\ratio,\ColorNegYellowGamma)}%
      \pgfmathtruncatemacro{\hue}{max(0,min(100,round(\hueraw)))}%
    \fi
    \edef\ApplyColorSpec{BurntOrange!\hue!Yellow!\mix!white}%
    \expandafter\cellcolor\expandafter{\ApplyColorSpec}%
  \else
    \pgfmathsetmacro{\den}{\ColorMax}%
    \ifdim \den pt = 0pt
      \def\mix{0}%
    \else
      \pgfmathsetmacro{\ratio}{min(1,max(0,\val/\den))}%
      \pgfmathsetmacro{\rpow}{pow(\ratio,\ColorGammaPos)}%
      \pgfmathsetmacro{\invpow}{pow(1-\ratio,\ColorGammaPos)}%
      \pgfmathsetmacro{\mixraw}{100*\rpow/(\rpow+\invpow)}%
      \pgfmathtruncatemacro{\mix}{max(0,min(100,round(\mixraw)))}%
    \fi
    \edef\ApplyColorSpec{CornflowerBlue!\mix!white}%
    \expandafter\cellcolor\expandafter{\ApplyColorSpec}%
  \fi
}

\newcommand{\fmtplain}[1]{#1}

\newcommand{\fmtul}[1]{\underline{#1}}

\newcommand{\applycolorwith}[2]{
  \applycellcolor{#1}%
  #2{\pgfmathprintnumber[fixed,precision=1,zerofill,showpos]{#1}}%
}

\newcommand{\applycolor}[1]{\applycolorwith{#1}{\fmtplain}}   
\newcommand{\applycolorUL}[1]{\applycolorwith{#1}{\fmtul}}    

\setcolorrange{-30}{35}
\setcolorgammas{0.9}{0.8}
\setnegyellow{1.2}

\definecolor{forgettingcolor}{HTML}{dd8e43}
\definecolor{learningcolor}{HTML}{519ec9}
\newcommand{\learn}[1]{\textcolor{learningcolor}{#1}}
\newcommand{\forget}[1]{\textcolor{forgettingcolor}{#1}}

\newcommand{\rebuttaledit}[1]{{#1}}

\newcommand{\citep}[1]{\cite{#1}}
\newcommand{\citet}[1]{\cite{#1}}

\providecommand{\mainlabprefix}{}
\newcommand{\mainref}[1]{\ref{\mainlabprefix#1}}

\renewcommand{\paragraph}[1]{%
  \vspace{2mm}                
  \noindent\textbf{#1}       
}

\makeatletter
\let\XR@orig@writefile\@writefile
\def\@writefile#1#2{}
\IfFileExists{supp.aux}{\externaldocument[S-]{supp}}{%
  \IfFileExists{build/supp.aux}{\externaldocument[S-]{build/supp}}{}%
}
\let\@writefile\XR@orig@writefile
\makeatother

\newcommand{\workatuiuc}{\textsuperscript{*}}

\begin{document}

\title{How To Teach Large Multimodal Models \texorpdfstring{\\}{ }New Skills?}


\author{
Zhen Zhu\inst{1,2}\workatuiuc \and
Yiming Gong\inst{1,3}\workatuiuc \and
Yao Xiao\inst{1} \and
Yaoyao Liu\inst{1} \and
Derek Hoiem\inst{1}
}

\institute{
University of Illinois Urbana-Champaign \and
Google DeepMind \and
Carnegie Mellon University \\
\email{\{zhenzhu4,yimingg8,yaox11,lyy,dhoiem\}@illinois.edu} \\
\textsuperscript{*} Work done at the University of Illinois Urbana-Champaign.
}

\maketitle

\begin{abstract}
  How can we teach large multimodal models (LMMs) new skills without erasing prior abilities?
  We study sequential fine-tuning on five target skills while monitoring general ability on eight held-out benchmarks across three model families.
  Surprisingly, we find that performance lost on held-out tasks after fine-tuning on one skill can partly recover when the model is subsequently tuned on a different skill.
  We trace this behavior to a measurable shift in the output token distribution, manifested through a simple counting-bias probe that shows the shift co-varies with forgetting.
  Guided by this insight, we identify two simple, robust tuning recipes that learn strongly while limiting drift:
  (i) updating only the self-attention projection layers (SA Proj., $\Delta$ learning +24.9 / $\Delta$ held-out forgetting -0.6), and
  (ii) updating only the MLP Gate\&Up while freezing the Down projection (+30.5 / -2.1).
  Both substantially outperform full-LLM tuning (+31.8 / -23.3) in the learning--forgetting trade-off.
  We also compare against common forgetting mitigation methods---Learning without Forgetting (LwF), LoRA, Mixture-of-Experts, and weight-space interpolation (WiSE-FT)---and find that our selective tuning recipes match or exceed their learning--stability balance while remaining simpler, requiring no replay, auxiliary parameters, or per-stage tuning.
  These results hold across LLaVA-OneVision, LLaVA-NeXT, and Qwen2.5-VL, confirming that the key to teaching LMMs new skills without forgetting lies in controlling output distribution shift by choosing which components to tune.
  Code will be made available.

\end{abstract}

\addtocontents{toc}{\protect\setcounter{tocdepth}{-1}}

\section{Introduction}

\begin{figure*}[htbp]
  \centering
  \includegraphics[width=\textwidth]{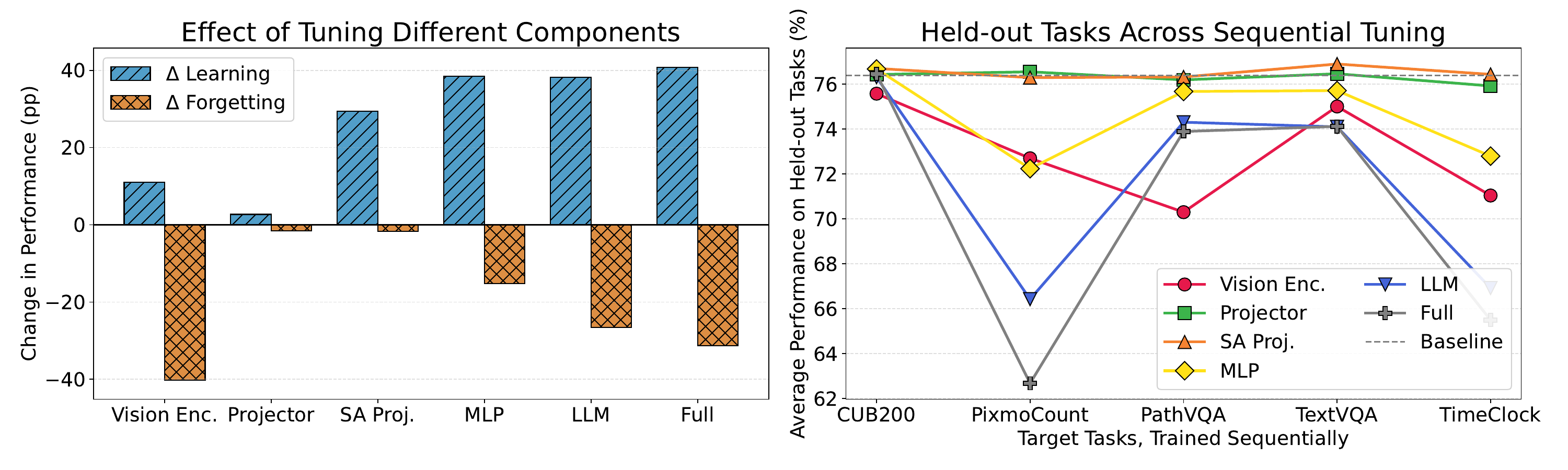}
  \caption{\small \textbf{Surprising Forgetting Behavior in LMMs}: \textbf{Left}: When fine-tuning most components on one target task, we see major improvement in that task (``Learning'') but a substantial drop in performance of other tasks (``Forgetting'', total across tasks shown here), as expected. But if we only tune self-attention projection layers (SA Proj.) in the language model, we still get substantial learning on the target task with minimal forgetting. 
  \textbf{Right}: Even fine-tuning SA Proj. for multiple tasks sequentially, we see no forgetting. For others, we see large forgetting on the PixmoCount task, but the models somehow partly recover what they "forgot" in learning the next specialized task. Our paper documents and analyzes these and other interesting phenomena of learning and forgetting in LMMs, leading to simple and effective ways to teach LMMs new skills.}
  \label{fig:teaser_new_new}
\end{figure*}


Large multimodal models (LMMs), such as LLaVA~\citep{liuVisualInstructionTuning2023} and Qwen2.5-VL~\citep{Bai_Qwen2.5-VL_2025}, are trained to generate natural language answers based on image(s) and natural language instruction. As such, these models can perform a wide range of tasks. However, for many special domains, such as medical images, or skills, such as counting, the models do not perform as desired.
How can we teach LMMs something new without degrading existing capabilities?



Training a new LMM can cost millions of dollars, weeks of time, and emit hundreds of tons of CO\textsubscript{2}; accordingly, practitioners often adapt existing models via fine-tuning rather than retraining from scratch.
Since some LMMs are trained with only a single pass over their data~\citep{liLLaVAOneVisionEasyVisual2024}, it is natural to keep adapting them as new data and use cases emerge.
However, naïve task-by-task fine-tuning can degrade broad vision--language ability (``catastrophic forgetting''), as reported in prior work~\citep{chenCoINBenchmarkContinual2024,Yu_BoostingContinualLearning_2024,Zhu2024ModelTailor}.
However, we show that forgetting is not uniform: it depends strongly on \emph{which parameters are updated}, and some performance drops after a narrow adaptation stage can partially reverse later.

To study the forgetting dynamics of LMMs systematically, we develop a controlled evaluation suite.
We sequentially fine-tune on five practical target skills spanning different answer formats (fine-grained bird classification, counting, medical VQA, OCR reading, and time reading), while monitoring general vision--language ability on eight widely used held-out benchmarks.
We report \emph{learning} as improvement on the current target task and \emph{forgetting} as the average drop on the held-out suite.
This protocol lets us compare tuning strategies on equal footing and directly measure the trade-off between acquiring new skills and preserving broad capabilities.



Our first goal is to identify tunable parts that deliver high target performance with minimal forgetting. We compare full‑model fine‑tuning to tuning each major component (vision encoder, projector, LLM) and then open the LLM into its two essential blocks—self‑attention projections (SA Proj.) and the feed‑forward network (MLP). Early experiments on \ov\ (Fig.~\ref{fig:teaser_new_new}) reveal two surprising results: 1) tuning SA Proj.\ learns with little or no measurable forgetting across a five‑task sequence; and 2) what appears forgotten after one stage can be recovered by tuning another specialized task.

These results lead us to ponder: why is SA Proj. tuning so robust to forgetting, and how is forgotten knowledge recovered without rehearsing? Consider the roles of the two essential components in the transformer decoder: self‑attention projections primarily route and combine information from the context~\citep{elhage2021mathematical,Olsson_InductionHeads_2022}, while MLPs act like key–value memories that project activated features back into the residual stream and thereby influence the next‑token distribution~\citep{Geva_TransformerFeedForwardLayersAreKeyValueMemories_2021}. We thus hypothesize that what appears to be forgetting after fine‑tuning on a narrow target task is often a bias in the output distribution induced by the task’s answer format. We test this on the \emph{counting} task: tuning the MLP improves the target performance but also increases the probability of emitting numeric tokens on a general \emph{captioning} task, and held‑out task accuracy drops accordingly. In contrast, tuning SA Proj.\ learns the counting skill without much bias to numeric tokens and without losing held‑out performance (Sec.~\ref{sec:dist-shift}).

Guided by this result, we explore tuning recipes that preserve learning while limiting output shift.
To avoid biasing the output distribution, we tune the MLP up/gating projections while keeping the down projection frozen, and find that it achieves similar learning to full MLP tuning with little forgetting. We experiment on \ov~\citep{liLLaVAOneVisionEasyVisual2024} by training five target tasks sequentially,  averaging over three sequence orders, measuring the learning and forgetting in target tasks and held-out tasks (Sec.~\ref{sec:sequential_tasks}).
We then confirm that similar trends hold for LLaVA-NeXT~\citep{li2024llavanext-strong} and Qwen2.5-VL~\citep{Bai_Qwen2.5-VL_2025} (Sec.~\ref{sec:other_backbones}).

To our knowledge, we are the {\bf first} to systematically study
rehearsal-free continual learning in large multimodal models under sequential
skill adaptation while monitoring performance on broad vision--language benchmarks, and to
show that apparent forgetting can be partially recoverable across stages.
\textbf{Our main contribution} is a mechanistically motivated view of \emph{where}
to update an LMM to acquire new skills while \emph{limiting output-distribution
drift}, together with a thorough evaluation suite spanning \textbf{five} practical target
skills, \textbf{eight} held-out benchmarks, and \textbf{three} model families.
Our key findings are:
\begin{itemize}[leftmargin=*, itemsep=2pt, topsep=2pt]
\item \textbf{Tuning the LLM (\learn{$\Delta$ learning +31.8}/\forget{$\Delta$ forgetting -23.3}) is critical for learning new tasks}, while tuning the vision encoder (\learn{+9.6}/\forget{-10.8}) brings little gain and harms general ability.
\item \textbf{Tuning only the self-attention projection weights (\learn{+24.9}/\forget{-0.6}) or the up\&gate layers of the MLP (\learn{+30.5}/\forget{-2.1}) provides excellent learning with limited forgetting}, evaluated on a five-target task sequence, eight held-out benchmarks, and three model families.
\item \textbf{Forgetting is largely a manifestation of output distribution shift.} We use a counting-bias probe to show that the rise in number‑token likelihood grows with MLP tuning and remains near baseline for self‑attention tuning; the magnitude of this shift covaries with performance drop in held‑out tasks. This explains why some methods -- tuning only SA Proj. or MLP Up\&Gate, or full tuning with distillation loss on the output tokens -- can learn with minimal forgetting.
\end{itemize}

\section{Related work}
Continual learning~\cite{Aljundii2017ExpertGate,Chen2018Lifelong,Chaudhry2019AGEM} aims to train models on a sequence of tasks or continuous data streams while preserving previously acquired knowledge. Much of the foundational work has been developed in closed-vocabulary image classification, where approaches can be broadly organized into three families:
(1) \emph{regularization} methods preserve knowledge from earlier model states by matching logits~\cite{li2017learning,Rebuffi17iCARL}, feature maps~\cite{Douillard2020PODNet}, or other representational statistics~\cite{Tao2020Topology,Wang2022FOSTER,Simon2021Learning,Joseph2022Energy,Pourkeshavarzi2022Looking,Liu2023Online};
(2) \emph{exemplar replay} methods maintain a reservoir of samples from previous training stages~\cite{Prabhu2020GDumb,Liu2024PlaceboCIL,Luo2023CVPR,Liu2020Mnemonics,Rebuffi17iCARL,Shin2017GenerativeReplay,Bang2021Rainbow} and revisit them during later phases to reinforce past knowledge; and
(3) \emph{network expansion} methods~\cite{Liu2020AANets,Wang2022FOSTER} increase model capacity for new target data while freezing a subset of parameters to retain prior knowledge. More recently, prompt tuning has emerged as an effective continual learning strategy~\cite{Jin2022GoodPrompt,Khattak2023multimodalprompt,Khattak2023selfregulatingprompts,coda_prompt_Smith_2023_CVPR}, keeping all model weights frozen while introducing learnable prompts to accommodate new tasks.

\paragraph{Large language models (LLMs).}
A growing body of work evaluates the learning-forgetting trade-off across various fine-tuning strategies for LLMs with billions of parameters, including full-model tuning, adapters, LoRA, and prompt tuning. Luo et al.~\cite{Luo_EmpericalStudy_2023} find that decoder-only models exhibit greater robustness to forgetting than encoder-decoder models.
Lin et al.~\cite{Lin_MitigatingAlignmentTax_2024}
observe that models fine-tuned for narrow domains can lose general-task ability, though weight interpolation~(WiSE-FT)~\cite{WiSE-FT} helps maintain a balance between specialization and retention. Biderman et al.~\cite{biderman2024lora} show that LoRA reduces both learning and forgetting relative to full fine-tuning.
Li et al.~\cite{li2025analyzing} propose a dual-memory replay framework with interpolated LoRA to further manage this balance.
Huang et al.~\cite{Huang_MitigatingForgetting_2024} generate pseudo-data from the model itself to mitigate forgetting without requiring access to original training data. Xiang et al.~\cite{xiang2023language} apply regularization strategies such as Elastic Weight Consolidation~(EWC)~\cite{EWC} and hierarchical importance-based penalties to preserve general knowledge by constraining updates to important parameters. Wang et al.~\cite{Wang2023Orthogonal} learn orthogonal LoRA weights for new tasks, reducing interference with previously learned representations. Concurrent with our work, Shenfeld et al.~\cite{shenfeld2025rlsrazoronlinereinforcement} report that the degree of forgetting correlates with the distributional shift between the base and tuned models (measured by KL divergence), and use this finding to explain why on-policy RL training is more robust to forgetting than supervised fine-tuning. 

\paragraph{Roles of attention and FFN in LLMs.}
Mechanistic studies of transformer blocks reveal a division of labor. 
Attention heads act primarily as \emph{routing} and retrieval mechanisms: they select \emph{where} to read from using query–key patterns and then mix the corresponding values; this view is formalized in the Transformer Circuits framework and supported by analyses of ``induction heads,'' which implement a simple copying algorithm and closely track the emergence of in‑context learning during training~\cite{Olsson_InductionHeads_2022}. Complementing this routing role, feed‑forward (FFN/MLP) blocks behave like \emph{key–value memories}: learned keys detect input patterns while values write features that align with groups of vocabulary items, thereby shifting the model's output preferences~\cite{Geva_TransformerFeedForwardLayersAreKeyValueMemories_2021}. Meng et al.~\cite{meng2022locating} show that directly modifying MLP weights updates specific facts while preserving unrelated behavior, providing further evidence that FFN layers serve as a principal site where factual associations are stored.  Earlier analyses in BERT and NMT have also found that a minority of attention heads specialize in linguistically interpretable roles (e.g., syntax, coreference) while many heads are prunable with little loss, reinforcing the view of attention as selective routing rather than the main repository of lexical knowledge~\cite{clark2019analysis,voita2019analyzing}. Our findings are consistent with this literature: self‑attention updates tend to preserve global behavior, while MLP updates are the main driver of output‑distribution shift.

\paragraph{Vision-text contrastive models}such as CLIP~\cite{radford_icml2021_clip}, are trained to align images and texts for open-vocabulary image classification and retrieval. While CLIP achieves strong zero-shot performance, it may underperform on fine-grained or specialized tasks~\cite{radford_icml2021_clip,treeprobe}, motivating the need for reliable continual learning approaches.
Zhu et al.~\cite{treeprobe,Zhu_AnytimeContinualLearning_2024} propose learning to blend predictions from original and tuned image encoders, enabling fast online learning without forgetting for open-vocabulary classification.
Yu et al.~\cite{Yu_BoostingContinualLearning_2024} add parameter-efficient adapters to a mixture-of-experts on a frozen CLIP model to prevent forgetting. Zhou et al.~\cite{zhou2025learning} design task-specific projection layers and cross-modal fusion modules for vision-language models in class-incremental learning. Liu et al.~\cite{liu2025c} incorporate continual low-rank adaptation and knowledge consolidation to prevent forgetting. Zheng et al.~\cite{zheng2023preventing} use knowledge distillation~\cite{li2017learning} on CLIP to maintain zero-shot performance. These methods focus on classification with frozen or lightly adapted encoders; our work instead addresses generative LMMs and investigates which decoder components to update.

\paragraph{Large multimodal models (LMMs).} Although continual learning in generative instruction-following LMMs remains relatively underexplored, interest is growing rapidly. Chen et al.~\cite{chenCoINBenchmarkContinual2024} find that LMMs suffer from catastrophic forgetting when learning a sequence of new tasks. Much of the existing work focuses on visual question answering~(VQA)~\cite{zhang2023vqacl,nikandrou2024enhancing,lin2025vlm,marouf2025no,del2020ratt} or image captioning~\cite{nguyen2019contcap}.
Zhang et al.~\cite{zhang2023vqacl} leverage both sample-specific and sample-invariant features to learn representations that are both discriminative and generalizable for VQA tasks.
Nikandrou et al.~\cite{nikandrou2024enhancing} distill knowledge separately for each modality, ensuring that both image features and question features retain their relevant information when new tasks arrive.
Lin et al.~\cite{lin2025vlm} combine selective memory replay and knowledge distillation for VQA.
Marouf et al.~\cite{marouf2025no} store only past questions from previous tasks as memory for rehearsal. More recent approaches rely on architectural expansions and dynamic routing, such as decoupling modalities via mixture-of-experts~\cite{wei2025mitigating} or using dynamic MLLM-based routing mechanisms~\cite{zhao2025mllmcl}. In earlier model generations,
Nguyen et al.~\cite{nguyen2019contcap} integrate continual learning techniques, such as finetuning schemas and regularization, into the captioning pipeline to address forgetting, and Del Chiaro et al.~\cite{del2020ratt} introduce an attention-based LSTM architecture.


In comparison, our study (i) evaluates rehearsal‑free sequential skill adaptation on five diverse target skills while monitoring eight broad held‑out benchmarks across three model families; (ii) identifies two simple, widely applicable tuning recipes, SA Proj.\ and MLP Gate\&Up (with Down frozen), that provide strong learning with limited forgetting; and (iii) provides a mechanistic analysis showing that much of the observed forgetting is tied to output‑token distribution drift, which can be limited (Sec.~\ref{sec:dist-shift}) or regularized (Sec.~\ref{sec:mitigate_forgetting_result}) without adding new modules or storing rehearsal data.
\section{Method}
\label{sec:method}

\begin{figure*}
    \centering
    \includegraphics[width=\textwidth]{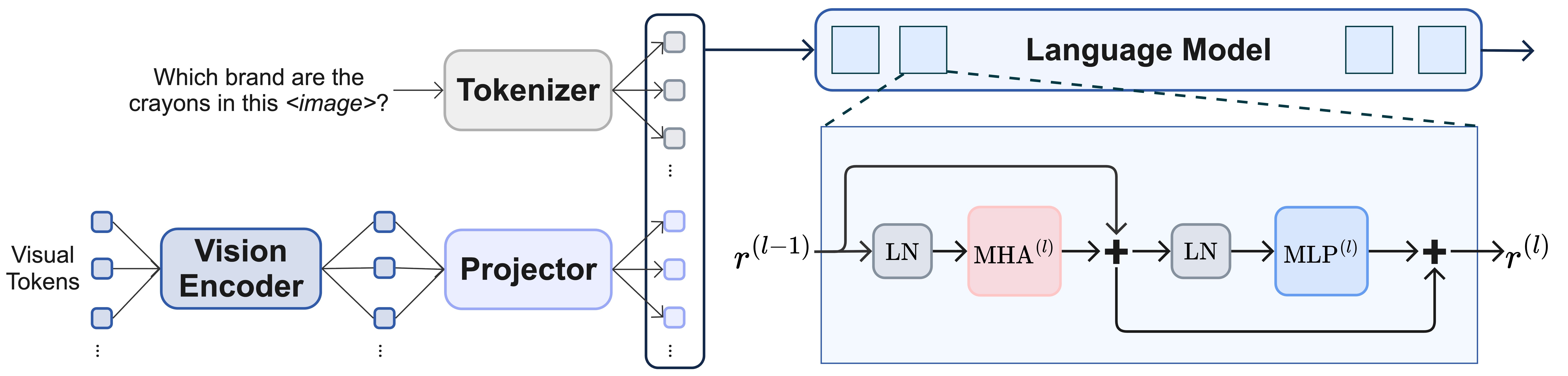}
    \caption{\small Architecture of our evaluated LMMs. The input contains visual inputs such as images or videos, which are converted to visual tokens by the vision encoder, and text input is processed by a tokenizer containing a visual placeholder token {\tt <image>}. Visual tokens are converted by the projector and concatenated with text tokens as input for the language model. We visualize the architecture of the transformer decoder layer of the language model. "LN", "MHA", and "MLP" represent layer norm, multi-head attention, and multi-layer perceptron, respectively. $r^{(l)}$ is the final output of layer $l$.
    }
    \label{fig:main_diagram}
  \end{figure*}
  

\paragraph{Setting.}
We adapt a pretrained large multimodal model on either a \emph{single‑target} task or a \emph{sequential} stream of tasks.
In the single‑target case, given a target dataset $\mathcal{D}_{\mathrm{T}}$ and a held‑out suite $\mathcal{D}_{\mathrm{H}}$, the goal is to improve performance on $\mathcal{D}_{\mathrm{T}}$ while preserving performance on $\mathcal{D}_{\mathrm{H}}$.
In the sequential case, tasks $\{\mathcal{D}_{\mathrm{T}}^{(1)},\ldots,\mathcal{D}_{\mathrm{T}}^{(K)}\}$ arrive in stages; unless noted, we update at each stage without rehearsal (no mixing of earlier tasks) and assess both the current target and the aggregated held‑out suite after every stage.

\subsection{Model}
\label{sec:model}

\paragraph{Overview of the LMM.}
Our evaluated LMMs have three major parts (Fig.~\ref{fig:main_diagram}): a vision encoder that turns an image into visual tokens, a projector that maps those tokens to the language width $d$, and a decoder‑only transformer language model that produces next‑token logits given the visual and text tokens.

\paragraph{Vision encoder and projector.}
The vision encoder produces $v=f_{\mathrm{vis}}(I)\in\mathbb{R}^{S_v\times d_v}$.
The projector maps to the language representation width,
\[
  x_{\mathrm{vis}}=g_{\psi}(v)\in\mathbb{R}^{S_v\times d},
\]
where $\psi$ are the projector’s trainable parameters.

\paragraph{Language model.}
The language model is a pre‑norm, decoder‑only transformer with $L$ identical blocks.
As illustrated in Fig.~\ref{fig:main_diagram}, the sublayer outputs and residual update at block $l$ are
\begin{equation}
  \label{eq:block-update}
  \adjustbox{max width=0.9\linewidth}{
    $
    a^{(l)}=\mathrm{MHA}^{(l)}\!\big(\mathrm{LN}(r^{(l-1)})\big),\quad
    f^{(l)}=\mathrm{MLP}^{(l)}\!\big(\mathrm{LN}(r^{(l-1)}+a^{(l)})\big),\quad
    r^{(l)}=r^{(l-1)}+a^{(l)}+f^{(l)}, $
  }
\end{equation}
where MHA and MLP denote the \textbf{multi‑head self‑attention} and \textbf{feed‑forward network}, respectively, and LN is layer normalization. 

\noindent\emph{Self‑attention.}
With input $x$ and per‑head key width $d_k$,
\begin{equation}
  \label{eq:attn}
  \scalebox{0.85}{$
    Q=xW_Q,\;
    K=xW_K,\;
    V=xW_V,\;
    A=\mathrm{softmax}\!\left(\frac{QK^\top}{\sqrt{d_k}}\right),\;
    \mathrm{MHA}(x)=(AV)W_O
  $}
\end{equation}
Here $W_Q,W_K$ determine \emph{where} to attend (routing), $W_V$ selects what content is mixed, and $W_O$ projects the mixed value features back into the residual stream at model width $d$.

\noindent\emph{Feed‑forward.}
With input $x$ and gating nonlinearity $\phi=\mathrm{SiLU}$,
\begin{equation}
  \label{eq:mlp}
  \scalebox{0.85}{$
      \mathrm{MLP}(x)=W_{\mathrm{down}}\!\left(\phi(xW_{\mathrm{gate}})\odot (xW_{\mathrm{up}})\right),
  $}
\end{equation}
where $\odot$ denotes the elementwise (Hadamard) product. Matrices $W_{\mathrm{gate}},W_{\mathrm{up}}$ detect features (key‑like pattern match), and $W_{\mathrm{down}}$ projects the activated features back to the residual stream at width $d$. 
We use $U\in\mathbb{R}^{d\times |V|}$ for the LM head (unembeddings) and denote the final block output as $r^{(L)}$, so the output logits are $z=U^\top r^{(L)}$.

\paragraph{Residual stream.}
Let $x_{\text{text}}\!\in\!\mathbb{R}^{S_t\times d}$ be the text embeddings and $x_{\mathrm{vis}}\!\in\!\mathbb{R}^{S_v\times d}$ the projected visual tokens.
The transformer input stacks them along the sequence:
$r^{(0)}=\big[\,x_{\text{text}};\,x_{\mathrm{vis}}\,\big]$.
Because the model relies on additive residual connections, the final output state $r^{(L)}$ is exactly the sum of this initial state and the accumulated updates from all subsequent layers:
\begin{equation}
\label{eq:residual_sum}
\scalebox{0.9}{$
  r^{(L)} = r^{(0)} + \sum_{l=1}^L \big( a^{(l)} + f^{(l)} \big).
$}
\end{equation}
This formulation highlights two critical points. First, the initial multimodal representation $r^{(0)}$ is a direct contributor to the final output logits $z$. Second, the sublayers are highly coupled; the key write-back paths—$W_O$ (attention output) and $W_{\mathrm{down}}$ (MLP output)—inject new content into this shared stream, layer by layer, directly shifting the final predictions.

\subsection{Which parts to tune?}
\label{sec:groups}

When keeping the token embeddings, the LM head $U$, and layer‑norm parameters frozen, Eq.~\ref{eq:residual_sum} highlights the importance of two primary components: the initial state $r^{(0)}$ (driven by the \textbf{vision encoder} and \textbf{projector}) and the \textbf{language model} submodules (which accumulate into $r^{(L)}$). 

Within the language model submodules, our selection of tunable parameters is guided by mechanistic studies of transformer components~\cite{Geva_TransformerFeedForwardLayersAreKeyValueMemories_2021,Olsson_InductionHeads_2022}. Prior work suggests that self‑attention layers act primarily as routers that associate and organize information from the context, while MLPs behave like key–value memories that project activated features back into the residual stream, strongly influencing the next‑token distribution. We hypothesize that much of rehearsal‑free forgetting is driven by unconstrained write‑back that biases this output distribution. Therefore, updating components that mainly change routing (SA Proj.) or updating only the MLP’s feature detectors while freezing its write‑back projection ($W_{\mathrm{down}}$) ideally should help limit global output drift and reduce interference.
Guided by this idea and combining Eqs.~\ref{eq:attn}–\ref{eq:residual_sum}, we consider:
\begin{itemize}[leftmargin=*, itemsep=2pt, topsep=2pt]
  \item \textbf{SA Proj.}: Update $W_Q,W_K,W_V,W_O$ in all blocks (routing + write‑back for attention).
  \item \textbf{SA Proj. (QKV)}: Freeze $W_O$ to emphasize routing without directly modifying write-back.
  \item \textbf{MLP}: Update $W_{\mathrm{gate}},W_{\mathrm{up}},W_{\mathrm{down}}$ (concept activation + write‑back).
  \item \textbf{MLP (Gate \& Up)}: Update $W_{\mathrm{gate}},W_{\mathrm{up}}$ while freezing $W_{\mathrm{down}}$ to regulate write‑back.
\end{itemize}

\subsection{Training objective}
\label{sec:objective}


\paragraph{Target task loss.}
We use next‑token cross‑entropy on the current target dataset with teacher forcing. For a batch $\mathcal{B}\subset\mathcal{D}_{\mathrm{T}}^{(k)}$ at stage $k$,
\begin{equation}
  \scalebox{0.9}{$
    \mathcal{L}_{\mathrm{task}}(\theta)=
    \mathbb{E}_{(I,y)\sim\mathcal{B}}\!\left[-\sum_{t=1}^{|y|}
    \log p_\theta\big(y_t\mid y_{<t},\,x_{\mathrm{vis}}\big)\right],
    \qquad x_{\mathrm{vis}}=g_{\psi}\!\left(f_{\mathrm{vis}}(I)\right).
  $}
\end{equation}

\paragraph{Learning‑without‑Forgetting (optional).}
To curb the output‑distribution drift, we can enforce a KL-divergence constraint between the outputs of the current model at stage $k$ with a frozen teacher model (checkpoint after stage $k{-}1$). Let $\theta_{k-1}$ be the frozen teacher and $\theta$ the current model tuned on $\mathcal{D}_{\mathrm{T}}^{(k)}$. The objective is
\begin{equation}
\scalebox{0.9}{$
  \mathcal{L}(\theta)
  =\mathcal{L}_{\mathrm{task}}(\theta)\;+\;\lambda\,\mathcal{L}_{\mathrm{distill}}(\theta;\theta_{k-1}),
   $}
\end{equation}
with
\begin{equation}
  \scalebox{0.9}{$
    \mathcal{L}_{\mathrm{distill}}(\theta;\theta_{k-1})
    =\mathbb{E}_{(I,y)\sim\tilde{\mathcal{B}}}\!\left[
      \frac{\tau^2}{|\mathcal{S}(y)|}\sum_{j\in\mathcal{S}(y)}
      \mathrm{KL}\!\Big(
        \mathrm{softmax}\!\big(\tfrac{z_{\theta_{k-1},j}}{\tau}\big)
        \;\big\|\;
        \mathrm{softmax}\!\big(\tfrac{z_{\theta,j}}{\tau}\big)
    \Big)\right],
  $}
\end{equation}
where $\tilde{\mathcal{B}}\!\subset\!\mathcal{D}_{\mathrm{T}}^{(k)}$ is a target minibatch, $\tau$ is the distillation temperature, and $\mathcal{S}(y)$ is a uniformly random subset of positions with $|\mathcal{S}(y)|=\min(|y|,1000)$ so we distill over many tokens while capping compute/memory. The coefficient $\lambda$ balances fitting the new supervision against preserving the model’s earlier behavior.

\section{Experiment Design}


Our experiments are designed to answer four questions:
1) \textbf{Where to tune?}—which components of an LMM can be updated to learn new skills while preserving prior abilities (Sec.~\ref{sec:sequential_tasks});
2) \textbf{Why does forgetting occur?}—whether performance loss is tied to a shift in the model’s output distribution (Sec.~\ref{sec:dist-shift});
3) \textbf{How generalizable is our selective tuning strategy}—whether the same selective‑tuning recipes (SA Proj., MLP Gate\&Up) transfer across model families (Sec.~\ref{sec:other_backbones});
4) \textbf{How does our selective tuning compare to other simple forgetting mitigation approaches?} (Sec.~\ref{sec:mitigate_forgetting_result}). Due to space limits, we provide additional task/implementation details, per-task tables,
and extended analyses in the Appendix.



\subsection{Tasks and evaluation suites}
\label{sec:datasets}

\paragraph{Target tasks.}
We select {\bf 5 target tasks} that are common in real-world usage and are typically challenging for LMMs. Then we create a {\bf default sequential-tuning task curriculum}: {\bf 1) Bird classification} from the CUB dataset~\citep{CUB}. We reformat the dataset following the instructions of~\citep{liu2024revisit_llm} for training and evaluating LMMs; {\bf 2) Counting} from the PixmoCount dataset~\citep{deitkeMolmoPixMoOpen2024a}; {\bf 3) Medical VQA} from the PathVQA dataset~\citep{He2020PathVQA}; {\bf 4) OCR reading} from the TextVQA dataset~\citep{Sign2019TextVQA} which has 34,602 training samples; {\bf 5) Time reading} from the TimeClock dataset~\citep{gpiosenka_time_image_datasetclassification}. In total, the curriculum contains 107,910 training samples, providing a comprehensive stress test for forgetting and knowledge transfer.

\paragraph{Held‑out suite.}
To measure generalization beyond the training stream, we evaluate on eight held‑out benchmarks:
AI2D \citep{kembhavi2016ai2d}, ChartQA \citep{masry2022chartqa}, DocVQA \citep{mathew2021docvqa}, InfoVQA \citep{mathew2022infovqa}, RealWorldQA \citep{visheratin_realworldqa_2024}, SeedBench \citep{li2023seedbench}, ScienceQA \citep{lu2022scienceqa}, and MMStar \citep{chen2024mmstar}.
Several of these are multi-skill evaluation suites (e.g., SeedBench, MMStar), so the held‑out average reflects a broad collection of capabilities rather than a single narrow dataset.
InfoVQA and DocVQA use ANLS; since ANLS is in $[0,1]$, we average it with accuracies from the other held‑out tasks when reporting the mean held‑out score.

\subsection{Sequence‑level metrics}
\label{sec:seq-metrics}

We summarize performance over the five‑stage curriculum with four metrics computed for every method.

\begin{itemize}[leftmargin=*, itemsep=2pt]
\item \textbf{Target Learning.}
At each stage, consider only the task being tuned and measure its improvement over the base model on that task. We then average these stage‑wise gains across all stages. This captures how well a method learns the task it is currently trained on.

\item \textbf{Target Forgetting.}
To measure forgetting on target tasks trained earlier in the sequence, we report the average difference between their accuracy immediately after they were trained and their accuracy at the end of the sequence. More negative means more forgetting.

\item \textbf{Target Overall.}
After training the full sequence, we compute the average performance change vs.\ the base model across all target tasks. This yields the net end‑of‑sequence effect on the target suite, combining the learned task and the previously learned targets.

\item \textbf{Held‑out Forgetting.}
After training the full sequence of target tasks, we measure the average performance across all eight held-out benchmarks, in comparison to the base model. Negative values indicate forgetting on general vision–language ability; positive values indicate positive transfer.
\end{itemize}

\section{Results}

\subsection{Component tuning on \ov}
\label{sec:sequential_tasks}

\begin{table*}[t]
\setlength\tabcolsep{6pt}
\centering
\small
\caption{\small \textbf{Effect of tuning different components: learning, forgetting, and overall performance, averaged over three five-task sequences.}
Cells are colored using a blue-orange colormap to show performance changes. \colorbox{CornflowerBlue!110}{Blue} indicates a positive change, where a darker shade is better. \colorbox{BurntOrange}{Orange} indicates a negative change, where a lighter shade is better. \rebuttaledit{We \underline{underline} numbers that do not reflect a significantly different task-average distribution from the best, based on a two-sided paired sample t-test.}}
\label{tab:sequential_effects_component_summary}
\resizebox{\textwidth}{!}{%
\begin{tabular}{lcccc}
\toprule
Method & Target Learning & Target Forgetting & Target Overall & Held-out Forgetting \\ 
\midrule
Baseline & 43.9 & 0.0 & 43.9 & 76.4 \\ 
\midrule
Full & \applycolorUL{29.9} & \applycolor{-25.9} & \applycolor{9.2} & \applycolor{-27.4} \\ 
\quad \texttt{|--} Vision Encoder & \applycolor{9.6} & \applycolor{-12.7} & \applycolor{-0.5} & \applycolor{-10.8} \\ 
\quad \texttt{|--} Projector & \applycolor{2.3} & \applycolorUL{-0.8} & \applycolor{1.7} & \applycolor{-1.3} \\ 
\quad \texttt{|--} LLM & \applycolorUL{31.8} & \applycolor{-23.5} & \applycolor{13.0} & \applycolor{-23.3} \\ 
\quad\quad \texttt{|--} SA Proj. & \applycolor{24.9} & \applycolor{-2.3} & \applycolor{23.1} & \applycolor{-0.6} \\ 
\quad\quad \texttt{|--} SA Proj. (QKV) & \applycolor{14.9} & \applycolorUL{-0.5} & \applycolor{14.5} & \applycolorUL{0.2} \\ 
\quad\quad \texttt{|--} MLP & \applycolorUL{31.1} & \applycolor{-19.5} & \applycolor{15.5} & \applycolor{-15.7} \\ 
\quad\quad \texttt{|--} MLP (Gate\&Up) & \applycolorUL{30.5} & \applycolor{-4.2} & \applycolorUL{27.1} & \applycolor{-2.1} \\ 
\bottomrule
\end{tabular}}
\end{table*}

\Cref{fig:teaser_new_new} previews learning (single–task tuning, left) and forgetting (held‑out along the default sequence, right). \Cref{tab:sequential_effects_component_summary}  summarizes the four sequence‑level metrics on \ov\ for each component‑tuning configuration, averaged over three five‑task curricula whose task sequences and detailed results are in the Appendix. Entries are percentage‑point deltas from the base model; the baseline row reports absolute scores.
For each column, we run paired sample t-tests on the per-sequence/per-task averages to test whether tuning different components leads to significantly different per-task learning, forgetting, or overall performance.
We underline numbers that are \textit{not} significantly different ($p$ > 0.1) than the best.
Detailed single‑task and sequential results, together with per‑task performance tables, are provided in the Appendix.

From the table, we find the following patterns:
\begin{itemize}[leftmargin=*, itemsep=2pt, topsep=2pt]
\item \textbf{Full‑model tuning attains large learning but maximizes forgetting}: Target Learning \(+29.9\) is coupled with the worst Target/Held‑out Forgetting (\(-25.9\)/\(-27.4\)).

\item \textbf{Vision‑side updates are weak or near‑neutral}: the vision encoder yields a modest Target Learning \(+9.6\) with negative Target Overall \(-0.5\) and Held‑out Forgetting \(-10.8\); projector‑only updates result in little change.

\item \textbf{Language‑model tuning has the best learning}: LLM shows the strongest Target Learning \(+31.8\) and a solid Target Overall \(+13.0\), but still substantial Target/Held‑out Forgetting (\(-23.5\)/\(-23.3\)).

\item \textbf{Self‑attention projection is the most stable among LLM choices}: SA Proj.\ achieves high Target Overall \(+23.1\) with minimal forgetting (Target \(-2.3\), Held‑out \(-0.6\)); the conservative variant without \(W_O\) further reduces forgetting (Target \(-0.5\), Held‑out \(+0.2\)) at the cost of learning (+14.5 Target Overall), indicating over‑regularization when the attention write‑back is frozen.

\item \textbf{Regulating the MLP write‑back offers the best balance}: MLP (Gate\&Up) delivers near‑maximal Target Learning \(+30.5\) and the highest Target Overall \(+27.1\) while keeping forgetting small (Target \(-4.2\), Held‑out \(-2.1\)); by contrast, full‑MLP pushes learning slightly higher on the current task \(+31.1\) but increases forgetting (Target \(-19.5\), Held‑out \(-15.7\)).
\end{itemize}

Overall, methods that mainly modify routing/associations (SA Proj. and SA Proj. (QKV)) or regulate how activated features are projected back (MLP Gate\&Up) provide the most favorable learning–stability trade‑off on \ov.

\subsection{Output‑distribution probe (counting bias)}
\label{sec:dist-shift}

\begin{figure*}[htbp]
  \centering
  \includegraphics[width=\textwidth]{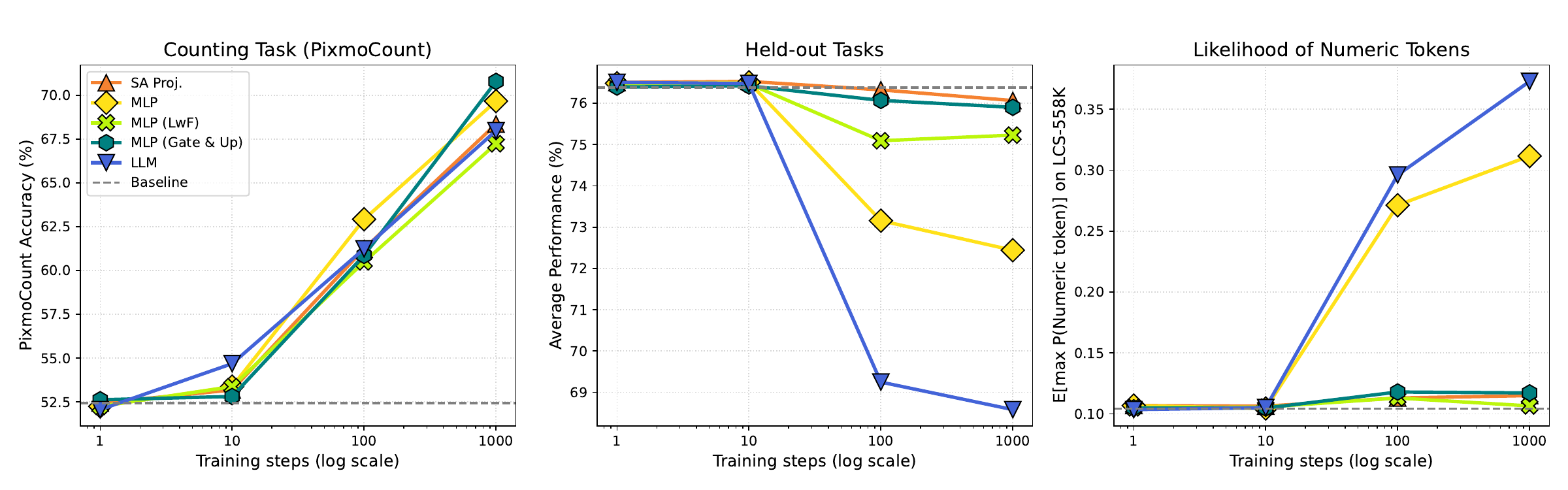}
\caption{\small \textbf{Learning–forgetting tracks output–distribution shift.}
On LLaVA‑OneVision tuned for counting, we plot five curves over log‑spaced steps for {LLM}, {SA Proj.}, {MLP}, {MLP (Gate\&Up)} and {MLP (LwF)}. The dashed line represents the base model.
\textbf{Left:} PixmoCount accuracy rises for all methods.
\textbf{Middle:} mean held‑out performance drops sharply for LLM and MLP, remains nearly unchanged for SA Proj., and is preserved by MLP (LwF); 
\textbf{Right:} the expected likelihood of number tokens on non‑counting captions (LCS‑558K~\citep{liu2023visualinstructiontuning}) surges for LLM and MLP, stays near baseline for SA Proj., and has little changes for MLP (LwF).}
  \label{fig:drift_on_counting}
\end{figure*}

Forgetting could be caused by interference or loss of access to memory (as commonly thought), or by biasing responses. We measure numeric-token bias (NTB) before, during, and after training on counting. Concretely, let $C$ be a fixed set of numeric tokens (digits and common spelled numerals), and let $\mathcal{B}=\{(I,y)\}$ be a fixed held-out batch of captioning examples, where $I$ is an image and $y$ its reference caption. At training step $s$ (with parameters $\theta_s$), we generate a caption $\hat{y}$ for each image $I$ using greedy decoding. At each decoding position $j$:
\begin{equation}
\notag
\scalebox{0.95}{$
    \mathrm{NTB}_s
=\frac{1}{|\mathcal{B}|}\sum_{(I,y)\in\mathcal{B}}
\left[\frac{1}{|\hat{y}|}\sum_{j=1}^{|\hat{y}|}\;
\max_{v\in C}\;p_{\theta_s}\!\bigl(v \mid \hat{y}_{<j},\,x_{\mathrm{vis}}(I)\bigr)\right],
$}
\end{equation}
where $\hat{y}_{<j}$ denotes the generated prefix before position $j$ and $p_{\theta_s}(\cdot\mid \hat{y}_{<j},x_{\mathrm{vis}}(I))$ is the next-token distribution.

\Cref{fig:drift_on_counting} shows that the likelihood of outputting numeric tokens for a captioning task is highly correlated with forgetting (Spearman $\rho=0.84$; Appendix~\ref{sec:counting-bias}), over training steps and across methods. We further show the forgetting effect is also reversible with subsequent fine-tuning (Appendix~\ref{sec:output-dist-dynamics}, Fig.~\ref{fig:recovery_probe}), and layer-wise analysis attributes most of the induced logit shift to MLP blocks (Fig.~\ref{fig:layer_contribution_to_shift} in Appendix). These results support the hypothesis that response bias is a dominant cause of forgetting.

\begin{figure*}[ht]
  \centering
  \includegraphics[width=\textwidth]{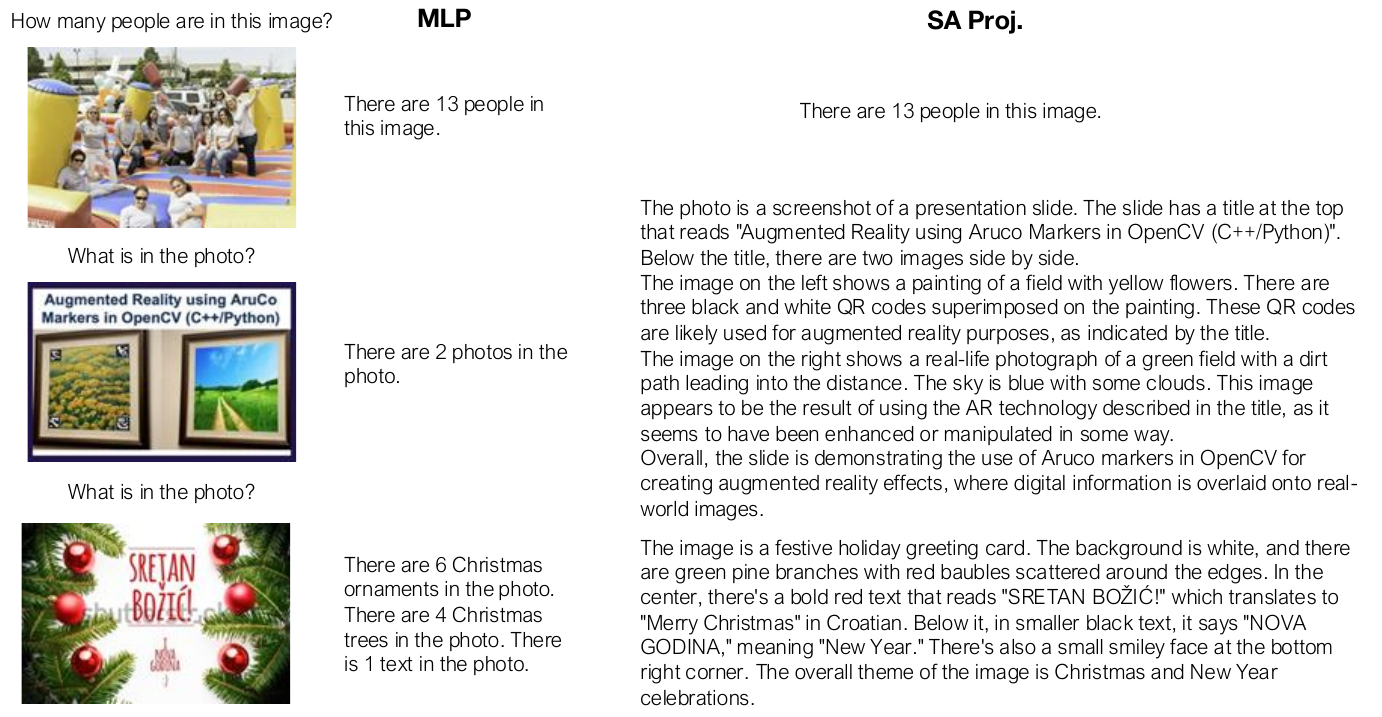}
  \caption{Visualizations on counting and captioning examples after tuning tuning MLP and SA Proj. on the counting task. The counting and examples are sampled from the PixmoCount dataset and the LCS‑558K~\citep{liu2023visualinstructiontuning} dataset, respectively.}
  \label{fig:counting_caption_viz}
\end{figure*}

\paragraph{Qualitative results on counting and captioning.}
~\Cref{fig:counting_caption_viz} contrasts the output on counting and captioning tasks for models that have tuned MLP or SA Proj weights for 1K steps on PixmoCount.
Both answer counting questions correctly, but the model with SA Proj tuning retains detailed captioning ability, while the MLP tuning leads to reframing captioning as counting statements, such as ``There are 2 photos in the photo'' (second row).  This pattern suggests that MLP tuning does not erase visual understanding (objects and relations are still recognized) but biases the output distribution.

\subsection{Beyond \ov: generalization to other backbones}
\label{sec:other_backbones}

\begin{table*}[t]
\setlength\tabcolsep{2pt}
\centering
\small
\caption{\small Component-level tuning experiments with \textbf{LLaVA-NeXT} and \textbf{Qwen2.5-VL}. "T" represents "Target" and "H" is for "Held-out". Underlined text of each column denotes the best method.}
\label{tab:llama3_component_summary}
\resizebox{\textwidth}{!}{%
\begin{tabular}{ l | cccc | cccc}
\toprule
\multirow{2}{*}{\textbf{Method}} & \multicolumn{4}{c|}{\textbf{LLaVA-NeXT (LLaMA-3 8B)}} &
\multicolumn{4}{c}{\textbf{Qwen2.5-VL (7B)}} \\
\cmidrule(lr){2-9}
  & T. Learn & T. Forget & T. Overall & H. Forget & T. Learn & T. Forget & T. Overall & H. Forget \\
\midrule
Baseline & 31.5 & 0.0 & 31.5 & 59.9 & 52.1 & 0.0 & 52.1 & 77.9 \\ 
\midrule
Full & \applycolor{31.7} & \applycolor{-20.3} & \applycolor{15.4} & \applycolor{-32.0} & \applycolor{17.3} & \applycolor{-5.2} & \applycolor{13.1} & \applycolor{-17.5} \\ 
\quad \texttt{|--} Vision + Projector & \applycolor{0.1} & \applycolorUL{-1.8} & \applycolor{-1.3} & \applycolor{-13.4} & \applycolor{12.1} & \applycolor{-9.1} & \applycolor{4.9} & \applycolor{-6.2} \\ 
\quad \texttt{|--} LLM & \applycolorUL{36.2} & \applycolor{-21.2} & \applycolor{19.3} & \applycolor{-35.9} & \applycolor{16.8} & \applycolor{-5.9} & \applycolor{12.1} & \applycolor{-24.6} \\ 
\quad\quad \texttt{|--} SA Proj. & \applycolor{28.3} & \applycolor{-7.9} & \applycolor{21.9} & \applycolorUL{-7.7} & \applycolor{16.1} & \applycolor{-1.6} & \applycolor{14.9} & \applycolorUL{0.6} \\ 
\quad\quad \texttt{|--} MLP & \applycolor{34.9} & \applycolor{-10.3} & \applycolorUL{26.6} & \applycolor{-16.3} & \applycolorUL{17.7} & \applycolor{-4.8} & \applycolor{13.9} & \applycolor{-10.9} \\ 
\quad\quad \texttt{|--} MLP ($W_{\text{gate}},W_{\text{up}}$) & \applycolor{28.0} & \applycolor{-8.9} & \applycolor{20.9} & \applycolor{-8.7} & \applycolor{16.8} & \applycolorUL{0.4} & \applycolorUL{17.1} & \applycolor{-4.6} \\ 
\bottomrule
\end{tabular}}
\end{table*}

We further repeat the default five‑task curriculum on two additional backbones, \textbf{LLaVA‑NeXT} (LLaMA‑3 8B) and \textbf{Qwen2.5‑VL} (7B), using the same training protocol and sequence‑level metrics as for \ov. For vision side updates, we tune the vision encoder and projector jointly, since they form a single interface that produces the visual token sequence consumed by the language model.

Across both backbones, the broad picture echoes \ov: updating the {language model} is consistently effective for learning new skills; full‑model and full‑LLM tuning achieve large target task gains but come with the largest drops on held‑out. Within the LM, two settings stand out as robust: {self‑attention projections} deliver meaningful target learning with small held‑out change, and {MLP (Gate\&Up)} preserves most of the learning of full‑MLP while limiting forgetting. There are, however, model‑specific nuances worth noting. For example, on \textbf{LLaVA‑NeXT}, MLP achieves the greatest target‑overall improvement but incurs a noticeably larger held‑out decrease than SA Proj.\ or Gate\&Up, which remain the most stable choices. LLaVA-NeXT is much more susceptible to forgetting in general than the other models. On \textbf{Qwen2.5‑VL}, SA Proj.\ is particularly stable: held‑out performance is maintained or slightly improved; MLP (Gate\&Up) attains the best target‑overall score with near‑zero target forgetting; vision\,+\,projector tuning also yields non‑trivial target gains with moderate stability cost, in contrast to its weaker effect on LLaVA‑NeXT. Though Qwen2.5-VL appears to learn less, it is worth noting that its baseline performance is much higher than that of other models.
Overall, taking \ov, LLaVA‑NeXT, and Qwen2.5‑VL together, the clearest cross‑model takeaway is to prefer {SA Proj.} when stability on held‑out is paramount and {MLP (Gate\&Up)} when seeking near‑maximal target learning with limited forgetting; projector‑only updates are generally weak, and full‑model / full‑MLP tuning maximizes short‑term gains at a clear stability cost.

\subsection{Mitigating forgetting}
\label{sec:mitigate_forgetting_result}

\vspace{-2mm}
\begin{figure*}[htbp]
  \centering
  \includegraphics[width=\textwidth]{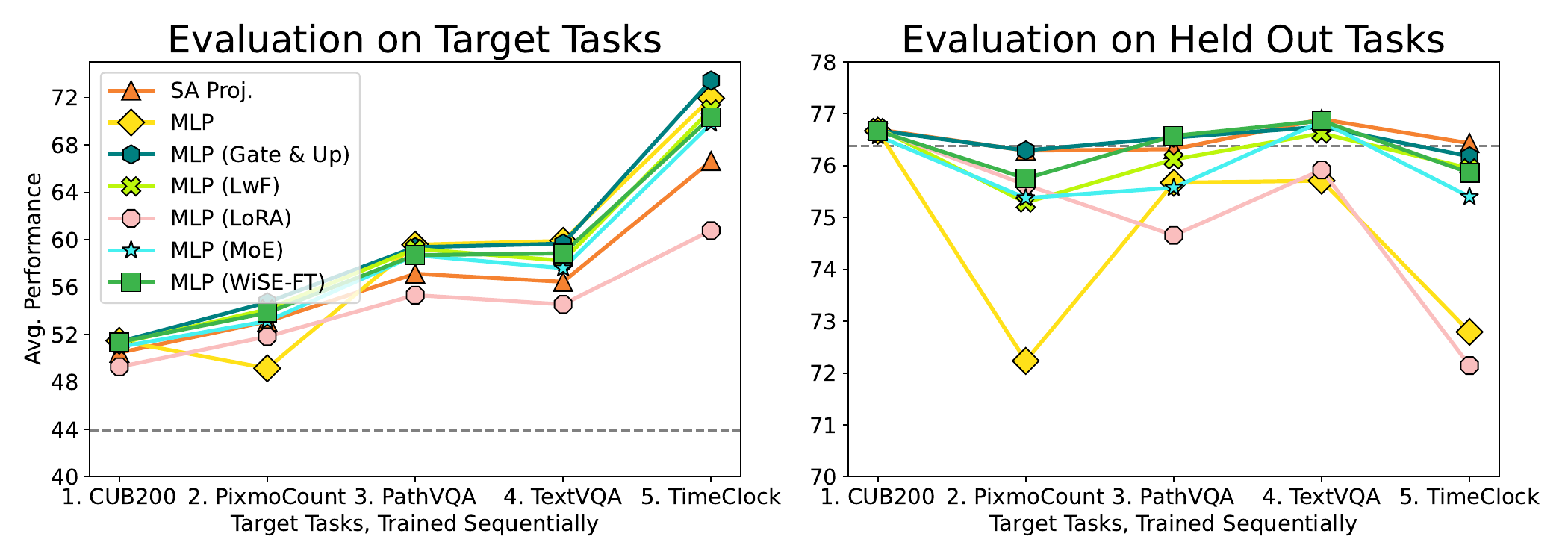}
  \caption{\small Comparison of different continual learning techniques in the default sequential task curriculum. For LwF, WiSE-FT, only the MLP layers are tuned. LoRA adapters are wrapped only on the MLP layers. MoE is also applied to the MLP layers.
  }
  \label{fig:method_comparison}
\end{figure*}

\paragraph{Comparisons to Continual Learning Baselines.} Recent continual learning methods for multimodal models often rely on highly specialized architectures or are restricted to discriminative VQA settings~\cite{lin2025vlm,nikandrou2024enhancing,zhang2023vqacl}. Because our goal is to evaluate open-ended generative capabilities across diverse skills without storing rehearsal data, we benchmark against the fundamental, widely applicable pillars of continual learning: Regularization (LwF)~\cite{li2017learning}, Parameter-Efficient Tuning (LoRA)~\cite{huLoRALowRankAdaptation2022}, Model Expansion (MoE), and Weight Averaging (WiSE-FT)~\cite{WiSE-FT}. See Appendix Sec.~\ref{sec:forgetting_mitigation_approach} for method details.

\Cref{fig:method_comparison} compares our selective tuning recipes to these baselines. Two patterns emerge. First, SA Proj.\ and MLP (Gate\&Up) provide the best learning–stability trade-off: SA Proj.\ keeps held-out performance essentially flat while achieving meaningful target gains, and MLP (Gate\&Up) delivers stronger target improvements with only a small held-out change, being substantially more stable than MLP. Second, among the compared methods, WiSE-FT can preserve held-out accuracy better than LwF but requires careful selection of task-dependent blending coefficients; LwF reliably curbs forgetting yet may impact target task gains; MoE and LoRA do not match the learning–stability balance of SA Proj.\ or MLP (Gate\&Up), with LoRA often lagging behind on target performance. Also note that selective tuning is highly training-efficient. When evaluated on 4$\times$H100 GPUs, our SA Proj. method achieves a throughput of 1.46 samples/sec, outperforming LoRA (1.27 samples/sec) and MoE (0.44 samples/sec), as it avoids the forward/backward pass overhead associated with adapter modules and routing logic (detailed in Appendix F.5). Overall, selectively tuning SA Proj.\ or the MLP Gate\&Up pair matches or exceeds these mitigation methods while remaining simple (no extra modules, no replay, no per-stage weight blending)

\section{Conclusion}
\label{sec:conclusion}

To answer our titular question, we can teach LMMs new skills by restricting tuning to SA Proj. or MLP Gate\&Up layers or applying LwF-like knowledge distillation~\cite{li_2018_lwf_pami}. These approaches all limit forgetting by constraining output distribution shift, leading to strong performance in serial task learning while retaining existing capabilities. We demonstrate the strong relationship between measured forgetting and output shift with a counting bias probe and layerwise logit analysis, explaining how ``forgetting'' from fine-tuning on one task can be recovered by fine-tuning on another. These results and recommendations for how to manage the learning-forgetting trade-off hold across three model families, as validated with five target skills and eight broad benchmarks.

\paragraph{Limitations.}A primary limitation of our work is that our evaluation does not extend to massive or commercial multimodal models. Furthermore, we have not tested whether the selective tuning recipes are still effective when scaling the training data to billions of samples.

\section*{Acknowledgment}
This work is supported in part by ONR award N00014-23-1-2383, and U.S.~DARPA ECOLE Program No.~\#HR00112390060.
The views and conclusions contained herein are those of the authors and should not be interpreted as necessarily representing the official policies, either expressed or implied, of DARPA, ONR, or the U.S. Government. We used GPUs at NCSA Delta through allocation  CIS230397 from the Advanced Cyberinfrastructure Coordination Ecosystem: Services \& Support (ACCESS)
program, which is supported by U.S. National Science Foundation grants 2138259, 2138286, 2138307,
2137603, and 2138296.

\clearpage  


\bibliographystyle{splncs04}
\bibliography{ref}


\addtocontents{toc}{\protect\setcounter{tocdepth}{2}}
\appendix
\clearpage



\appendix

\tableofcontents

\section{Task design}
\label{sec:tasks}

\subsection{Tasks and evaluation suites}
\label{sec:app-datasets}

\paragraph{Target tasks.} Our criteria for target task selection are: (1) prefer tasks that occur in daily experience, like counting and reading clocks preferred; (2) prefer tasks that LMMs are known to be typically weak, such as fine-grained species classification~\citep{liu2024revisit_llm}; and (3) exclude tasks used to train the original \ov\ model.
Based on this, we select {\bf 5 target tasks} and create a {\bf default sequential-tuning task curriculum}:
\begin{enumerate}[leftmargin=*, itemsep=2pt]
  \item {\bf Bird classification} from the CUB dataset~\citep{CUB} which contains 5,994 training samples. We reformat the dataset following the instructions of~\citep{liu2024revisit_llm} for training and evaluating LMMs.
  \item {\bf Counting} from the PixmoCount dataset~\citep{deitkeMolmoPixMoOpen2024a} which contains 36,140 training samples with object count labels.
  \item {\bf Medical VQA} from the PathVQA dataset~\citep{He2020PathVQA} which contains 19,654 radiology question answer pairs.
  \item {\bf OCR reading} from the TextVQA dataset~\citep{Sign2019TextVQA} which has 34,602 training samples.
  \item {\bf Time reading} from the TimeClock dataset~\citep{gpiosenka_time_image_datasetclassification} containing 11,520 training images of analogue clocks with ground truth times.
\end{enumerate}
In total, the curriculum contains 107,910 training samples, providing a comprehensive stress test for forgetting and knowledge transfer.

\paragraph{Held‑out suite.}
To measure generalization beyond the training stream, we evaluate on eight held‑out benchmarks:
AI2D \citep{kembhavi2016ai2d}, ChartQA \citep{masry2022chartqa}, DocVQA \citep{mathew2021docvqa}, InfoVQA \citep{mathew2022infovqa}, RealWorldQA \citep{visheratin_realworldqa_2024}, SeedBench \citep{li2023seedbench}, ScienceQA \citep{lu2022scienceqa}, and MMStar \citep{chen2024mmstar}.
InfoVQA and DocVQA use ANLS; since ANLS is in $[0,1]$, we average it with accuracies from the other held‑out tasks when reporting the mean held‑out score.

\subsection{On building target tasks}
Our training and evaluation scripts are built on the \href{https://github.com/LLaVA-VL/LLaVA-NeXT}{{\tt LLaVA-NeXT}} and \href{https://github.com/EvolvingLMMs-Lab/lmms-eval}{{\tt lmms-eval}} public GitHub repositories. Since some of the target tasks are not supported by {\tt lmms-eval}, we need to implement support for evaluation of target tasks. Details are as follows.

\paragraph{Bird classification.}
We reformulate the bird classification dataset CUB200~\citep{CUB} to a multiple choice VQA task following~\citep{liu2024revisit_llm}. Specifically, for a \textcolor{blue}{<image>} and \textcolor{blue}{<class name>} pair, we mix the correct label with 31 randomly chosen labels from the whole dataset and then compose a question like:
\begin{ColorVerbatim}
  \color{blue}{<image>} What species is the bird in this photo?
  Answer with the option’s letter from the given choices directly.
  \string\n \textcolor{red}{A.<class name A>} \string\n \textcolor{blue}{B.<class name>} \string\n ··· \color{red}{Z.<class name Z>}
\end{ColorVerbatim}
This task has 5,794 validation samples. The {\bf instruction prompt} for this task is: {\it ``Answer with the option's letter from the given choices directly.''} And only exact match can be deemed as correct by lowercasing model's output and compared to lowercased ground-truth answer.

\paragraph{Counting.}
The original \href{https://huggingface.co/datasets/allenai/pixmo-count}{PixmoCount} dataset provides download links rather than actual image files. By the time of downloading, not all links are valid. In the end, besides 36,140 training samples, we collected 535 and 536 validation and test samples. We use the validation set to report numbers in the paper, as done in the technical report of Pixmo dataset~\citep{deitkeMolmoPixMoOpen2024a}. The {\bf instruction prompt} for this task is {\it ``Answer with integer and nothing else. For example, if the answer is 1, you should output 1.''}. We convert the output by the model to digits and then use exact match to compute accuracy.

\paragraph{Medical VQA.}
We use the test split of the \href{https://huggingface.co/datasets/flaviagiammarino/path-vqa}{PathVQA dataset} for evaluation, containing 6,719 samples. The {\bf instruction prompt} for this task is {\it ``For questions that can be answered with a yes or no, just answer yes or no. Otherwise, provide an answer in the medical domain.''} We use the exact match score as the metric for this task, using the \href{https://github.com/KaveeshaSilva/PathVQA/tree/main}{official evaluation algorithm}.

\paragraph{OCR reading.}
\href{https://github.com/EvolvingLMMs-Lab/lmms-eval}{{\tt lmms-eval}} has support for TextVQA evaluation and we use the accuracy on the validation set as the performance for this task.

\paragraph{Time reading.}
We evaluate on the validation split of the
\href{https://www.kaggle.com/datasets/gpiosenka/time-image-datasetclassification}{TimeClock} dataset, which contains 1,440 samples.
The instruction prompt is
\emph{``Answer with the time in HH:MM format. For example, if it is 3:45, output 3:45.''}
To compute accuracy, we parse the model’s output to extract the hour and minute.
A prediction is marked correct only when both values match the ground truth.

\subsection{Task curriculum}
We provide three task sequences for sequential-tuning:
\begin{enumerate}
  \item CUB200 $\rightarrow$ PixmoCount $\rightarrow$ PathVQA $\rightarrow$ TextVQA $\rightarrow$ TimeClock
  \item PathVQA $\rightarrow$ CUB200 $\rightarrow$ TextVQA $\rightarrow$ TimeClock $\rightarrow$ PixmoCount
  \item TimeClock $\rightarrow$ TextVQA $\rightarrow$ PathVQA $\rightarrow$ PixmoCount $\rightarrow$ CUB200
\end{enumerate}
Unless otherwise stated, we use the first as the default sequential-tuning sequence.

\section{Implementation details}
\label{sec:impl}

\subsection{Implementation details for \ov}

We adopt the \href{https://huggingface.co/lmms-lab/llava-onevision-qwen2-7b-ov}{7B Qwen2 language model checkpoint} for experiments on \ov. Experiments run primarily on 4×NVIDIA H100 GPUs with 1 sample per GPU.
We use 8 gradient‑accumulation steps (effective batch 32), a learning rate of $5\!\times\!10^{-6}$ with cosine decay, and a 3\% warm‑up. Following the practice from~\citep{li2024llavanext-ablations}, we use a smaller learning rate at $2\!\times\!10^{-6}$ when tuning the vision encoder and the projector.
Evaluation uses \texttt{lmms‑eval} \citep{Zhang_LMMs-Eval_2024} with added support for our targets.
When we activate LwF, we set $\lambda\!=\!1$ and $\tau=2$. For all the experiments, we perform single epoch training.

\subsection{Parameter count for \ov}

\begin{table}[t]
  \centering
  \small
  \begin{threeparttable}
    \caption{Parameter groups and counts for LLaVA OneVision Qwen2-7B}
    \label{tab:param-groups-llavaov7b}
    \setlength{\tabcolsep}{6pt}
    \renewcommand{\arraystretch}{1.05}
    \begin{tabularx}{\linewidth}{@{}l >{\raggedright\arraybackslash}X r@{}}
      \toprule
      \textbf{Group} & \textbf{Components} & \textbf{\#Params} \\
      \midrule
      $\Theta_{\mathrm{VE}}$ & SigLIP vision encoder & $\approx 400\,$M \\
      $\Theta_{\mathrm{Proj}}$ & Multimodal projector & $\approx 20\,$M \\
      \midrule
      $\Theta_{\mathrm{SA}}$ & All blocks: $W_Q,W_K,W_V,W_O$ & $\approx 822\,$M \\
      $\Theta_{\mathrm{MLP}}$ & All blocks: $W_{\mathrm{gate}},\,W_{\mathrm{up}},\,W_{\mathrm{down}}$ & $\approx 5{,}703\,$M \\
      \midrule
      $\Theta_{\mathrm{Emb}}$ & Input token embeddings & $\approx 545\,$M \\
      $\Theta_{\mathrm{LM}}$ & LM head $U$ & $\approx 545\,$M \\
      \bottomrule
    \end{tabularx}
    \begin{tablenotes}[flushleft]
      \footnotesize
    \item[a] SigLIP So400M vision backbone, about $400$M parameters.
    \item[b] OneVision Stage 1 projector is about $20$M parameters for the 7B class.
    \item[c] Per layer counts with $d{=}3584$, $d_{kv}{=}512$, $L{=}28$:
      SA $29{,}364{,}736$, MLP up $135{,}790{,}592$, MLP $203{,}685{,}888$; totals multiply by $L$.
    \item[d] Vocab size $152{,}128$ and width $d{=}3584$ give $152{,}128\times 3584 = 545{,}226{,}752$ parameters for embeddings and for the LM head (untied).
    \end{tablenotes}
  \end{threeparttable}
\end{table}

Tab.~\ref{tab:param-groups-llavaov7b} lists the parameter groups and counts of each part in the \ov\ model. It can be seen that the language model takes a large part of the total capacity of the model. Within the language model, MLP is the major consumer of parameters.

\subsection{Implementation details for Qwen2.5-VL and LLaVA-NeXT (LLaMA 3)}
We adopt the \href{https://huggingface.co/Qwen/Qwen2.5-VL-7B-Instruct}{7B Qwen2.5 checkpoint} for experiments on Qwen2.5-VL. For all experiments on Qwen2.5-VL, we use 4 H100 GPUs and set learning rate at 2e-5 for all components in the model. Per-GPU batch size is set to 4, with 4 gradient accumulation steps. Therefore, the effective batch size is 64.

For experiments on LLaVA-NeXT (LLaMA-3), we use the \href{https://huggingface.co/lmms-lab/llama3-llava-next-8b}{8B LLaMA-3 checkpoint}. We adopt the same learning rate, warm-up ratio, batch size, gradient accumulation steps as tuning \ov.

For all the relevant experiments, we perform single epoch training.

\subsection{Implementation details for Sec.~\mainref{sec:dist-shift}: Output‑distribution probe (counting bias)}
\label{sec:counting-bias}


\paragraph{Setup.}
Fix a token subset $C$ (digits and common spelled numerals; exact list in the repo) and a held‑out batch $\hat{\mathcal{B}}=\{(I,y)\}$ of $|\hat{\mathcal{B}}|=100$ image–caption pairs sampled once from LCS‑558K (reused for all checkpoints and methods). For each checkpoint $s$ and each $(I,y)\in\hat{\mathcal{B}}$, generate a caption $\hat{y}$ with deterministic greedy decoding using identical preprocessing and decoding settings across methods.

\paragraph{Per‑position score.}
At generation step $j$, before committing the token, read the model’s next‑token probabilities and compute the per‑position number tendency
\[
  p^{\max}_C(\theta_s; I,\hat{y},j) \;=\; \max_{v\in C}\; p_{\theta_s}\!\bigl(v \,\big|\, \hat{y}_{<j},\, x_{\mathrm{vis}}\bigr).
\]

\paragraph{Per‑example and batch aggregates.}
Summarize each example by the sequence average
\[
  \mathrm{SeqAvg}_C(\theta_s; I)\;=\;\frac{1}{|\hat{y}|}\sum_{j=1}^{|\hat{y}|} p^{\max}_C(\theta_s; I,\hat{y},j),
\]
and aggregate over the batch to obtain the \emph{number‑token bias} at checkpoint $s$:
\[
  \mathrm{NTB}_s \;=\; \frac{1}{|\hat{\mathcal{B}}|}\sum_{(I,y)\in\hat{\mathcal{B}}}\mathrm{SeqAvg}_C(\theta_s; I).
\]

\paragraph{Correlation with held-out performance.} To quantify the relationship between output‑distribution drift and forgetting, we compute Spearman’s rank correlation between the average held‑out score and $\mathrm{NTB}_s$ across checkpoints and methods in Fig.~\mainref{fig:drift_on_counting}. We find a strong negative correlation ($\rho{=}{-}0.84$): checkpoints with higher numeric‑token bias systematically have worse held‑out performance.



\subsection{Numerical token list used for counting bias probe}
\begin{table}[t]
  \centering
  \footnotesize
  \setlength{\tabcolsep}{6pt} 
  \caption{Numeric token indices and their corresponding tokens.}
  \begin{tabular}{r l | r l | r l | r l}
    \toprule
    \multicolumn{2}{c}{Index / Token} & \multicolumn{2}{c}{Index / Token}
    & \multicolumn{2}{c}{Index / Token} & \multicolumn{2}{c}{Index / Token} \\
    \midrule
    15 & 0      & 16 & 1      & 17 & 2      & 18 & 3 \\
    19 & 4      & 20 & 5      & 21 & 6      & 22 & 7 \\
    23 & 8      & 24 & 9      & 603 & one   & 1960 & ten \\
    3966 & One  & 5225 & ONE   & 11613 & Two & 14154 & zero \\
    17999 & Zero & 19641 & Three & 19789 & two & 26972 & Four \\
    27856 & three & 32687 & Ten & 34024 & four & 37020 & Five \\
    41460 & Six  & 50364 & six  & 52670 & five & 58313 & million \\
    59085 & Eight & 59528 & Seven & 67532 & eight & 73956 & ZERO \\
    75796 & Twenty & 80185 & seven & 83329 & Nine & 91602 & Thirty \\
    93223 & nine & 93965 & twenty & \multicolumn{2}{c}{} & \multicolumn{2}{c}{} \\
    \bottomrule
  \end{tabular}
  \label{tab:numeric_tokens}
\end{table}

In Tab.~\ref{tab:numeric_tokens}, we list the total 38 numeric tokens we used for counting bias probe, and their indices in the tokenizer. The numeric tokens include numeric digits and words such as "one", "ONE", etc.

\section{Discovery Process}
\label{sec:discussion_claims}

In science, the ordering of observation, hypotheses, experimental results, and explanation is important --- to know whether claims are post-hoc rationalization of results or experiments are a confirmation of hypotheses that were based on prior observations. Therefore, we wish to give a full accounting.

In beginning this research, we first sought to verify the problem of ``catastrophic forgetting'' in LMMs. While prior works had largely confirmed the forgetting, these works tending to involve a limited range of tasks, so we created a diverse set of target tasks, some of which we expected to be very hard for the LLaVA-OneVision model (e.g. counting and telling time), and others to be easy (e.g. bird identification and TextVQA). After confirming that typical tuning practices, such as tuning the vision component, LLM, or full model, led to substantial forgetting, we thought we would turn to mitigation strategies, such as experience replay, model expansion with mixture-of-experts, knowledge distillation, and weight averaging. We also noticed a surprising result, that the model performance would drop significantly in held out benchmarks after training on the counting task, it would mostly recover on PathVQA, another specialized task that is not well represented in the benchmarks. Meanwhile, while performing the forgetting mitigation experiments, we also tried separately tuning only the self-attention projection (SA Proj) or MLP layers, motivated by the finding that tuning only the LLM was generally better than tuning the full model. This led to another very surprising result -- that tuning only self-attention projection layers led to very good learning of the target tasks with no drop in performance in held out tasks, even after training all five target tasks in a sequence. This was surprising because we were not aware of other instances of strong learning without forgetting behavior, in the absence of model expansion, rehearsal, or strong regularization.  A third interesting result was that knowledge distillation turned out to be the most effective method for mitigating forgetting that we tried, outperforming e.g. replay of examples from earlier target tasks and a mixture-of-experts scheme for model expansion.

Initially, we sought to stress test these results. Indeed, we found that if we keep training new tasks, such as the large long-tailed task of iNaturalist~\citep{van_horn_2023_inaturalist} classification, we see a little bit of forgetting.  Also, in tuning other models, LLaVA-NeXT and Qwen2.5-VL (Table~\mainref{tab:llama3_component_summary}),  we do not see exactly the same numbers, of course, but the major trends hold.  With Qwen2.5-VL, we actually get a little forward transfer on the held-out tasks when tuning SA Proj.  With LLaVA-Next, we get a 7.7 point drop in held out tasks, but less than half the forgetting of tuning MLP layers and less than one-quarter as much as tuning the full LLM.  We also tried other sequences of target task training and found that the post-counting recovery of forgetting was not a fluke. For example, we see recovery from both PixmoCount and TimeClock when reversing the sequence order. By-and-large, the results held --- fine-tuning self-attention is remarkably robust to forgetting, what was ``forgotten'' can be recovered without rehearsal, and regularizing the outputs with knowledge distillation also is highly effective in mitigating forgetting when tuning the MLP.

We performed many other experiments, but our breakthrough in understanding came from reviewing the literature, particularly in work, such as Geva et al.~\citep{Geva_TransformerFeedForwardLayersAreKeyValueMemories_2021} and Olsson et al.~\citep{Olsson_InductionHeads_2022}, that experimentally explore the roles of transformer components.  Their key results are that MLPs are responsible for storing and applying memories, with the up layer(s) looking up the memories (or activating concepts) and the down layers applying the activated concepts to modify the output token distribution.  Attention, on the other hand, is responsible for processing and organizing the inputs. This led us to consider that a model can adapt to a task in many ways: acquiring skills to make better use of its inputs, acquiring new memories and concepts, better applying those concepts, or simply biasing toward the output distribution.  We hypothesized that, when training the full LLM, the model is at least partially taking a shortcut to bias toward the output distribution, rather than focusing on skill or memory improvement. This hypothesis could explain all three observed phenomena.  The SA Proj is robust to forgetting because it does not directly tune the MLP layers that produce the output distribution.  The forgetting is sometimes recoverable because subsequent training on a task with more varied outputs reverses the narrow output distribution shift.  Knowledge distillation directly penalizes shift in the output distribution.

This led us to propose two experiments to test this hypothesis.  First, we reasoned, we should see that, as the counting task is trained, the model becomes more predisposed to output numbers, since the counting task answers are always of the form, ``There are [number] [object(s)] in this image.''  We also should see some correlation between this bias toward numeric tokens and forgetting in held-out benchmarks.  As we show in Fig.~\mainref{fig:drift_on_counting} and exemplify in Fig.~\mainref{fig:counting_caption_viz}, the results are quite striking with a strong effect of the output distribution bias and a strong correlation with forgetting.  Second, we proposed, tuning the MLP except for the down layers that most directly modify the output distribution should mitigate the output bias and, therefore, reduce forgetting.  Again, the confirmation was strong --- with LLaVA-OneVision, tuning only MLP up layers achieved the best overall target performance with only a little more forgetting than tuning SA Proj. We believe our results are conclusive, especially given the observe-hypothesize-test-confirm pattern of our research.

Finally, we would like to stress that our experiments have been more thorough than we can relate in the main text. Just in generating the results of Table~\mainref{tab:sequential_effects_component_summary}, we fine-tuned the 7B parameter model on 5 tasks 21 times (3 sequences, 7 components) and evaluated 8 broad benchmarks and 5 target tasks 105 times (after each target task was trained).  That is 105 task trainings and 1365 task evaluations.  We include many other experimental results in the main text and appendix below.  While it is always possible to train more models, more component variations, more mitigation strategies, on more datasets and with more evaluations, we have pushed our resources to the brink and hope that the reader finds our claims sufficiently supported, as we do.





\section{More results}

\subsection{Single task fine-tuning}
\label{sec:single_task}

\begin{table*}[h]
\setlength\tabcolsep{5pt}
\centering
\caption{\textbf{Single task fine-tuning by component.} Each individual target task is fine-tuned from the original model, and the performance for that task (``Target'') and average held-out performance (``Held-out'') is measured.  Each row is for tuning a different component or set of components:
``Proj.'', ``Vis. Enc.'', ``SA Proj.'',  ``MLP'', ``LM'', ``Full'' represent tuning the projector, vision encoder, self-attention projections in the LLM, MLP in LLM, full LLM, and all parameters, respectively. ``+'' is an increment over the baseline (original \ov-7B checkpoint), and ``-'' is a decrease.  
}
\resizebox{1\textwidth}{!}{
    \begin{tabular}{@{}l | cc | cc | cc | cc | cc | cc}
    \toprule
    
   \multirow{3}{*}{\textbf{Method}} & \multicolumn{2}{c|}{\textbf{CUB200}}  & \multicolumn{2}{c|}{\textbf{PixmoCount}}  & \multicolumn{2}{c|}{\textbf{PathVQA}}  & \multicolumn{2}{c|}{\textbf{TextVQA}}  & \multicolumn{2}{c|}{\textbf{TimeClock}} & \multicolumn{2}{c}{\textbf{Average}}\\
    
    \cmidrule(lr){2-3}  \cmidrule(lr){4-5}  \cmidrule(lr){6-7}  \cmidrule(lr){8-9} \cmidrule(lr){10-11}  \cmidrule{12-13}
    
    &  {Target} & {Held-out} &  {Target} & {Held-out} &  {Target} & {Held-out} &  {Target} & {Held-out} &  {Target} & {Held-out} & {Target} & {Held-out}
    \\ 
    
    \midrule
    Baseline & 53.7 & 76.4 & 52.4 & 76.4 & 36.3 & 76.4 & 76.0 & 76.4 & 1.1 & 76.4 & 43.9 &  76.4\\ 
    \midrule
    Proj. & +5.7 & -0.0 & +4.2 & -0.1 & +0.6 & -0.1 & +0.5 & +0.2 & +0.4 & -0.5 & +2.3 & -0.1 \\
    Vis. Enc. & +16.1 & -0.8 & +11.6 & -4.7 & +3.7 & -2.8 & +1.0 & -0.7 & +12.7 & -11.9 & +9.0 & -4.2 \\
    SA Proj. & +31.8 & +0.3 & +15.2 & -0.2 & +14.4 & -0.3 & +3.5 & +0.3 & +56.0 & -0.1 & +24.2 & +0.0  \\
    MLP & +36.4 & +0.3 & +17.8 & -4.0 & +26.5 & -0.4 & +3.8 & +0.0 & +73.3 & -3.1 & +31.6 & -1.4  \\
    LM & +40.0 & -0.0 & +16.3 & -7.7 & +26.8& -0.7 & +3.5 & -0.7 & +72.6 & -4.6 & +31.8 & -2.8 \\
    Full & +37.0 & +0.1 & +19.0 & -9.0 & +27.4& -0.9&+3.4 & -0.7 & +79.8 & -5.4 & +33.3 & -3.2\\
    \bottomrule
    \end{tabular}
}
\label{tab:individual_test}
\end{table*}

In Fig.~\mainref{fig:teaser_new_new}, we show the learning and forgetting of tuning different components on one target task at a time, and then recording the performance for that target task and the average held-out performance. In Tab.~\ref{tab:individual_test} we show the actual performance of each component by taking the delta based on the baseline (original model), ordered from least to most parameters.


As a general trend, tuning more parameters increases both {\em learning} (improvement in target task) and {\em forgetting} (decrease in average held-out tasks), with vision encoder and self-attention projection as the notable exceptions.
Tuning the {\bf full} network or only the {\bf language model} yields the greatest learning (+33.3 and +31.8 percentage points, on average), yet these gains are accompanied by significant forgetting (-3.2 and -2.8 points). Adjusting the {\bf MLP} of the LLM provides a good trade-off, with similar learning (+31.6) and substantially lower forgetting (-1.4).
Adjusting only the {\bf self-attention projection layers} achieves a respectable +24.2 in learning and, surprisingly, {\em no measurable forgetting}.  The {\bf vision encoder} and the {\bf projector} offer relatively little gain, and the vision encoder has the most forgetting (-4.2), indicating that tuning the vision features is particularly disruptive.

Now, consider the variations by task.  There is only a weak correlation between the amount learned and forgotten per task.  For instance, CUB200 has the second-most learning (after TimeClock) but the least forgetting. Also, some tasks benefit from visual tuning while others do not. Fine-grained bird recognition (CUB200) and medical question answering (PathVQA) benefit almost exclusively from language model updates, gaining +40.0 and +26.8 points, respectively, with little or no benefit from additional vision tuning.
Conversely, for PixmoCount and TimeClock, tuning the full model handily outperforms tuning only the LLM portion.

\subsection{Sequential fine‑tuning}
\label{sec:app_sequential_tasks}

\begin{figure*}[htbp]
  \centering
  \includegraphics[width=\textwidth]{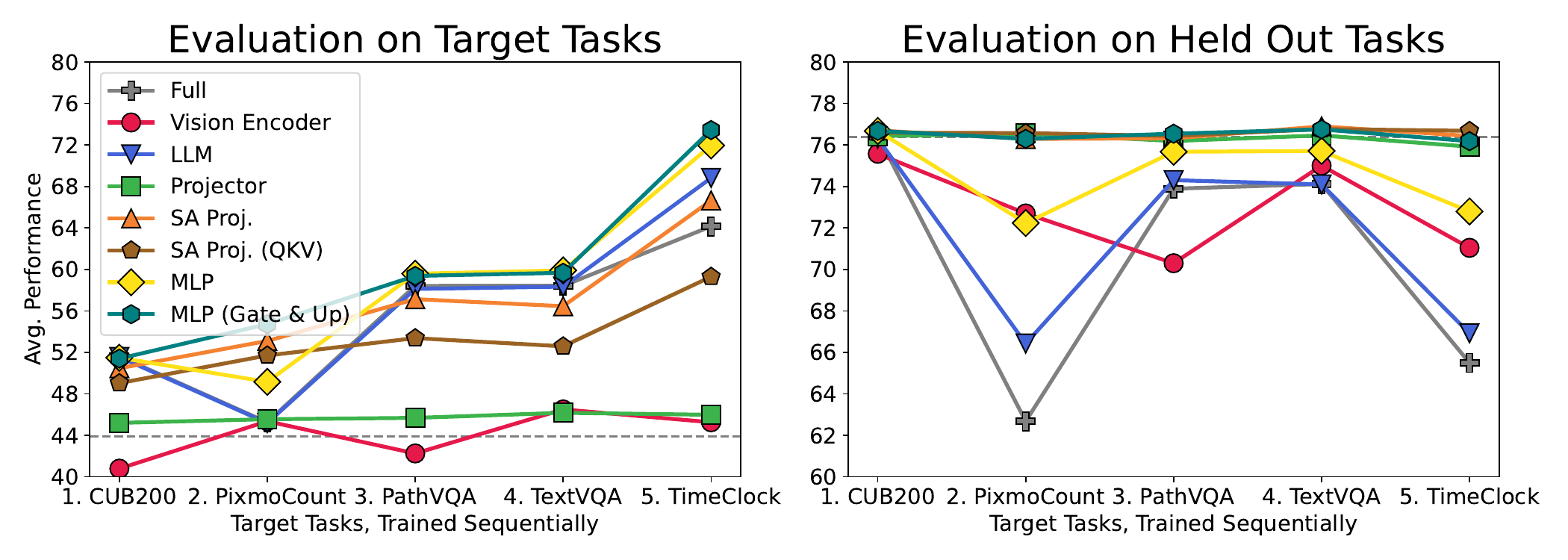}
  \caption{\textbf{Sequential fine-tuning by component.} The target tasks in the $x$-axis are trained sequentially, from left to right.  After training each task, the average performance of all target tasks (\textbf{left}) and all held-out tasks (\textbf{right}) are measured. Each line shows the performance after tuning a different component or set of components: \ov\ (Full, Vision Encoder, Projector, LLM, SA Proj., SA Proj. (QKV),, MLP, MLP (Gate\&Up)). The dashed horizontal \textcolor{gray}{gray} line marks the average 
  performance of the original model.
  }
  \label{fig:tunable_parts}
\end{figure*}

In Fig.~\ref{fig:tunable_parts}, we display how sequentially tuning different components on the default sequence of all five tasks affects the {\em average performance of all target tasks} and held-out tasks. In this case, forgetting in later learning can affect the performance of target tasks learned earlier. Per-task performance is attached in the later sections.

Updating the MLP (Gate\&Up) gives the best target-task performance overall. Multiple methods have stable results on held-out tasks through out the whole sequence, such as SA Proj., SA Proj. (QKV), and MLP (Gate\&Up).

Another interesting phenomenon which is also mentioned in the main paper is that held-out performance does not continually drop as more tasks are trained, but rises and falls.  For example, training on PixmoCount causes substantial loss in held out performance (0.76 to 0.63 for the full model), but the loss is largely recovered  (to 0.74) by training the next task PathVQA.  This means in ``forgetting'', much of the information is not permanently lost but temporarily inaccessible.

\begin{figure*}[t]
  \centering
  \includegraphics[width=\textwidth]{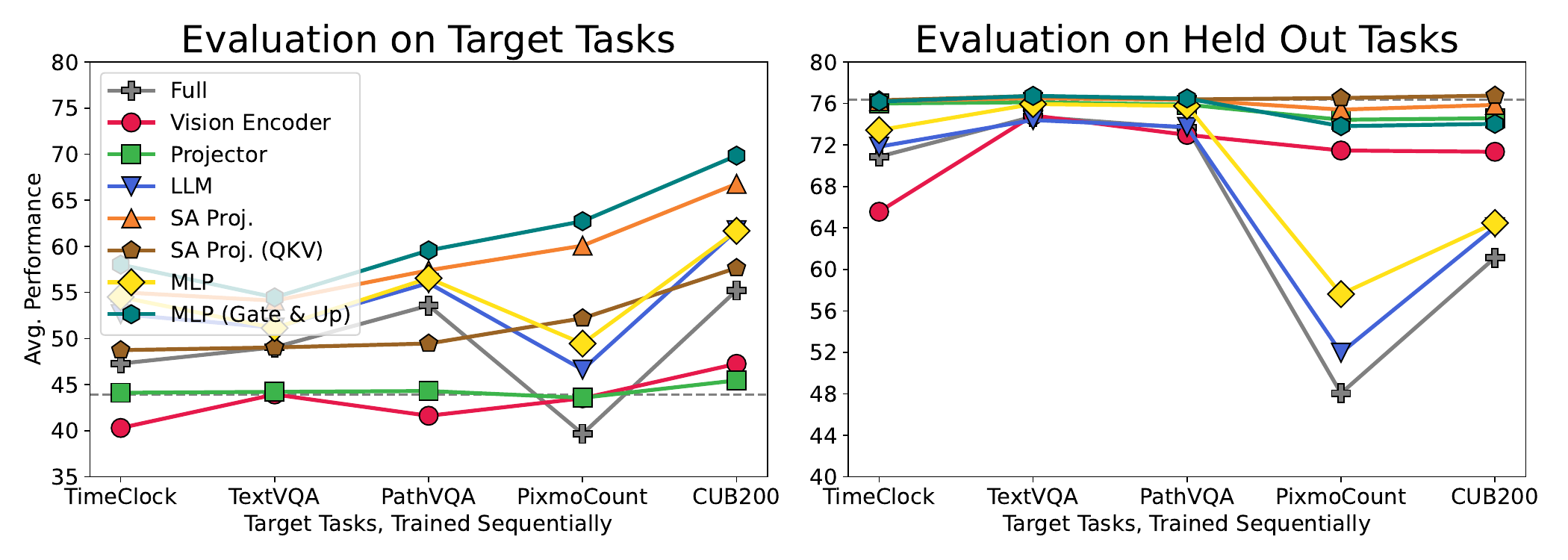}
  \caption{\textbf{Sequential fine-tuning by component.} Tasks are arranged as TimeClock $\rightarrow$ TextVQA $\rightarrow$ PathVQA $\rightarrow$ PixmoCount $\rightarrow$ CUB200.} 
  \label{fig:tunable_parts_tcorder}
\end{figure*}

\begin{figure*}[t]
  \centering
  \includegraphics[width=\textwidth]{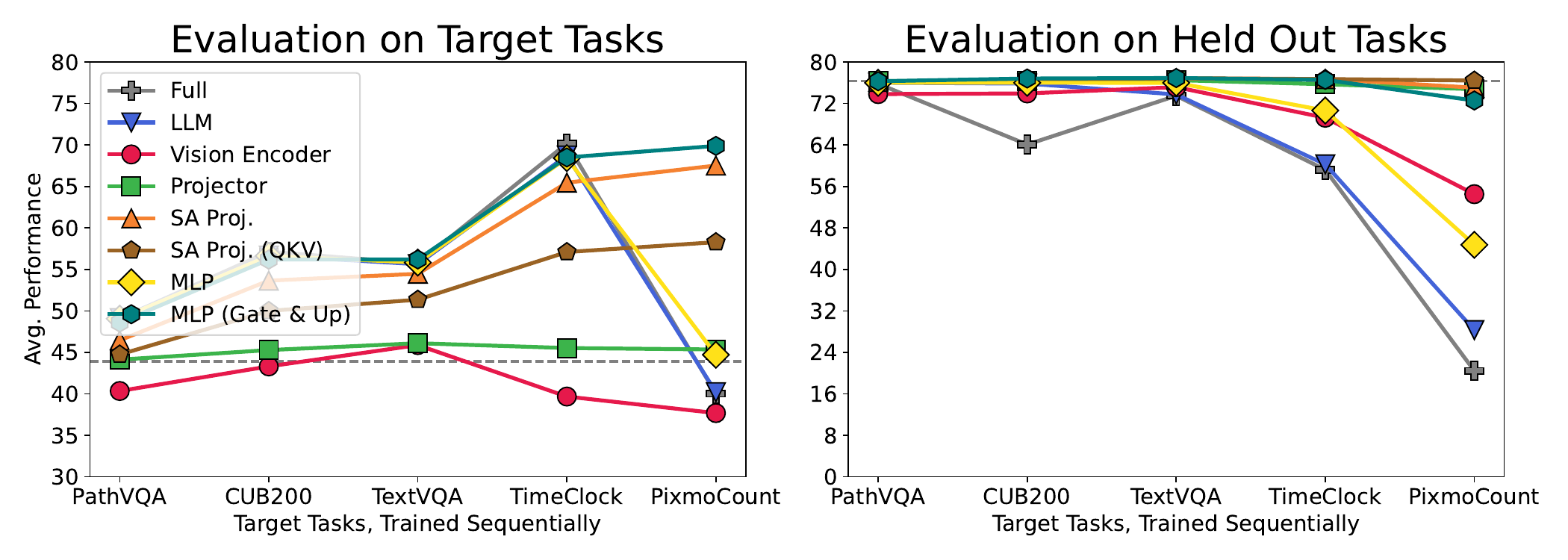}
  \caption{\textbf{Sequential fine-tuning by component.} Tasks are arranged as PathVQA $\rightarrow$ CUB200 $\rightarrow$ TextVQA $\rightarrow$ TimeClock $\rightarrow$ PixmoCount.} 
  \label{fig:tunable_parts_altorder}
\end{figure*}

In Figs.~\ref{fig:tunable_parts_altorder} and~\ref{fig:tunable_parts_tcorder}, we show the sequential tuning results by component on \ov\ in the other two orders. Fig.~\ref{fig:tunable_parts_tcorder} validates that forgetting recovery is not order-specific: methods that forget significantly on PixmoCount, rebound after tuning on CUB200. Both figures indicate the robustness of SA Proj., SA Proj. (QKV), and MLP (Gate\&Up) on held-out tasks as they essentially keep flat throughout. Especially, MLP (Gate\&Up) has a huge benefit in target learning.

\section{More analysis}



\subsection{Composing stable tuning strategies: SA Proj.\ + MLP Gate\&Up}
\label{sec:appendix-composition}

\begin{figure*}[t]
  \centering
  \includegraphics[width=\textwidth]{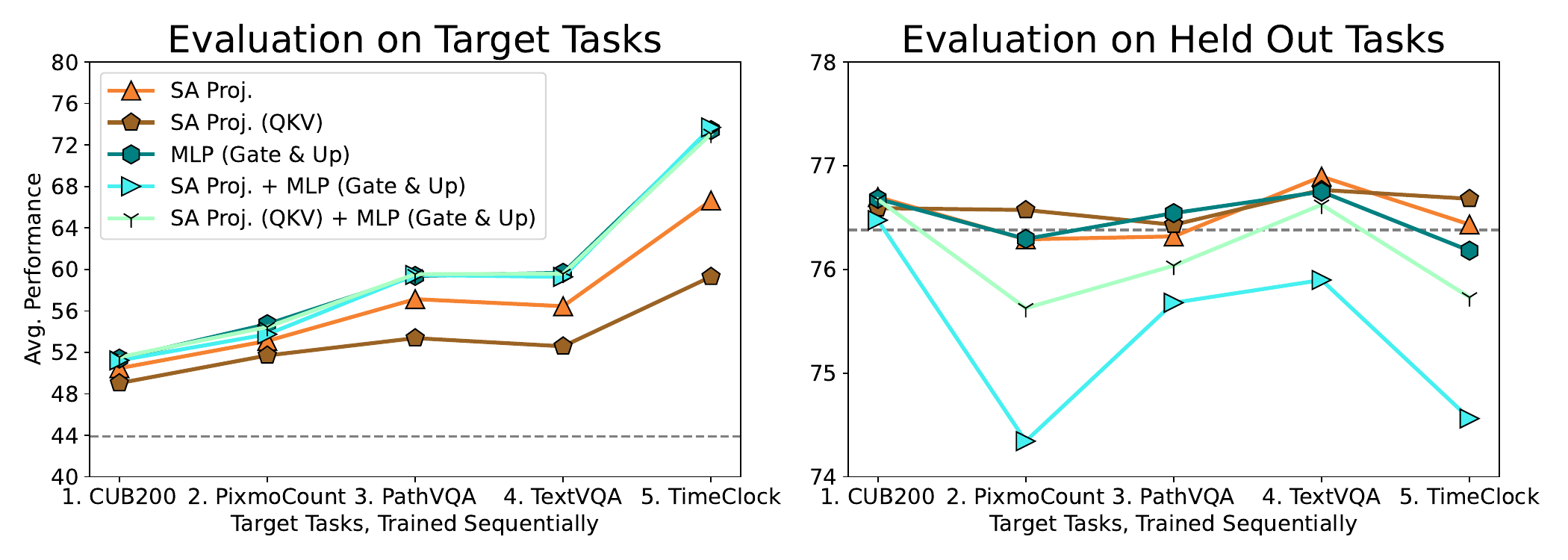}
  \caption{\small \textbf{Composing stable updates.} We compare SA Proj.\ and MLP (Gate\&Up) to two compositions: \textbf{SA Proj.\ + Gate\&Up} and \textbf{SA Proj.\,(QKV only) + Gate\&Up} using the default five-task sequential tuning.}
  \label{fig:composition_comparison}
\end{figure*}

We asked whether combining the two most stable, high‑learning settings from the main paper, i.e., SA Proj. and MLP (Gate\&Up), has further benefits. We evaluate two compositions: \textbf{SA Proj.\ + MLP (Gate\&Up)} and \textbf{SA Proj.\,(QKV only) + MLP (Gate\&Up)} under the same five‑task curriculum, reporting the same sequence‑level metrics and counting‑bias probe. In aggregate, the composed variants match or slightly underperform the two standalone settings on target learning while keeping held‑out changes small; the QKV‑only composition is better than the other composition in the held-out performance. But across tasks and checkpoints, neither composition consistently dominates MLP (Gate\&Up) alone, indicating that most of the achievable learning–stability trade‑off is already realized by Gate\&Up for \ov.



\subsection{Layer‑wise residual–to–logit contribution analysis}
\label{sec:layer-contrib-setup}

\begin{figure*}[htbp]
  \centering
  \includegraphics[width=\textwidth]{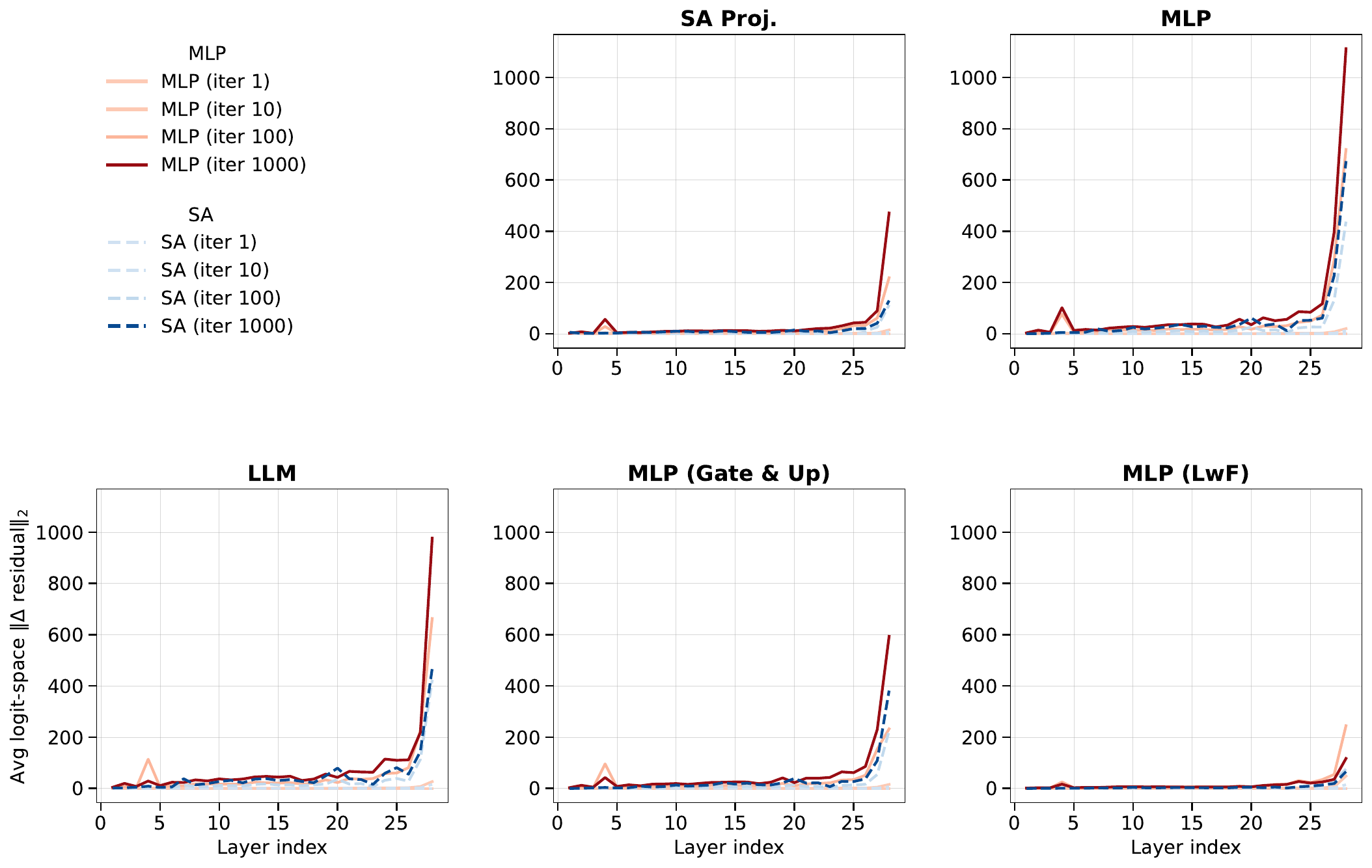}
  \caption{Layer-wise residual-delta magnitude across training iterations. Each subplot shows the average logit-space $\lVert \Delta \text{residual} \rVert_2$ attributable to MLP (solid reds) and self-attention projections (dashed blues) versus transformer layer index. The five method configurations are: SA Proj., MLP, LLM, MLP (Gate \& Up), and MLP (LwF). Color shade encodes checkpoint iteration (darker = later). The top-left panel is a legend; other panels omit legends for clarity.}
  \label{fig:layer_contribution_to_shift}
\end{figure*}

We quantify where (by depth) and how strongly (by pathway) adaptation perturbs the output distribution by comparing the \emph{logit‑space} effects of self‑attention versus MLP residual updates across layers and training steps. We evaluate on a fixed, held‑out multimodal shard sampled once from LCS‑558K and reuse it for all methods and checkpoints; unless noted, statistics are computed under teacher forcing over the assistant answer span (target tokens). For each tuning configuration (SA Proj., MLP, LLM, MLP (Gate\&Up), and MLP (LwF)) we compare tuned checkpoints to the frozen stage‑0 base model at log‑spaced training steps (e.g., \(1,10,100,1000\)), excluding the combined SA Proj.\,+\,MLP (Gate\&Up) condition. We register forward hooks on every decoder layer’s self‑attention and MLP submodules in both the base and tuned models to capture their residual increments \(a^{(l)}\) and \(f^{(l)}\) at each token \(j\); with the LM head \(U\) fixed, we form logit‑space deltas by projecting the \emph{difference} of residual contributions through \(U\):
\[
  \Delta z_{\text{SA}}^{(l)}(j)=U^{\!\top}\!\big(a^{(l)}_{\text{tuned}}(j)-a^{(l)}_{\text{base}}(j)\big),\qquad
  \Delta z_{\text{MLP}}^{(l)}(j)=U^{\!\top}\!\big(f^{(l)}_{\text{tuned}}(j)-f^{(l)}_{\text{base}}(j)\big).
\]
For each layer we aggregate token‑wise vectors into a scalar via the \(\ell_2\) norm and then average across tokens and examples to obtain per‑layer logit‑space magnitudes:
\[
  \text{SA}(l)=\sqrt{\mathbb{E}_{j}\!\big[\|\Delta z_{\text{SA}}^{(l)}(j)\|_2^2\big]},\qquad
  \text{MLP}(l)=\sqrt{\mathbb{E}_{j}\!\big[\|\Delta z_{\text{MLP}}^{(l)}(j)\|_2^2\big]}.
\]
We report these two curves (dashed blue for self‑attention, solid red for MLP) per checkpoint, sharing axes across panels for comparability.

\paragraph{Key observations.}
(1) \textbf{MLP dominates the shift.} Across configurations and steps, MLP curves exceed self‑attention curves—often by \(>\!\!2\times\) in later layers—indicating that most logit‑space change comes from the MLP pathway.
(2) \textbf{Drift grows with training.} For settings that forget (e.g., full LLM or full MLP), per‑layer magnitudes increase monotonically with checkpoint step, mirroring the counting‑bias rise and held‑out decline.
(3) \textbf{Late layers drive the effect.} The last 4–5 transformer layers account for the vast majority of the shift, with the final two layers contributing the largest deltas; early layers remain comparatively stable.
(4) \textbf{Regulating write‑back reduces drift.} MLP (Gate\&Up) and MLP (LwF) substantially shrink late‑layer MLP magnitudes relative to full MLP, aligning with their smaller held‑out drops.
(5) \textbf{Self‑attention changes are smaller and flatter.} SA Proj. curves are consistently below the corresponding MLP curves and vary less across steps, indicating weaker and less step‑sensitive contribution to the overall distribution shift.

\subsection{Effect of tuning different LLM layers}
\label{sec:llm-layer-split}
\begin{figure*}[htbp]
  \centering
  \includegraphics[width=\linewidth]{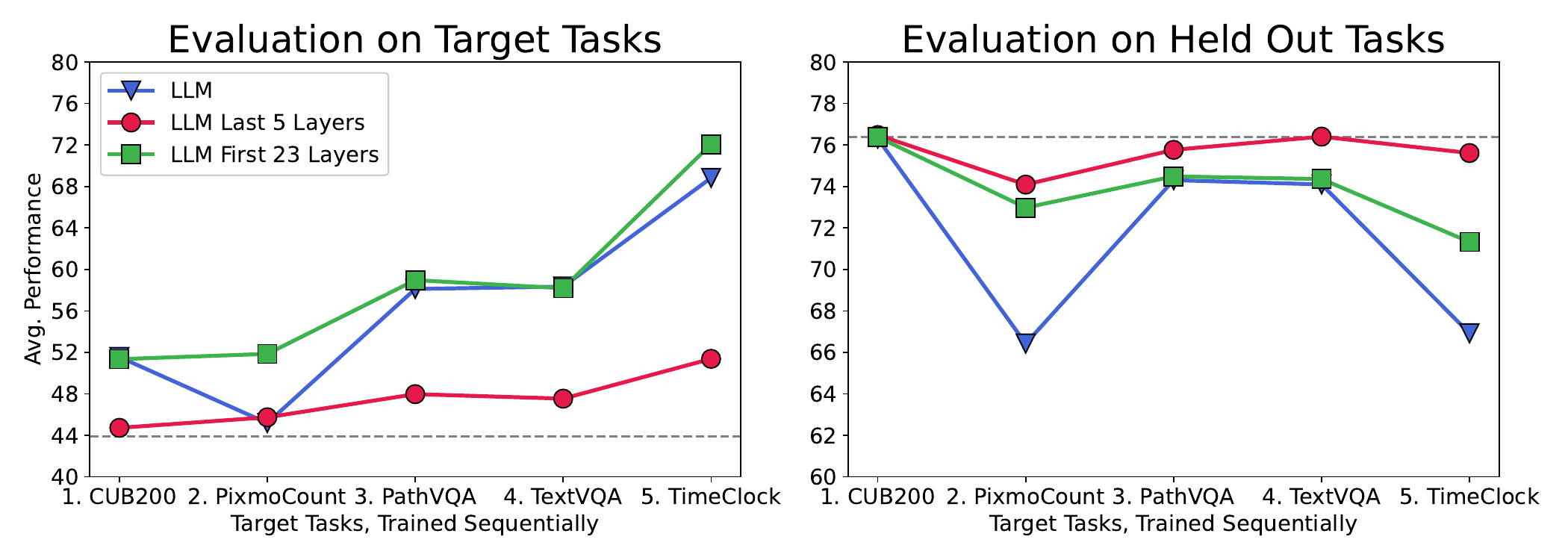} 
  \caption{\rebuttaledit{\textbf{Sequential tuning of LLM layer subsets.} We compare tuning the full LLM against tuning only the first 23 layers or only the last 5 layers of the 28-layer Qwen2-7B model.}}
  \label{fig:llm_layers_split}
\end{figure*}

\rebuttaledit{Guided by the observation that late transformer layers are the primary drivers of output distribution shift (Sec.~\ref{sec:layer-contrib-setup}), we investigate how different layer subsets contribute to learning versus forgetting. We partition the 28-layer Qwen2-7B language model into two subsets: the \textbf{First 23 Layers} (blocks 0–22) and the \textbf{Last 5 Layers} (blocks 23–27). We perform sequential fine-tuning on our default five-task curriculum, updating only one subset while freezing the rest.}

\rebuttaledit{We can read the following from Fig.~\ref{fig:llm_layers_split}:
  \begin{itemize}[leftmargin=*, itemsep=2pt]
    \item \textbf{Tuning only the \emph{Last 5 Layers} yields minimal learning gains.} This suggests the late layers alone lack the capacity to adapt to new tasks, and as a consequence, they also induce little forgetting.
    \item \textbf{Tuning only the \emph{First 23 Layers} drives strong learning,} achieving target task performance comparable to, and at times exceeding, tuning the full LLM. Because the \emph{Last 5 Layers} are frozen in this "First 23" experiment, the forgetting seen in the full LLM is largely alleviated. The sharp drop from the counting task (PixmoCount) is significantly mitigated.
  \end{itemize}}

\rebuttaledit{In summary, these results directly confirms our analysis that constraining the output distribution drift is an effective strategy to mitigate forgetting.}

\subsection{Qualitative results of MLP and SA Proj. tuning}


In Fig.~\mainref{fig:counting_caption_viz}, we demonstrate the response differences between tuning the MLP and tuning SA Proj. in the LLM on PixmoCount for 1K iterations. Hence, counting examples can be regarded as the target evaluation and we draw two image captioning samples from LCS‑558K~\citep{liu2023visualinstructiontuning} as a held-out evaluation. It can be seen that SA Proj. can both output the correct answers for counting examples and remain the capability to give detailed responses when asked to caption images. As a contrast, MLP uses the learned counting skill to describe the image contents, for example, "There are 2 photos in the photo." for the second row. It demonstrates that MLP temporarily forgets how to answer this question but still remains the capability to conceptualize image contents and recognize objects.

\subsection{Training efficiency comparison of different forgetting mitigation methods}

\begin{table}[t]
  \centering
  \caption{\rebuttaledit{Training efficiency and parameter footprint of the evaluated model variants
           (4 × H100 96 GB, DeepSpeed, bfloat16).
           \textbf{Total Params} is the total number of model parameters in billions.
           \textbf{Trainable Params} is the subset that requires gradient updates in billions.
           \textbf{Train SPS} is the number of training samples processed per second, collected from training models on the same 384 samples. }}
  \label{tab:train_efficiency}
  \setlength{\tabcolsep}{6pt}
  \footnotesize
  \begin{tabular}{lrrr}
    \toprule
    \textbf{Method} &
    \textbf{Total Params} &
    \textbf{Trainable Params} &
    \textbf{Train SPS} $\uparrow$ \\
    & (B) & (B) & (\#samples / sec.) \\
    \midrule
    SA Proj.             &  8.03 &  0.82 & 1.46 \\
    MLP (Gate\&Up)       &  8.03 &  3.80 & 1.45 \\
    MLP                  &  8.03 &  5.70 & 1.44 \\
    LoRA                 &  8.51 &  0.50 & 1.27 \\
    LwF                  & 16.06 &  5.70 & 0.81 \\
    MoE                  & 13.73 &  5.70 & 0.44 \\
    \bottomrule
  \end{tabular}
\end{table}

\rebuttaledit{We compare model size, number of trainable parameters, and training speed for the forgetting mitigation methods shown in Fig.~6.
  The results appear in Tab.~\ref{tab:train_efficiency}.
  All experiments were run on four NVIDIA H100\,96 GB GPUs with DeepSpeed in \texttt{bfloat16}, and no competing processes were active.
  Each run processed exactly 384 training samples.
  \textbf{Total Params} and \textbf{Trainable Params} are reported in billions.
\textbf{Train SPS} reports the average number of training samples processed per second, reflecting data loading and optimization steps.}

\rebuttaledit{We observe the following:
  \begin{itemize}
    \item \textbf{SA Proj.} achieves the highest throughput at 1.46 samples / sec. while using only 0.82 B trainable parameters.
    \item \textbf{MLP (Gate\&Up)} follows closely at 1.45 samples / sec. but requires larger number of trainable parameters (3.8 B).
    \item \textbf{MLP} uses more trainable parameters (5.7 B) though the train SPS is very close to the above two variants.
    \item \textbf{LoRA} uses the fewest trainable parameters at 0.5 B, though its throughput is lower at 1.27 samples / sec. due to processing every token through adapter weights.
    \item \textbf{LwF} maintains a teacher network, resulting in the largest total parameter count of 16.06 B and a reduced throughput of 0.81 samples / sec. because of the extra forward pass and distillation loss.
    \item \textbf{MoE} introduces two experts per MLP module and a layerwise gating network, inflating total parameters to 13.73 B and reducing throughput further to 0.44 samples / sec.
  \end{itemize}}

\subsection{Output Distribution Dynamics Across Forgetting and Recovery}
\label{sec:output-dist-dynamics}

\rebuttaledit{To understand the output distribution dynamics during both forgetting and subsequent recovery, we conducted an additional analysis. Building upon the methodology of our counting-bias probe (Sec.~\mainref{sec:dist-shift} and Sec.~\ref{sec:counting-bias}), we extended this analysis across three sequential stages of our curriculum: initial training on CUB200, followed by PixmoCount (where forgetting typically occurs), and finally PathVQA (where recovery is observed).}
\begin{figure*}[htbp]
  \centering
  \includegraphics[width=\textwidth]{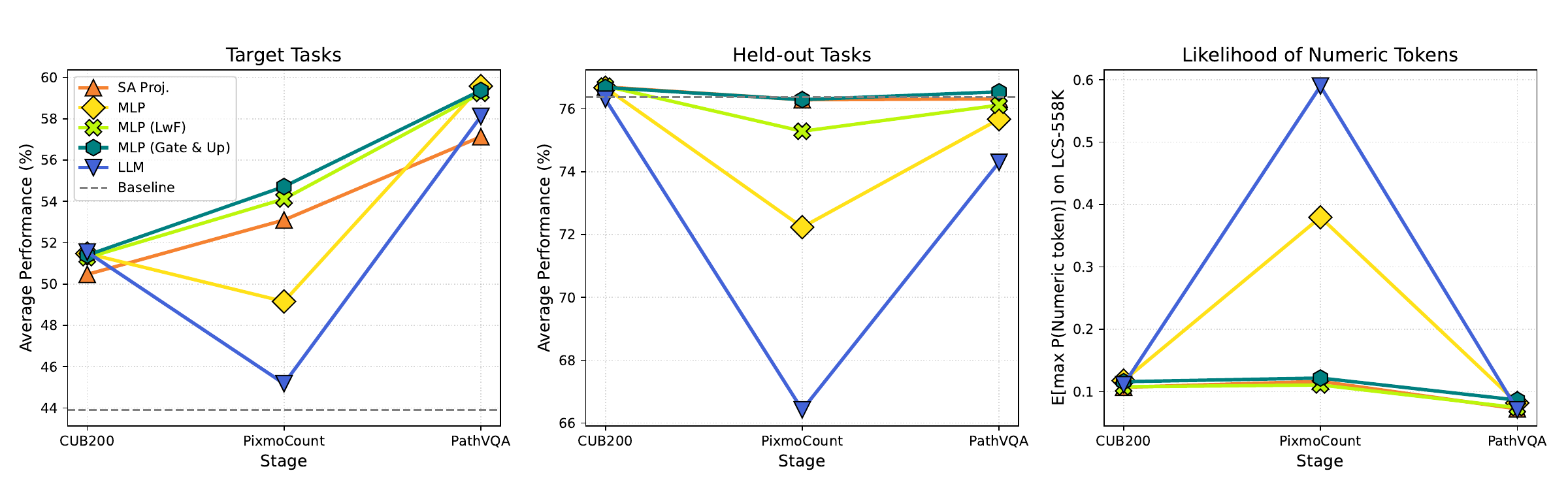}
  \caption{ \rebuttaledit{
    \textbf{Analysis of the forgetting and recovery cycle over three sequential tasks.} 
    We plot performance on \textbf{(Left)} average target tasks, \textbf{(Middle)} average held-out tasks, and \textbf{(Right)} the likelihood of numeric tokens on a general held-out dataset (LCS-558K). The plots track performance after training on CUB200 (Stage 1), PixmoCount (Stage 2), and PathVQA (Stage 3). 
    The key finding is in the rightmost plot: for methods susceptible to forgetting (e.g., LLM, MLP), the numeric token bias (output shift) \textbf{spikes} at Stage 2 (correlating with the performance drop) and then \textbf{recedes} at Stage 3 (correlating with the performance recovery). This provides direct quantitative evidence that the "recovery" phenomenon is a re-correction of the output distribution shift.}
  }
  \label{fig:recovery_probe}
\end{figure*} 

\rebuttaledit{The results are presented in Fig.~\ref{fig:recovery_probe}. The rightmost plot directly visualizes the output distribution shift for numeric tokens and its recovery for different fine-tuning methods. Comparing this with the middle plot for the held-out performance, we observe:}
\rebuttaledit{
  \begin{itemize}[leftmargin=*, itemsep=2pt]
    \item \textbf{Forgetting (PixmoCount):} For methods susceptible to forgetting (e.g., LLM Full, MLP), training on PixmoCount leads to a significant drop in held-out performance (middle plot) and a simultaneous spike in the likelihood of numeric tokens (right plot). This re-confirms our finding that forgetting is directly correlated with an output distribution shift towards the task-specific bias.
    \item \textbf{Recovery (PathVQA):} Crucially, after training on PathVQA, which requires diverse, non-numeric outputs, we observe a clear recovery in held-out performance for these same methods. This recovery is directly mirrored by a reduction in the likelihood of numeric tokens, bringing the output distribution closer to its pre-counting state.
  \end{itemize}}

\rebuttaledit{This extended probe explicitly demonstrates that the "recovery" of held-out performance is indeed a consequence of the model's output distribution shifting away from the previous task's bias and back towards a more general, un-biased state. This provides strong evidence for the mechanism of recoverable output distribution shift throughout the entire forgetting and recovery cycle. We note, however, that the held-out performance does not fully recover to its original baseline. We hypothesize that, this is likely because the PathVQA task, while more general than counting, still introduces its own subtle, task-specific distribution shifts on other tokens. This new bias prevents a perfect return to the original general-purpose state, even as it corrects the more damaging numeric bias from the previous task. Quantifying these more complex, overlapping distribution shifts remains a challenging direction for future work.}

\subsection{Qualitative Analysis of Forgetting and Recovery}
\label{sec:qualitative_recovery}
\begin{figure*}[ht]
  \centering
  \includegraphics[width=\textwidth]{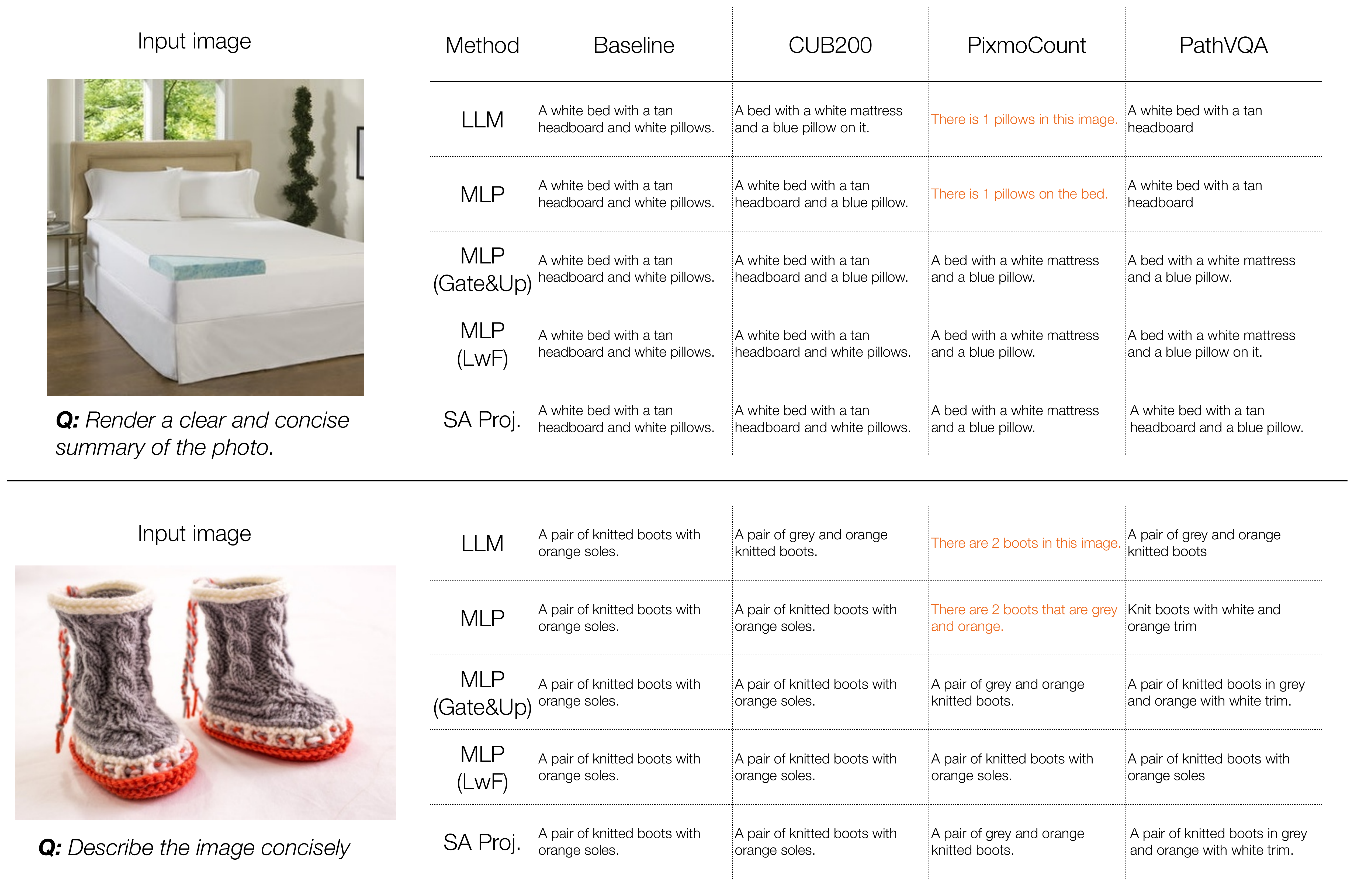}
  \caption{\rebuttaledit{Results of different models on captioning examples from the LCS‑558K~\citep{liu2023visualinstructiontuning} dataset after sequentially fine-tuning on CUB200, PixmoCount, and PathVQA. }}
  \label{fig:image_caption_vis_seq_tuning}
\end{figure*} 

\rebuttaledit{In addition to the quantitative probes, Figure~\ref{fig:image_caption_vis_seq_tuning} provides a direct qualitative visualization of the forgetting and recovery cycle. This figure displays model responses to general, held-out image captioning prompts (e.g., "Describe the image concisely") at key sequential stages: Baseline, after CUB200, after PixmoCount (the "forgetting" stage), and after PathVQA (the "recovery" stage).}

\rebuttaledit{The results clearly illustrate the output shift mechanism. We observe two key behaviors in the methods most prone to forgetting (\textbf{LLM} and \textbf{MLP}):
  \begin{itemize}[leftmargin=*, itemsep=2pt]
    \item \textbf{Forgetting (After PixmoCount):} After training on the narrow counting task, these models exhibit a strong output bias. They incorrectly reframe the general captioning request as a counting problem, producing outputs like "There are 2 boots in this image". This qualitatively demonstrates the "forgetting" as a task-specific output shift, not a loss of core concepts (the model still identifies "boots").
    \item \textbf{Recovery (After PathVQA):} Crucially, in the final column, after training on the more diverse PathVQA task, these same models \textbf{regain their captioning ability}. The output bias is corrected, and they once again produce the correct, descriptive caption (e.g., "A pair of grey and orange knitted boots").
  \end{itemize}
This provides a powerful, intuitive visualization to confirm the hypothesis: "forgetting" is likely a temporary output distribution shift, and "recovery" is the re-correction of that shift by a subsequent task's training signal.}


\section{Forgetting mitigation approach descriptions}
\label{sec:forgetting_mitigation_approach}

\paragraph{Low-rank adaptation (LoRA).}
When memory or compute restricts full fine‑tuning, LoRA~\citep{huLoRALowRankAdaptation2022} offers a lightweight alternative.
For a weight matrix $W_0 \in \mathbb{R}^{d\times k}$, we introduce two trainable low‑rank matrices $A \in \mathbb{R}^{r\times k}$ and $B \in \mathbb{R}^{d\times r}$ and model the update as $\Delta W = BA$.
After optimization, the effective weight becomes
\[
  W  =  W_0  +  \tfrac{\alpha}{r} BA,
\]
where $\alpha$ is a scalar scaling factor.
Because only $A$ and $B$ are updated, the number of learned parameters per task drops from $dk$ to $r(d+k)$, which is substantial when $r \ll \min(d,k)$.

In the continual‑learning setting, we instantiate a fresh pair $(A^{t},B^{t})$ for each task $\mathcal{T}_t$ while keeping the backbone weights frozen.
After completing task~$\mathcal{T}_t$, we merge the low‑rank update into the backbone weight $W_{t} \leftarrow W_{t-1} + \frac{\alpha}{r}B^{t}A^{t}$ and then discard the adapters.
This maintains a {\it constant} parameter footprint across tasks and avoids accumulating a growing set of task‑specific modules.

\paragraph{Weight-space interpolation}
Weight‑space interpolation~\citep{WiSE-FT} forms an implicit ensemble by linearly combining the pretrained/base checkpoint with the fine‑tuned checkpoint.
Given the base weights $W_{\mathrm{base}}$ (the original \ov\ checkpoint) and the fine‑tuned weights after stage $t$, $W_{t}^{\mathrm{FT}}$, we build an interpolated model
\begin{equation}
  W_{t}^{(\beta)} \;=\; (1-\beta)\, W_{\mathrm{base}} \; +\; \beta\, W_{t}^{\mathrm{FT}}, \quad \beta \in [0,1].
  \label{eq:wiseft}
\end{equation}
The coefficient $\beta$ trades off specialization on the current target task (larger $\beta$) against retention of general capabilities (smaller $\beta$).

In our sequential setting, we apply Eq.~\eqref{eq:wiseft} \emph{after} finishing fine‑tuning on task $\mathcal{T}_t$ and evaluate $W_{t}^{(\beta)}$.  Unless otherwise noted, optimization for the next stage continues from $W_{t}^{\mathrm{FT}}$ (not from $W_{t}^{(\beta)}$) to avoid repeatedly biasing training toward the base weights.
We sweep $\beta \in \{0.1, 0.3, 0.5, 0.7, 0.9\}$ and report $\beta{=}0.3$'s result for comparison as it leads to better learning and forgetting tradeoff compared to results obtained from other $\beta$-s.

\paragraph{Mixture of Experts.} We next leverage the \emph{Mixture of Experts} (MoE) architecture to expand model capacity without overwriting knowledge learned during pretraining~\citep{Wei_Skywork-MoE_2024}.
An MoE layer combines a set of specialist networks (experts) $\{E_{i}\}_{i=1}^{N}$ through a learnable \emph{gating network} $g$ that produces input dependent weights.
The layer output is
\[
  l \;=\; \sum_{i=1}^{N} g_{i}(x)\,E_{i}(x),
\]
typically with a sparsity constraint such as top‑$k$ gating so that only a few experts are active per input.
We follow standard practice and replace the feed‑forward (MLP) submodule in every transformer decoder block of the language model with an MoE layer.


At the start of continual training, each decoder block contains
1) the \textbf{pretrained expert} $E_{\mathrm{pt}}$ that stores upstream knowledge and
2) a new \textbf{tuned expert} $E_{\mathrm{new}}$ that is a copy of $E_{\mathrm{pt}}$.
The gating network is a linear layer initialized to all zeros, which initially routes the entire token sequence through $E_{\mathrm{pt}}$.
During task $t$, we {\it freeze} $E_{\mathrm{pt}}$ and update only the parameters of $E_{\mathrm{new}}$ and the gate.
Because $E_{\mathrm{pt}}$ remains untouched, it acts as a safeguard when the tuned expert fails, giving MoE an inherent resistance to forgetting.
We repeat this procedure for every new task, always reusing the same pair~$(E_{\mathrm{pt}},E_{\mathrm{new}})$ and thus adding no extra parameters beyond the current tuned expert and gate.

\section{Detailed task performance}

\subsection{Forgetting mitigation methods sequential tuning tables}

\begin{table}[t]
\caption{Detailed performance of using \textbf{MoE} to mitigate forgetting by performing sequential fine-tuning on each target task.}
\label{tab:moe_baseline_task_stage_performance_extracted}
\centering
\begin{tabular}{l c c c c c c}
\toprule
\multicolumn{1}{c}{\multirow{2}{*}{Dataset}} & \multicolumn{1}{c}{Baseline} & \multicolumn{1}{c}{Stage 1} & \multicolumn{1}{c}{Stage 2} & \multicolumn{1}{c}{Stage 3} & \multicolumn{1}{c}{Stage 4} & \multicolumn{1}{c}{Stage 5} \\
& \multicolumn{1}{c}{--} & \multicolumn{1}{c}{CUB200} & \multicolumn{1}{c}{PixmoCount} & \multicolumn{1}{c}{PathVQA} & \multicolumn{1}{c}{TextVQA} & \multicolumn{1}{c}{TimeClock} \\
\midrule
\rowcolor{Gray}\multicolumn{7}{l}{\textbf{Target}}\\
CUB200 & 53.7 & 87.7 & 87.6 & 87.4 & 87.3 & 87.0 \\
PixmoCount & 52.4 & 53.9 & 65.5 & 68.2 & 63.5 & 59.9 \\
PathVQA & 36.3 & 36.2 & 35.4 & 61.3 & 57.4 & 57.8 \\
TextVQA & 76.0 & 76.0 & 75.3 & 75.3 & 78.5 & 76.3 \\
TimeClock & 1.1 & 1.0 & 1.9 & 1.2 & 1.2 & 68.0 \\
\midrule
\textbf{Average} & 43.9 & 51.0 & 53.1 & 58.7 & 57.6 & 69.8 \\
\midrule
\rowcolor{Gray}\multicolumn{7}{l}{\textbf{Held out}}\\
AI2D & 81.4 & 81.5 & 81.2 & 80.5 & 81.3 & 80.8 \\
ChartQA & 80.1 & 80.2 & 79.6 & 79.6 & 80.0 & 76.1 \\
DocVQA & 87.1 & 87.2 & 85.2 & 85.8 & 86.6 & 84.5 \\
InfoVQA & 65.9 & 65.6 & 64.0 & 64.8 & 66.4 & 64.4 \\
MMStar & 61.8 & 62.1 & 62.6 & 62.6 & 62.5 & 61.4 \\
RealWorldQA & 66.4 & 67.7 & 62.2 & 63.9 & 69.3 & 68.0 \\
ScienceQA & 95.9 & 95.8 & 95.9 & 95.7 & 96.3 & 96.0 \\
SeedBench & 72.4 & 72.5 & 72.2 & 71.6 & 72.4 & 72.1 \\
\midrule
\textbf{Average} & 76.4 & 76.6 & 75.4 & 75.6 & 76.8 & 75.4 \\
\bottomrule
\end{tabular}
\end{table}

\begin{table}[t]
\caption{Detailed performance of using \textbf{LoRA} to mitigate forgetting by performing sequential fine-tuning on each target task.}
\label{tab:Lora-7B}
\centering
\begin{tabular}{l c c c c c c}
\toprule
\multicolumn{1}{c}{\multirow{2}{*}{Dataset}} & \multicolumn{1}{c}{Baseline} & \multicolumn{1}{c}{Stage 1} & \multicolumn{1}{c}{Stage 2} & \multicolumn{1}{c}{Stage 3} & \multicolumn{1}{c}{Stage 4} & \multicolumn{1}{c}{Stage 5} \\
& \multicolumn{1}{c}{--} & \multicolumn{1}{c}{CUB200} & \multicolumn{1}{c}{PixmoCount} & \multicolumn{1}{c}{PathVQA} & \multicolumn{1}{c}{TextVQA} & \multicolumn{1}{c}{TimeClock} \\
\midrule
\rowcolor{Gray}\multicolumn{7}{l}{\textbf{Target}}\\
CUB200 & 53.7 & 80.6 & 80.0 & 78.0 & 78.2 & 77.0 \\
PixmoCount & 52.4 & 52.6 & 67.8 & 64.2 & 62.5 & 63.7 \\
PathVQA & 36.3 & 36.2 & 35.1 & 58.1 & 51.6 & 53.3 \\
TextVQA & 76.0 & 76.0 & 75.2 & 75.2 & 79.1 & 76.2 \\
TimeClock & 1.1 & 1.0 & 1.0 & 1.0 & 1.2 & 33.7 \\
\midrule
\textbf{Average} & 43.9 & 49.3 & 51.8 & 55.3 & 54.5 & 60.8 \\
\midrule
\rowcolor{Gray}\multicolumn{7}{l}{\textbf{Held out}}\\
AI2D & 81.4 & 81.9 & 80.8 & 79.7 & 81.3 & 79.7 \\
ChartQA & 80.1 & 80.1 & 79.2 & 78.1 & 79.1 & 71.7 \\
DocVQA & 87.1 & 87.1 & 85.0 & 83.4 & 84.4 & 74.1 \\
InfoVQA & 65.9 & 66.1 & 63.9 & 62.5 & 64.4 & 59.3 \\
MMStar & 61.8 & 62.1 & 62.2 & 60.5 & 61.2 & 60.3 \\
RealWorldQA & 66.4 & 67.6 & 66.4 & 65.9 & 68.8 & 66.3 \\
ScienceQA & 95.9 & 96.0 & 95.4 & 95.2 & 95.6 & 93.5 \\
SeedBench & 72.4 & 72.5 & 72.2 & 71.9 & 72.6 & 72.4 \\
\midrule
\textbf{Average} & 76.4 & 76.7 & 75.6 & 74.7 & 75.9 & 72.2 \\
\bottomrule
\end{tabular}
\end{table}

\begin{table}[t]
\caption{Detailed performance of using \textbf{LwF} to mitigate forgetting by performing sequential fine-tuning on each target task.}
\label{tab:lwf_task_stage_performance_extracted}
\centering
\begin{tabular}{l c c c c c c}
\toprule
\multicolumn{1}{c}{\multirow{2}{*}{Dataset}} & \multicolumn{1}{c}{Baseline} & \multicolumn{1}{c}{Stage 1} & \multicolumn{1}{c}{Stage 2} & \multicolumn{1}{c}{Stage 3} & \multicolumn{1}{c}{Stage 4} & \multicolumn{1}{c}{Stage 5} \\
& \multicolumn{1}{c}{--} & \multicolumn{1}{c}{CUB200} & \multicolumn{1}{c}{PixmoCount} & \multicolumn{1}{c}{PathVQA} & \multicolumn{1}{c}{TextVQA} & \multicolumn{1}{c}{TimeClock} \\
\midrule
\rowcolor{Gray}\multicolumn{7}{l}{\textbf{Target}}\\
CUB200 & 53.7 & 90.0 & 89.8 & 89.6 & 89.5 & 89.3 \\
PixmoCount & 52.4 & 53.2 & 67.6 & 67.8 & 61.6 & 61.0 \\
PathVQA & 36.3 & 35.9 & 35.0 & 61.1 & 58.3 & 57.6 \\
TextVQA & 76.0 & 76.3 & 76.5 & 76.6 & 80.3 & 79.4 \\
TimeClock & 1.1 & 1.0 & 1.7 & 1.0 & 1.4 & 67.8 \\
\midrule
\textbf{Average} & 43.9 & 51.3 & 54.1 & 59.2 & 58.2 & 71.0 \\
\midrule
\rowcolor{Gray}\multicolumn{7}{l}{\textbf{Held out}}\\
AI2D & 81.4 & 81.9 & 81.6 & 81.6 & 81.6 & 81.7 \\
ChartQA & 80.1 & 79.9 & 79.9 & 79.9 & 80.2 & 78.0 \\
DocVQA & 87.1 & 87.1 & 86.6 & 86.5 & 86.5 & 85.6 \\
InfoVQA & 65.9 & 66.3 & 66.0 & 65.8 & 66.1 & 65.2 \\
MMStar & 61.8 & 62.4 & 63.2 & 62.6 & 62.0 & 61.6 \\
RealWorldQA & 66.4 & 67.8 & 59.6 & 64.4 & 67.8 & 66.9 \\
ScienceQA & 95.9 & 96.0 & 93.0 & 95.8 & 96.3 & 96.0 \\
SeedBench & 72.4 & 72.4 & 72.5 & 72.2 & 72.5 & 72.5 \\
\midrule
\textbf{Average} & 76.4 & 76.7 & 75.3 & 76.1 & 76.6 & 75.9 \\
\bottomrule
\end{tabular}
\end{table}

\begin{table}[t]
\caption{Detailed performance of using \textbf{WiSE-FT} using $\beta=0.3$ to mitigate forgetting by performing sequential fine-tuning on each target task.}
\label{tab:wise_ft_0.3_for_prev_model_task_stage_performance_extracted}
\centering
\begin{tabular}{l c c c c c c}
\toprule
\multicolumn{1}{c}{\multirow{2}{*}{Dataset}} & \multicolumn{1}{c}{Baseline} & \multicolumn{1}{c}{Stage 1} & \multicolumn{1}{c}{Stage 2} & \multicolumn{1}{c}{Stage 3} & \multicolumn{1}{c}{Stage 4} & \multicolumn{1}{c}{Stage 5} \\
& \multicolumn{1}{c}{--} & \multicolumn{1}{c}{CUB200} & \multicolumn{1}{c}{PixmoCount} & \multicolumn{1}{c}{PathVQA} & \multicolumn{1}{c}{TextVQA} & \multicolumn{1}{c}{TimeClock} \\
\midrule
\rowcolor{Gray}\multicolumn{7}{l}{\textbf{Target}}\\
CUB200 & 53.7 & 89.9 & 89.2 & 89.5 & 89.0 & 89.0 \\
PixmoCount & 52.4 & 53.7 & 69.7 & 68.0 & 68.0 & 66.1 \\
PathVQA & 36.3 & 35.9 & 34.2 & 59.1 & 56.3 & 56.1 \\
TextVQA & 76.0 & 76.4 & 74.2 & 76.1 & 79.7 & 78.3 \\
TimeClock & 1.1 & 1.0 & 1.8 & 1.0 & 1.4 & 62.2 \\
\midrule
\textbf{Average} & 43.9 & 51.4 & 53.8 & 58.7 & 58.9 & 70.3 \\
\midrule
\rowcolor{Gray}\multicolumn{7}{l}{\textbf{Held out}}\\
AI2D & 81.4 & 81.8 & 81.8 & 81.6 & 81.6 & 81.8 \\
ChartQA & 80.1 & 80.1 & 80.4 & 80.2 & 80.2 & 78.3 \\
DocVQA & 87.1 & 87.3 & 85.2 & 86.7 & 86.4 & 85.1 \\
InfoVQA & 65.9 & 65.7 & 64.0 & 65.3 & 65.9 & 64.1 \\
MMStar & 61.8 & 61.9 & 61.4 & 61.6 & 62.3 & 60.9 \\
RealWorldQA & 66.4 & 67.7 & 63.9 & 68.4 & 69.7 & 67.8 \\
ScienceQA & 95.9 & 96.3 & 96.3 & 96.3 & 96.4 & 96.3 \\
SeedBench & 72.4 & 72.6 & 72.9 & 72.6 & 72.6 & 72.6 \\
\midrule
\textbf{Average} & 76.4 & 76.7 & 75.7 & 76.6 & 76.9 & 75.9 \\
\bottomrule
\end{tabular}
\end{table}

Tab.~\ref{tab:Lora-7B}, Tab.~\ref{tab:lwf_task_stage_performance_extracted}, Tab.~\ref{tab:wise_ft_0.3_for_prev_model_task_stage_performance_extracted}, and Tab.~\ref{tab:moe_baseline_task_stage_performance_extracted} are detailed sequential tuning tables for forgetting mitigation approaches tested in the paper, i.e., LoRA, LwF~\citep{li2017learning}, WiSE-FT~\citep{WiSE-FT}, and MoE.

\subsection{Sequential tuning detailed performance tables on \ov}

We include the detailed task performances for sequential fine-tuning experiments on \ov\ here.
Tab.~\ref{tab:tune_full_task_stage_performance_extracted}, Tab.~\ref{tab:tune_vision_tower_task_stage_performance_extracted}, Tab.~\ref{tab:tune_projector_task_stage_performance_extracted},
Tab.~\ref{tab:tune_llm_task_stage_performance_extracted},
Tab.~\ref{tab:tune_sa_task_stage_performance_extracted},
Tab.~\ref{tab:tune_sa_wo_o_proj_task_stage_performance_extracted},
Tab.~\ref{tab:tune_mlp_task_stage_performance_extracted}, and Tab.~\ref{tab:tune_parts_mlp_gate_up_task_stage_performance_extracted} are detailed performance tables of sequentially fine-tuning the Full model, Vision Encoder, Projector, LLM, SA projection layers in LLM, SA Proj. (QKV), MLP layers in LLM, MLP (Gate\&Up), respectively.

\begin{table}[t]
\caption{Detailed performance of sequentially fine-tuning the \textbf{full} model on each target task.}
\label{tab:tune_full_task_stage_performance_extracted}
\centering
\begin{tabular}{l c c c c c c}
\toprule
\multicolumn{1}{c}{\multirow{2}{*}{Dataset}} & \multicolumn{1}{c}{Baseline} & \multicolumn{1}{c}{Stage 1} & \multicolumn{1}{c}{Stage 2} & \multicolumn{1}{c}{Stage 3} & \multicolumn{1}{c}{Stage 4} & \multicolumn{1}{c}{Stage 5} \\
& \multicolumn{1}{c}{--} & \multicolumn{1}{c}{CUB200} & \multicolumn{1}{c}{PixmoCount} & \multicolumn{1}{c}{PathVQA} & \multicolumn{1}{c}{TextVQA} & \multicolumn{1}{c}{TimeClock} \\
\midrule
\rowcolor{Gray}\multicolumn{7}{l}{\textbf{Target}}\\
CUB200 & 53.7 & 90.7 & 89.2 & 88.1 & 88.0 & 86.9 \\
PixmoCount & 52.4 & 54.9 & 73.0 & 64.6 & 63.1 & 59.4 \\
PathVQA & 36.3 & 34.8 & 3.7 & 63.6 & 59.8 & 58.6 \\
TextVQA & 76.0 & 76.6 & 59.0 & 74.6 & 79.6 & 68.9 \\
TimeClock & 1.1 & 1.0 & 1.4 & 1.2 & 1.5 & 46.9 \\
\midrule
\textbf{Average} & 43.9 & 51.6 & 45.3 & 58.4 & 58.4 & 64.1 \\
\midrule
\rowcolor{Gray}\multicolumn{7}{l}{\textbf{Held out}}\\
AI2D & 81.4 & 81.4 & 57.9 & 80.4 & 80.3 & 74.7 \\
ChartQA & 80.1 & 80.3 & 63.8 & 77.9 & 77.6 & 66.9 \\
DocVQA & 87.1 & 87.4 & 74.1 & 83.1 & 82.9 & 68.6 \\
InfoVQA & 65.9 & 65.7 & 54.2 & 62.5 & 61.9 & 50.3 \\
MMStar & 61.8 & 60.6 & 59.6 & 58.9 & 59.0 & 53.9 \\
RealWorldQA & 66.4 & 68.6 & 44.2 & 63.4 & 66.1 & 55.8 \\
ScienceQA & 95.9 & 95.0 & 76.0 & 94.6 & 93.5 & 87.9 \\
SeedBench & 72.4 & 72.6 & 71.7 & 70.3 & 71.6 & 65.9 \\
\midrule
\textbf{Average} & 76.4 & 76.5 & 62.7 & 73.9 & 74.1 & 65.5 \\
\bottomrule
\end{tabular}
\end{table}

\begin{table}[t]
\caption{Detailed performance of sequentially fine-tuning the \textbf{vision tower} on each target task.}
\label{tab:tune_vision_tower_task_stage_performance_extracted}
\centering
\begin{tabular}{l c c c c c c}
\toprule
\multicolumn{1}{c}{\multirow{2}{*}{Dataset}} & \multicolumn{1}{c}{Baseline} & \multicolumn{1}{c}{Stage 1} & \multicolumn{1}{c}{Stage 2} & \multicolumn{1}{c}{Stage 3} & \multicolumn{1}{c}{Stage 4} & \multicolumn{1}{c}{Stage 5} \\
& \multicolumn{1}{c}{--} & \multicolumn{1}{c}{CUB200} & \multicolumn{1}{c}{PixmoCount} & \multicolumn{1}{c}{PathVQA} & \multicolumn{1}{c}{TextVQA} & \multicolumn{1}{c}{TimeClock} \\
\midrule
\rowcolor{Gray}\multicolumn{7}{l}{\textbf{Target}}\\
CUB200 & 53.7 & 69.9 & 57.5 & 58.8 & 61.5 & 55.3 \\
PixmoCount & 52.4 & 22.1 & 64.2 & 44.0 & 59.0 & 37.8 \\
PathVQA & 36.3 & 35.4 & 31.8 & 37.0 & 34.4 & 34.2 \\
TextVQA & 76.0 & 75.5 & 72.2 & 70.7 & 76.4 & 72.7 \\
TimeClock & 1.1 & 1.0 & 1.0 & 0.8 & 1.1 & 26.3 \\
\midrule
\textbf{Average} & 43.9 & 40.8 & 45.3 & 42.3 & 46.5 & 45.3 \\
\midrule
\rowcolor{Gray}\multicolumn{7}{l}{\textbf{Held out}}\\
AI2D & 81.4 & 81.3 & 79.7 & 78.9 & 80.6 & 77.3 \\
ChartQA & 80.1 & 80.0 & 76.9 & 74.3 & 79.5 & 76.0 \\
DocVQA & 87.1 & 85.8 & 79.6 & 75.1 & 85.1 & 78.9 \\
InfoVQA & 65.9 & 63.6 & 60.3 & 56.3 & 64.4 & 59.6 \\
MMStar & 61.8 & 61.4 & 57.6 & 56.4 & 59.5 & 55.6 \\
RealWorldQA & 66.4 & 66.3 & 65.0 & 61.3 & 65.6 & 61.8 \\
ScienceQA & 95.9 & 94.6 & 91.9 & 90.5 & 94.3 & 89.4 \\
SeedBench & 72.4 & 71.5 & 70.6 & 69.6 & 71.0 & 69.5 \\
\midrule
\textbf{Average} & 76.4 & 75.6 & 72.7 & 70.3 & 75.0 & 71.0 \\
\bottomrule
\end{tabular}
\end{table}

\begin{table}[t]
\caption{Detailed performance of sequentially fine-tuning the \textbf{projector} in the LLM on each target task.}
\label{tab:tune_projector_task_stage_performance_extracted}
\centering
\begin{tabular}{l c c c c c c}
\toprule
\multicolumn{1}{c}{\multirow{2}{*}{Dataset}} & \multicolumn{1}{c}{Baseline} & \multicolumn{1}{c}{Stage 1} & \multicolumn{1}{c}{Stage 2} & \multicolumn{1}{c}{Stage 3} & \multicolumn{1}{c}{Stage 4} & \multicolumn{1}{c}{Stage 5} \\
& \multicolumn{1}{c}{--} & \multicolumn{1}{c}{CUB200} & \multicolumn{1}{c}{PixmoCount} & \multicolumn{1}{c}{PathVQA} & \multicolumn{1}{c}{TextVQA} & \multicolumn{1}{c}{TimeClock} \\
\midrule
\rowcolor{Gray}\multicolumn{7}{l}{\textbf{Target}}\\
CUB200 & 53.7 & 59.4 & 57.7 & 57.5 & 58.2 & 57.4 \\
PixmoCount & 52.4 & 53.2 & 56.6 & 57.9 & 58.2 & 57.7 \\
PathVQA & 36.3 & 36.1 & 36.0 & 35.5 & 36.4 & 35.9 \\
TextVQA & 76.0 & 76.1 & 76.3 & 76.4 & 77.0 & 76.9 \\
TimeClock & 1.1 & 1.1 & 1.2 & 1.2 & 1.0 & 1.9 \\
\midrule
\textbf{Average} & 43.9 & 45.2 & 45.6 & 45.7 & 46.2 & 46.0 \\
\midrule
\rowcolor{Gray}\multicolumn{7}{l}{\textbf{Held out}}\\
AI2D & 81.4 & 81.4 & 81.7 & 81.6 & 81.8 & 81.1 \\
ChartQA & 80.1 & 79.9 & 80.2 & 80.0 & 80.1 & 79.4 \\
DocVQA & 87.1 & 87.3 & 87.2 & 86.1 & 86.3 & 86.2 \\
InfoVQA & 65.9 & 66.1 & 66.1 & 65.5 & 66.3 & 65.1 \\
MMStar & 61.8 & 62.1 & 61.7 & 60.9 & 61.0 & 60.5 \\
RealWorldQA & 66.4 & 66.1 & 66.9 & 67.3 & 68.0 & 67.1 \\
ScienceQA & 95.9 & 95.9 & 96.0 & 95.9 & 95.8 & 95.6 \\
SeedBench & 72.4 & 72.6 & 72.5 & 72.3 & 72.5 & 72.4 \\
\midrule
\textbf{Average} & 76.4 & 76.4 & 76.5 & 76.2 & 76.5 & 75.9 \\
\bottomrule
\end{tabular}
\end{table}

\begin{table}[t]
\caption{Detailed performance of sequentially fine-tuning the \textbf{LLM} on each target task.}
\label{tab:tune_llm_task_stage_performance_extracted}
\centering
\begin{tabular}{l c c c c c c}
\toprule
\multicolumn{1}{c}{\multirow{2}{*}{Dataset}} & \multicolumn{1}{c}{Baseline} & \multicolumn{1}{c}{Stage 1} & \multicolumn{1}{c}{Stage 2} & \multicolumn{1}{c}{Stage 3} & \multicolumn{1}{c}{Stage 4} & \multicolumn{1}{c}{Stage 5} \\
& \multicolumn{1}{c}{--} & \multicolumn{1}{c}{CUB200} & \multicolumn{1}{c}{PixmoCount} & \multicolumn{1}{c}{PathVQA} & \multicolumn{1}{c}{TextVQA} & \multicolumn{1}{c}{TimeClock} \\
\midrule
\rowcolor{Gray}\multicolumn{7}{l}{\textbf{Target}}\\
CUB200 & 53.7 & 90.7 & 89.4 & 89.6 & 88.9 & 87.8 \\
PixmoCount & 52.4 & 54.3 & 70.2 & 62.2 & 63.3 & 56.6 \\
PathVQA & 36.3 & 35.2 & 4.4 & 63.2 & 58.6 & 56.7 \\
TextVQA & 76.0 & 76.6 & 60.5 & 74.7 & 79.6 & 71.2 \\
TimeClock & 1.1 & 1.0 & 1.4 & 1.0 & 1.3 & 71.8 \\
\midrule
\textbf{Average} & 43.9 & 51.6 & 45.2 & 58.1 & 58.3 & 68.8 \\
\midrule
\rowcolor{Gray}\multicolumn{7}{l}{\textbf{Held out}}\\
AI2D & 81.4 & 81.2 & 72.8 & 80.7 & 79.8 & 75.2 \\
ChartQA & 80.1 & 80.4 & 66.9 & 78.4 & 78.0 & 68.6 \\
DocVQA & 87.1 & 87.3 & 75.9 & 84.2 & 83.0 & 72.0 \\
InfoVQA & 65.9 & 65.8 & 54.9 & 62.8 & 61.6 & 51.8 \\
MMStar & 61.8 & 60.8 & 58.4 & 58.9 & 59.2 & 53.4 \\
RealWorldQA & 66.4 & 67.5 & 46.4 & 63.5 & 67.2 & 59.0 \\
ScienceQA & 95.9 & 94.8 & 83.6 & 94.7 & 92.5 & 90.0 \\
SeedBench & 72.4 & 72.7 & 72.3 & 71.0 & 71.5 & 65.3 \\
\midrule
\textbf{Average} & 76.4 & 76.3 & 66.4 & 74.3 & 74.1 & 66.9 \\
\bottomrule
\end{tabular}
\end{table}

\begin{table}[t]
\caption{Detailed performance of sequentially fine-tuning the \textbf{SA projection layers in the LLM} on each target task.}
\label{tab:tune_sa_task_stage_performance_extracted}
\centering
\begin{tabular}{l c c c c c c}
\toprule
\multicolumn{1}{c}{\multirow{2}{*}{Dataset}} & \multicolumn{1}{c}{Baseline} & \multicolumn{1}{c}{Stage 1} & \multicolumn{1}{c}{Stage 2} & \multicolumn{1}{c}{Stage 3} & \multicolumn{1}{c}{Stage 4} & \multicolumn{1}{c}{Stage 5} \\
& \multicolumn{1}{c}{--} & \multicolumn{1}{c}{CUB200} & \multicolumn{1}{c}{PixmoCount} & \multicolumn{1}{c}{PathVQA} & \multicolumn{1}{c}{TextVQA} & \multicolumn{1}{c}{TimeClock} \\
\midrule
\rowcolor{Gray}\multicolumn{7}{l}{\textbf{Target}}\\
CUB200 & 53.7 & 85.5 & 85.1 & 84.8 & 84.4 & 84.0 \\
PixmoCount & 52.4 & 53.9 & 67.8 & 68.2 & 64.8 & 66.3 \\
PathVQA & 36.3 & 35.7 & 35.0 & 55.9 & 52.6 & 51.4 \\
TextVQA & 76.0 & 76.1 & 76.4 & 75.8 & 79.3 & 78.9 \\
TimeClock & 1.1 & 1.0 & 1.2 & 1.0 & 1.2 & 52.6 \\
\midrule
\textbf{Average} & 43.9 & 50.4 & 53.1 & 57.1 & 56.5 & 66.6 \\
\midrule
\rowcolor{Gray}\multicolumn{7}{l}{\textbf{Held out}}\\
AI2D & 81.4 & 82.0 & 81.4 & 81.2 & 81.9 & 81.9 \\
ChartQA & 80.1 & 80.0 & 79.7 & 80.0 & 80.6 & 79.4 \\
DocVQA & 87.1 & 87.2 & 86.9 & 86.8 & 86.3 & 86.1 \\
InfoVQA & 65.9 & 66.0 & 64.7 & 65.3 & 66.0 & 64.9 \\
MMStar & 61.8 & 62.4 & 62.3 & 62.1 & 62.4 & 61.9 \\
RealWorldQA & 66.4 & 68.0 & 67.1 & 66.9 & 69.2 & 68.9 \\
ScienceQA & 95.9 & 95.7 & 95.9 & 96.1 & 96.3 & 96.1 \\
SeedBench & 72.4 & 72.3 & 72.4 & 72.0 & 72.5 & 72.4 \\
\midrule
\textbf{Average} & 76.4 & 76.7 & 76.3 & 76.3 & 76.9 & 76.5 \\
\bottomrule
\end{tabular}
\end{table}

\begin{table}[t]
\caption{Detailed performance of sequentially fine-tuning the \textbf{SA Proj. (QKV)} in the LLM on each target task.}
\label{tab:tune_sa_wo_o_proj_task_stage_performance_extracted}
\centering
\begin{tabular}{l c c c c c c}
\toprule
\multicolumn{1}{c}{\multirow{2}{*}{Dataset}} & \multicolumn{1}{c}{Baseline} & \multicolumn{1}{c}{Stage 1} & \multicolumn{1}{c}{Stage 2} & \multicolumn{1}{c}{Stage 3} & \multicolumn{1}{c}{Stage 4} & \multicolumn{1}{c}{Stage 5} \\
& \multicolumn{1}{c}{--} & \multicolumn{1}{c}{CUB200} & \multicolumn{1}{c}{PixmoCount} & \multicolumn{1}{c}{PathVQA} & \multicolumn{1}{c}{TextVQA} & \multicolumn{1}{c}{TimeClock} \\
\midrule
\rowcolor{Gray}\multicolumn{7}{l}{\textbf{Target}}\\
CUB200 & 53.7 & 78.7 & 78.4 & 78.4 & 78.2 & 77.6 \\
PixmoCount & 52.4 & 53.2 & 65.9 & 67.2 & 62.4 & 65.4 \\
PathVQA & 36.3 & 36.1 & 36.4 & 44.2 & 42.6 & 43.0 \\
TextVQA & 76.0 & 76.2 & 76.9 & 76.2 & 78.6 & 78.3 \\
TimeClock & 1.1 & 1.0 & 1.0 & 0.8 & 1.1 & 32.2 \\
\midrule
\textbf{Average} & 43.9 & 49.0 & 51.7 & 53.4 & 52.6 & 59.3 \\
\midrule
\rowcolor{Gray}\multicolumn{7}{l}{\textbf{Held out}}\\
AI2D & 81.4 & 81.9 & 81.7 & 81.2 & 82.0 & 81.9 \\
ChartQA & 80.1 & 79.9 & 79.9 & 79.8 & 80.4 & 79.9 \\
DocVQA & 87.1 & 87.2 & 87.3 & 86.9 & 86.7 & 86.4 \\
InfoVQA & 65.9 & 65.8 & 65.4 & 65.7 & 65.9 & 65.8 \\
MMStar & 61.8 & 62.3 & 62.5 & 62.7 & 62.3 & 62.3 \\
RealWorldQA & 66.4 & 67.6 & 67.6 & 67.1 & 68.4 & 68.8 \\
ScienceQA & 95.9 & 95.9 & 95.8 & 95.9 & 96.3 & 96.1 \\
SeedBench & 72.4 & 72.2 & 72.4 & 72.1 & 72.3 & 72.3 \\
\midrule
\textbf{Average} & 76.4 & 76.6 & 76.6 & 76.4 & 76.8 & 76.7 \\
\bottomrule
\end{tabular}
\end{table}

\begin{table}[t]
\caption{Detailed performance of sequentially fine-tuning the \textbf{MLP in the LLM} on each target task.}
\label{tab:tune_mlp_task_stage_performance_extracted}
\centering
\begin{tabular}{l c c c c c c}
\toprule
\multicolumn{1}{c}{\multirow{2}{*}{Dataset}} & \multicolumn{1}{c}{Baseline} & \multicolumn{1}{c}{Stage 1} & \multicolumn{1}{c}{Stage 2} & \multicolumn{1}{c}{Stage 3} & \multicolumn{1}{c}{Stage 4} & \multicolumn{1}{c}{Stage 5} \\
& \multicolumn{1}{c}{--} & \multicolumn{1}{c}{CUB200} & \multicolumn{1}{c}{PixmoCount} & \multicolumn{1}{c}{PathVQA} & \multicolumn{1}{c}{TextVQA} & \multicolumn{1}{c}{TimeClock} \\
\midrule
\rowcolor{Gray}\multicolumn{7}{l}{\textbf{Target}}\\
CUB200 & 53.7 & 90.1 & 89.5 & 89.6 & 89.3 & 88.9 \\
PixmoCount & 52.4 & 54.1 & 71.5 & 67.6 & 68.0 & 62.0 \\
PathVQA & 36.3 & 35.6 & 17.0 & 64.1 & 60.9 & 60.9 \\
TextVQA & 76.0 & 76.6 & 66.2 & 75.3 & 79.8 & 74.0 \\
TimeClock & 1.1 & 1.0 & 1.5 & 1.2 & 1.6 & 74.0 \\
\midrule
\textbf{Average} & 43.9 & 51.5 & 49.1 & 59.6 & 59.9 & 72.0 \\
\midrule
\rowcolor{Gray}\multicolumn{7}{l}{\textbf{Held out}}\\
AI2D & 81.4 & 81.7 & 80.9 & 81.0 & 80.5 & 80.4 \\
ChartQA & 80.1 & 80.4 & 75.7 & 79.7 & 79.9 & 75.1 \\
DocVQA & 87.1 & 87.2 & 80.0 & 85.5 & 84.5 & 78.9 \\
InfoVQA & 65.9 & 65.8 & 60.0 & 64.0 & 63.7 & 59.3 \\
MMStar & 61.8 & 61.5 & 61.2 & 61.0 & 60.7 & 59.5 \\
RealWorldQA & 66.4 & 68.0 & 53.7 & 65.9 & 68.1 & 62.1 \\
ScienceQA & 95.9 & 96.1 & 93.7 & 96.2 & 95.9 & 95.2 \\
SeedBench & 72.4 & 72.7 & 72.7 & 72.1 & 72.3 & 71.9 \\
\midrule
\textbf{Average} & 76.4 & 76.7 & 72.2 & 75.7 & 75.7 & 72.8 \\
\bottomrule
\end{tabular}
\end{table}

\begin{table}[t]
\caption{Detailed performance of sequentially fine-tuning the \textbf{MLP (Gate \& Up)} in the LLM on each target task.}
\label{tab:tune_parts_mlp_gate_up_task_stage_performance_extracted}
\centering
\begin{tabular}{l c c c c c c}
\toprule
\multicolumn{1}{c}{\multirow{2}{*}{Dataset}} & \multicolumn{1}{c}{Baseline} & \multicolumn{1}{c}{Stage 1} & \multicolumn{1}{c}{Stage 2} & \multicolumn{1}{c}{Stage 3} & \multicolumn{1}{c}{Stage 4} & \multicolumn{1}{c}{Stage 5} \\
& \multicolumn{1}{c}{--} & \multicolumn{1}{c}{CUB200} & \multicolumn{1}{c}{PixmoCount} & \multicolumn{1}{c}{PathVQA} & \multicolumn{1}{c}{TextVQA} & \multicolumn{1}{c}{TimeClock} \\
\midrule
\rowcolor{Gray}\multicolumn{7}{l}{\textbf{Target}}\\
CUB200 & 53.7 & 90.2 & 89.8 & 89.8 & 89.6 & 89.5 \\
PixmoCount & 52.4 & 53.4 & 71.5 & 67.8 & 68.4 & 67.2 \\
PathVQA & 36.3 & 36.1 & 35.0 & 61.9 & 58.5 & 58.9 \\
TextVQA & 76.0 & 76.4 & 75.5 & 76.1 & 80.0 & 79.3 \\
TimeClock & 1.1 & 0.9 & 1.8 & 1.2 & 1.9 & 72.2 \\
\midrule
\textbf{Average} & 43.9 & 51.4 & 54.7 & 59.4 & 59.7 & 73.4 \\
\midrule
\rowcolor{Gray}\multicolumn{7}{l}{\textbf{Held out}}\\
AI2D & 81.4 & 81.7 & 81.6 & 81.3 & 81.2 & 81.5 \\
ChartQA & 80.1 & 80.1 & 80.6 & 80.1 & 80.7 & 78.8 \\
DocVQA & 87.1 & 87.0 & 86.3 & 86.6 & 85.9 & 85.4 \\
InfoVQA & 65.9 & 66.1 & 65.4 & 65.0 & 65.3 & 64.9 \\
MMStar & 61.8 & 62.2 & 63.1 & 62.7 & 62.5 & 62.0 \\
RealWorldQA & 66.4 & 67.7 & 64.4 & 67.8 & 69.5 & 68.4 \\
ScienceQA & 95.9 & 96.3 & 96.4 & 96.5 & 96.2 & 96.0 \\
SeedBench & 72.4 & 72.4 & 72.6 & 72.3 & 72.6 & 72.5 \\
\midrule
\textbf{Average} & 76.4 & 76.7 & 76.3 & 76.5 & 76.7 & 76.2 \\
\bottomrule
\end{tabular}
\end{table}

\subsection{Sequential tuning detailed performance tables on LLaVA-NeXT (LLaMA 3)}

We include the detailed task performances for sequential fine-tuning experiments on LLaVA-NeXT (LLaMA 3) here.
Tab.~\ref{tab:llama3_tune_parts_full_model_task_stage_performance_extracted}, Tab.~\ref{tab:llama3_tune_parts_vision_and_projector_task_stage_performance_extracted}, Tab.~\ref{tab:llama3_tune_parts_llm_task_stage_performance_extracted},
Tab.~\ref{tab:LLaVA-LLaMA_3_SA_Proj_task_stage_performance_extracted},
Tab.~\ref{tab:LLaVA-LLaMA_3_MLP_task_stage_performance_extracted}, and
Tab.~\ref{tab:llama3_tune_parts_mlp_gate_and_up_task_stage_performance_extracted} are detailed performance tables of sequentially fine-tuning the Full model, Vision Encoder + Projector, LLM, SA projection layers in LLM, MLP layers in LLM, and MLP (Gate\&Up), respectively.

\begin{table}[t]
\caption{Detailed performance of sequentially fine-tuning the \textbf{full model} of \textbf{LLaVA-NeXT (LLaMA 3)} on each target task.}
\label{tab:llama3_tune_parts_full_model_task_stage_performance_extracted}
\centering
\begin{tabular}{l c c c c c c}
\toprule
\multicolumn{1}{c}{\multirow{2}{*}{Dataset}} & \multicolumn{1}{c}{Baseline} & \multicolumn{1}{c}{Stage 1} & \multicolumn{1}{c}{Stage 2} & \multicolumn{1}{c}{Stage 3} & \multicolumn{1}{c}{Stage 4} & \multicolumn{1}{c}{Stage 5} \\
& \multicolumn{1}{c}{--} & \multicolumn{1}{c}{CUB200} & \multicolumn{1}{c}{PixmoCount} & \multicolumn{1}{c}{PathVQA} & \multicolumn{1}{c}{TextVQA} & \multicolumn{1}{c}{TimeClock} \\
\midrule
\rowcolor{Gray}\multicolumn{7}{l}{\textbf{Target}}\\
CUB200 & 32.6 & 84.8 & 76.8 & 77.2 & 76.6 & 69.6 \\
PixmoCount & 45.7 & 37.6 & 63.3 & 48.5 & 32.4 & 44.0 \\
PathVQA & 13.2 & 24.8 & 0.7 & 62.0 & 55.6 & 45.2 \\
TextVQA & 65.4 & 52.2 & 31.0 & 56.1 & 72.9 & 42.8 \\
TimeClock & 0.8 & 0.3 & 0.1 & 0.6 & 0.5 & 33.1 \\
\midrule
\textbf{Average} & 31.5 & 39.9 & 34.4 & 48.9 & 47.6 & 46.9 \\
\midrule
\rowcolor{Gray}\multicolumn{7}{l}{\textbf{Held out}}\\
AI2D & 71.6 & 54.0 & 53.3 & 62.1 & 58.2 & 43.9 \\
ChartQA & 69.2 & 54.3 & 14.6 & 48.6 & 51.0 & 7.8 \\
DocVQA & 72.7 & 40.4 & 27.7 & 46.6 & 59.2 & 15.7 \\
InfoVQA & 31.9 & 23.4 & 14.6 & 27.2 & 33.9 & 10.2 \\
MMStar & 42.0 & 43.9 & 41.5 & 39.6 & 42.4 & 25.6 \\
RealWorldQA & 59.7 & 55.3 & 32.7 & 50.3 & 53.6 & 19.2 \\
ScienceQA & 73.2 & 63.3 & 57.5 & 69.7 & 66.4 & 58.3 \\
SeedBench & 58.5 & 56.8 & 55.8 & 53.9 & 56.6 & 42.0 \\
\midrule
\textbf{Average} & 59.8 & 48.9 & 37.2 & 49.7 & 52.7 & 27.8 \\
\bottomrule
\end{tabular}
\end{table}

\begin{table}[t]
\caption{Detailed performance of sequentially fine-tuning the \textbf{vision encoder and projector} of \textbf{LLaVA-NeXT (LLaMA 3)} on each target task.}
\label{tab:llama3_tune_parts_vision_and_projector_task_stage_performance_extracted}
\centering
\begin{tabular}{l c c c c c c}
\toprule
\multicolumn{1}{c}{\multirow{2}{*}{Dataset}} & \multicolumn{1}{c}{Baseline} & \multicolumn{1}{c}{Stage 1} & \multicolumn{1}{c}{Stage 2} & \multicolumn{1}{c}{Stage 3} & \multicolumn{1}{c}{Stage 4} & \multicolumn{1}{c}{Stage 5} \\
& \multicolumn{1}{c}{--} & \multicolumn{1}{c}{CUB200} & \multicolumn{1}{c}{PixmoCount} & \multicolumn{1}{c}{PathVQA} & \multicolumn{1}{c}{TextVQA} & \multicolumn{1}{c}{TimeClock} \\
\midrule
\rowcolor{Gray}\multicolumn{7}{l}{\textbf{Target}}\\
CUB200 & 32.6 & 4.2 & 4.2 & 12.8 & 22.1 & 13.4 \\
PixmoCount & 45.7 & 0.6 & 44.9 & 40.8 & 48.9 & 40.3 \\
PathVQA & 13.2 & 0.3 & 1.8 & 34.1 & 34.1 & 33.1 \\
TextVQA & 65.4 & 0.9 & 0.9 & 50.1 & 69.9 & 59.3 \\
TimeClock & 0.8 & 0.0 & 0.6 & 0.5 & 0.7 & 5.2 \\
\midrule
\textbf{Average} & 31.5 & 1.2 & 10.5 & 27.7 & 35.1 & 30.3 \\
\midrule
\rowcolor{Gray}\multicolumn{7}{l}{\textbf{Held out}}\\
AI2D & 71.6 & 13.8 & 5.8 & 57.1 & 64.9 & 58.0 \\
ChartQA & 69.2 & 0.1 & 0.5 & 34.9 & 55.3 & 42.7 \\
DocVQA & 72.7 & 0.6 & 1.4 & 36.3 & 59.2 & 38.9 \\
InfoVQA & 31.9 & 0.3 & 0.2 & 22.2 & 29.9 & 24.4 \\
MMStar & 42.0 & 9.9 & 1.9 & 38.1 & 42.3 & 35.8 \\
RealWorldQA & 59.7 & 14.4 & 2.4 & 55.6 & 59.7 & 52.3 \\
ScienceQA & 73.2 & 11.3 & 0.0 & 64.0 & 69.9 & 67.3 \\
SeedBench & 58.5 & 15.1 & 6.0 & 49.7 & 58.7 & 52.3 \\
\midrule
\textbf{Average} & 59.8 & 8.2 & 2.3 & 44.7 & 55.0 & 46.5 \\
\bottomrule
\end{tabular}
\end{table}

\begin{table}[t]
\caption{Detailed performance of sequentially fine-tuning the \textbf{LLM} of \textbf{LLaVA-NeXT (LLaMA 3)} on each target task.}
\label{tab:llama3_tune_parts_llm_task_stage_performance_extracted}
\centering
\begin{tabular}{l c c c c c c}
\toprule
\multicolumn{1}{c}{\multirow{2}{*}{Dataset}} & \multicolumn{1}{c}{Baseline} & \multicolumn{1}{c}{Stage 1} & \multicolumn{1}{c}{Stage 2} & \multicolumn{1}{c}{Stage 3} & \multicolumn{1}{c}{Stage 4} & \multicolumn{1}{c}{Stage 5} \\
& \multicolumn{1}{c}{--} & \multicolumn{1}{c}{CUB200} & \multicolumn{1}{c}{PixmoCount} & \multicolumn{1}{c}{PathVQA} & \multicolumn{1}{c}{TextVQA} & \multicolumn{1}{c}{TimeClock} \\
\midrule
\rowcolor{Gray}\multicolumn{7}{l}{\textbf{Target}}\\
CUB200 & 32.6 & 85.2 & 68.6 & 72.0 & 72.1 & 68.8 \\
PixmoCount & 45.7 & 32.0 & 57.5 & 55.1 & 21.9 & 41.6 \\
PathVQA & 13.2 & 23.3 & 14.1 & 62.7 & 56.4 & 42.7 \\
TextVQA & 65.4 & 56.9 & 35.0 & 57.8 & 72.6 & 40.1 \\
TimeClock & 0.8 & 0.1 & 0.0 & 0.8 & 0.6 & 60.9 \\
\midrule
\textbf{Average} & 31.5 & 39.5 & 35.0 & 49.7 & 44.7 & 50.8 \\
\midrule
\rowcolor{Gray}\multicolumn{7}{l}{\textbf{Held out}}\\
AI2D & 71.6 & 55.5 & 54.4 & 62.7 & 59.9 & 35.1 \\
ChartQA & 69.2 & 54.3 & 20.0 & 49.4 & 53.7 & 4.3 \\
DocVQA & 72.7 & 45.7 & 31.6 & 49.5 & 58.6 & 21.5 \\
InfoVQA & 31.9 & 25.2 & 11.5 & 27.4 & 33.4 & 8.7 \\
MMStar & 42.0 & 42.6 & 40.8 & 40.8 & 38.4 & 16.3 \\
RealWorldQA & 59.7 & 56.1 & 11.8 & 50.8 & 57.1 & 16.1 \\
ScienceQA & 73.2 & 65.5 & 24.3 & 70.8 & 68.7 & 52.2 \\
SeedBench & 58.5 & 56.8 & 55.4 & 54.7 & 56.8 & 37.2 \\
\midrule
\textbf{Average} & 59.8 & 50.2 & 31.2 & 50.8 & 53.3 & 23.9 \\
\bottomrule
\end{tabular}
\end{table}

\begin{table}[t]
\caption{Detailed performance of sequentially fine-tuning the \textbf{SA projection layers in the LLM} of \textbf{LLaVA-NeXT (LLaMA 3)} on each target task.}
\label{tab:LLaVA-LLaMA_3_SA_Proj_task_stage_performance_extracted}
\centering
\begin{tabular}{l c c c c c c}
\toprule
\multicolumn{1}{c}{\multirow{2}{*}{Dataset}} & \multicolumn{1}{c}{Baseline} & \multicolumn{1}{c}{Stage 1} & \multicolumn{1}{c}{Stage 2} & \multicolumn{1}{c}{Stage 3} & \multicolumn{1}{c}{Stage 4} & \multicolumn{1}{c}{Stage 5} \\
& \multicolumn{1}{c}{--} & \multicolumn{1}{c}{CUB200} & \multicolumn{1}{c}{PixmoCount} & \multicolumn{1}{c}{PathVQA} & \multicolumn{1}{c}{TextVQA} & \multicolumn{1}{c}{TimeClock} \\
\midrule
\rowcolor{Gray}\multicolumn{7}{l}{\textbf{Target}}\\
CUB200 & 32.6 & 78.1 & 68.2 & 72.3 & 70.6 & 70.3 \\
PixmoCount & 45.7 & 38.8 & 60.5 & 50.6 & 9.9 & 54.7 \\
PathVQA & 13.2 & 28.7 & 28.4 & 53.9 & 43.9 & 44.3 \\
TextVQA & 65.4 & 64.3 & 62.9 & 62.0 & 73.5 & 64.9 \\
TimeClock & 0.8 & 0.7 & 0.8 & 0.8 & 0.6 & 32.9 \\
\midrule
\textbf{Average} & 31.5 & 42.1 & 44.2 & 47.9 & 39.7 & 53.4 \\
\midrule
\rowcolor{Gray}\multicolumn{7}{l}{\textbf{Held out}}\\
AI2D & 71.6 & 67.6 & 65.3 & 68.5 & 67.4 & 65.4 \\
ChartQA & 69.2 & 60.4 & 58.3 & 61.3 & 63.8 & 49.4 \\
DocVQA & 72.7 & 60.8 & 58.9 & 58.3 & 63.6 & 48.9 \\
InfoVQA & 31.9 & 29.5 & 32.1 & 33.9 & 35.8 & 27.5 \\
MMStar & 42.0 & 46.9 & 45.2 & 43.1 & 42.5 & 41.1 \\
RealWorldQA & 59.7 & 58.6 & 55.4 & 58.3 & 63.3 & 53.9 \\
ScienceQA & 73.2 & 72.2 & 70.4 & 73.4 & 72.6 & 70.7 \\
SeedBench & 58.5 & 59.6 & 60.1 & 58.2 & 59.9 & 60.2 \\
\midrule
\textbf{Average} & 59.8 & 57.0 & 55.7 & 56.9 & 58.6 & 52.1 \\
\bottomrule
\end{tabular}
\end{table}

\begin{table}[t]
\caption{Detailed performance of sequentially fine-tuning the \textbf{MLP in the LLM} of \textbf{LLaVA-NeXT (LLaMA 3)} on each target task.}
\label{tab:LLaVA-LLaMA_3_MLP_task_stage_performance_extracted}
\centering
\begin{tabular}{l c c c c c c}
\toprule
\multicolumn{1}{c}{\multirow{2}{*}{Dataset}} & \multicolumn{1}{c}{Baseline} & \multicolumn{1}{c}{Stage 1} & \multicolumn{1}{c}{Stage 2} & \multicolumn{1}{c}{Stage 3} & \multicolumn{1}{c}{Stage 4} & \multicolumn{1}{c}{Stage 5} \\
& \multicolumn{1}{c}{--} & \multicolumn{1}{c}{CUB200} & \multicolumn{1}{c}{PixmoCount} & \multicolumn{1}{c}{PathVQA} & \multicolumn{1}{c}{TextVQA} & \multicolumn{1}{c}{TimeClock} \\
\midrule
\rowcolor{Gray}\multicolumn{7}{l}{\textbf{Target}}\\
CUB200 & 32.6 & 84.3 & 78.7 & 76.9 & 76.9 & 72.0 \\
PixmoCount & 45.7 & 34.6 & 58.6 & 53.9 & 35.8 & 52.8 \\
PathVQA & 13.2 & 28.5 & 26.3 & 61.8 & 56.2 & 52.7 \\
TextVQA & 65.4 & 61.5 & 54.6 & 62.0 & 73.2 & 59.2 \\
TimeClock & 0.8 & 0.5 & 0.8 & 0.8 & 0.7 & 54.2 \\
\midrule
\textbf{Average} & 31.5 & 41.9 & 43.8 & 51.1 & 48.6 & 58.2 \\
\midrule
\rowcolor{Gray}\multicolumn{7}{l}{\textbf{Held out}}\\
AI2D & 71.6 & 65.3 & 62.8 & 66.3 & 62.1 & 59.1 \\
ChartQA & 69.2 & 59.8 & 50.4 & 58.8 & 57.6 & 32.2 \\
DocVQA & 72.7 & 54.1 & 48.6 & 57.0 & 61.2 & 38.8 \\
InfoVQA & 31.9 & 29.7 & 24.3 & 32.5 & 35.3 & 22.4 \\
MMStar & 42.0 & 44.9 & 43.7 & 43.5 & 41.2 & 35.2 \\
RealWorldQA & 59.7 & 58.3 & 50.3 & 53.5 & 57.3 & 39.5 \\
ScienceQA & 73.2 & 71.7 & 69.8 & 70.3 & 62.8 & 65.6 \\
SeedBench & 58.5 & 59.5 & 58.5 & 57.6 & 58.8 & 55.7 \\
\midrule
\textbf{Average} & 59.8 & 55.4 & 51.1 & 54.9 & 54.5 & 43.6 \\
\bottomrule
\end{tabular}
\end{table}

\begin{table}[t]
\caption{Detailed performance of sequentially fine-tuning the \textbf{MLP (Gate \& Up) in the LLM} of \textbf{LLaVA-NeXT (LLaMA 3)} on each target task.}
\label{tab:llama3_tune_parts_mlp_gate_and_up_task_stage_performance_extracted}
\centering
\begin{tabular}{l c c c c c c}
\toprule
\multicolumn{1}{c}{\multirow{2}{*}{Dataset}} & \multicolumn{1}{c}{Baseline} & \multicolumn{1}{c}{Stage 1} & \multicolumn{1}{c}{Stage 2} & \multicolumn{1}{c}{Stage 3} & \multicolumn{1}{c}{Stage 4} & \multicolumn{1}{c}{Stage 5} \\
& \multicolumn{1}{c}{--} & \multicolumn{1}{c}{CUB200} & \multicolumn{1}{c}{PixmoCount} & \multicolumn{1}{c}{PathVQA} & \multicolumn{1}{c}{TextVQA} & \multicolumn{1}{c}{TimeClock} \\
\midrule
\rowcolor{Gray}\multicolumn{7}{l}{\textbf{Target}}\\
CUB200 & 32.6 & 78.4 & 73.9 & 74.3 & 71.5 & 71.9 \\
PixmoCount & 45.7 & 32.4 & 58.4 & 57.9 & 29.8 & 44.8 \\
PathVQA & 13.2 & 27.9 & 27.9 & 59.5 & 51.6 & 54.0 \\
TextVQA & 65.4 & 63.5 & 63.7 & 64.7 & 74.0 & 63.9 \\
TimeClock & 0.8 & 0.8 & 0.6 & 0.7 & 0.8 & 27.6 \\
\midrule
\textbf{Average} & 31.5 & 40.6 & 44.9 & 51.4 & 45.5 & 52.4 \\
\midrule
\rowcolor{Gray}\multicolumn{7}{l}{\textbf{Held out}}\\
AI2D & 71.6 & 67.6 & 67.4 & 68.9 & 66.7 & 66.1 \\
ChartQA & 69.2 & 60.6 & 63.3 & 65.3 & 64.0 & 46.6 \\
DocVQA & 72.7 & 59.7 & 62.4 & 63.8 & 64.1 & 47.6 \\
InfoVQA & 31.9 & 32.3 & 33.0 & 36.2 & 36.5 & 27.5 \\
MMStar & 42.0 & 45.8 & 45.5 & 46.0 & 43.6 & 40.9 \\
RealWorldQA & 59.7 & 60.4 & 52.4 & 54.9 & 59.0 & 49.2 \\
ScienceQA & 73.2 & 72.3 & 71.7 & 73.1 & 71.8 & 71.9 \\
SeedBench & 58.5 & 60.1 & 59.9 & 59.0 & 60.0 & 59.2 \\
\midrule
\textbf{Average} & 59.8 & 57.4 & 56.9 & 58.4 & 58.2 & 51.1 \\
\bottomrule
\end{tabular}
\end{table}

\subsection{Sequential tuning detailed performance tables on Qwen2.5-VL}

We include the detailed task performances for sequential fine-tuning experiments on Qwen2.5-VL here.
Tab.~\ref{tab:llama3_tune_parts_full_model_task_stage_performance_extracted}, Tab.~\ref{tab:llama3_tune_parts_vision_and_projector_task_stage_performance_extracted}, Tab.~\ref{tab:llama3_tune_parts_llm_task_stage_performance_extracted},
Tab.~\ref{tab:LLaVA-LLaMA_3_SA_Proj_task_stage_performance_extracted},
Tab.~\ref{tab:LLaVA-LLaMA_3_MLP_task_stage_performance_extracted}, and
Tab.~\ref{tab:llama3_tune_parts_mlp_gate_and_up_task_stage_performance_extracted} are detailed performance tables of sequentially fine-tuning the Full model, Vision Encoder + Projector, LLM, SA projection layers in LLM, MLP layers in LLM, and MLP (Gate\&Up), respectively.

\begin{table}[t]
\caption{Detailed performance of sequentially fine-tuning the \textbf{full model} of \textbf{Qwen2.5-VL} on each target task.}
\label{tab:qwen2_5_vl_tune_parts_full_task_stage_performance_extracted}
\centering
\begin{tabular}{l c c c c c c}
\toprule
\multicolumn{1}{c}{\multirow{2}{*}{Dataset}} & \multicolumn{1}{c}{Baseline} & \multicolumn{1}{c}{Stage 1} & \multicolumn{1}{c}{Stage 2} & \multicolumn{1}{c}{Stage 3} & \multicolumn{1}{c}{Stage 4} & \multicolumn{1}{c}{Stage 5} \\
& \multicolumn{1}{c}{--} & \multicolumn{1}{c}{CUB200} & \multicolumn{1}{c}{PixmoCount} & \multicolumn{1}{c}{PathVQA} & \multicolumn{1}{c}{TextVQA} & \multicolumn{1}{c}{TimeClock} \\
\midrule
\rowcolor{Gray}\multicolumn{7}{l}{\textbf{Target}}\\
CUB200 & 81.4 & 93.5 & 0.2 & 12.1 & 92.6 & 92.4 \\
PixmoCount & 58.6 & 55.6 & 50.6 & 48.3 & 47.4 & 51.1 \\
PathVQA & 29.2 & 18.5 & 0.0 & 60.8 & 58.2 & 59.1 \\
TextVQA & 83.0 & 69.5 & 17.2 & 73.3 & 81.5 & 62.7 \\
TimeClock & 8.2 & 0.1 & 0.0 & 0.0 & 6.3 & 60.8 \\
\midrule
\textbf{Average} & 52.1 & 47.4 & 13.6 & 38.9 & 57.2 & 65.2 \\
\midrule
\rowcolor{Gray}\multicolumn{7}{l}{\textbf{Held out}}\\
AI2D & 82.9 & 79.5 & 0.1 & 64.0 & 78.8 & 72.5 \\
ChartQA & 83.2 & 72.6 & 54.2 & 72.1 & 69.3 & 62.7 \\
DocVQA & 94.4 & 77.2 & 30.2 & 76.1 & 90.0 & 66.8 \\
InfoVQA & 80.3 & 61.9 & 33.6 & 64.8 & 74.5 & 47.1 \\
MMStar & 62.6 & 59.3 & 0.0 & 34.0 & 53.5 & 46.9 \\
RealWorldQA & 68.6 & 59.5 & 3.7 & 27.5 & 59.7 & 51.5 \\
ScienceQA & 76.7 & 77.8 & 0.4 & 43.4 & 77.3 & 71.6 \\
SeedBench & 74.1 & 72.0 & 0.0 & 24.6 & 68.9 & 63.7 \\
\midrule
\textbf{Average} & 77.9 & 70.0 & 15.3 & 50.8 & 71.5 & 60.4 \\
\bottomrule
\end{tabular}
\end{table}

\begin{table}[t]
\caption{Detailed performance of sequentially fine-tuning the \textbf{vision encoder and projector} of \textbf{Qwen2.5-VL} on each target task.}
\label{tab:qwen2_5_vl_tune_parts_vision_projector}
\centering
\begin{tabular}{l c c c c c c}
\toprule
\multicolumn{1}{c}{\multirow{2}{*}{Dataset}} & \multicolumn{1}{c}{Baseline} & \multicolumn{1}{c}{Stage 1} & \multicolumn{1}{c}{Stage 2} & \multicolumn{1}{c}{Stage 3} & \multicolumn{1}{c}{Stage 4} & \multicolumn{1}{c}{Stage 5} \\
& \multicolumn{1}{c}{--} & \multicolumn{1}{c}{CUB200} & \multicolumn{1}{c}{PixmoCount} & \multicolumn{1}{c}{PathVQA} & \multicolumn{1}{c}{TextVQA} & \multicolumn{1}{c}{TimeClock} \\
\midrule
\rowcolor{Gray}\multicolumn{7}{l}{\textbf{Target}}\\
CUB200 & 81.4 & 92.3 & 81.4 & 81.5 & 81.5 & 88.0 \\
PixmoCount & 58.6 & 56.4 & 59.0 & 59.2 & 58.2 & 33.1 \\
PathVQA & 29.2 & 30.3 & 29.1 & 29.3 & 30.2 & 35.4 \\
TextVQA & 83.0 & 82.5 & 83.2 & 83.1 & 83.1 & 71.0 \\
TimeClock & 8.2 & 8.5 & 8.4 & 8.6 & 8.8 & 57.5 \\
\midrule
\textbf{Average} & 52.1 & 54.0 & 52.2 & 52.3 & 52.4 & 57.0 \\
\midrule
\rowcolor{Gray}\multicolumn{7}{l}{\textbf{Held out}}\\
AI2D & 82.9 & 83.0 & 82.8 & 82.9 & 83.1 & 75.4 \\
ChartQA & 83.2 & 83.8 & 83.7 & 83.8 & 83.9 & 75.1 \\
DocVQA & 94.4 & 94.4 & 94.5 & 94.4 & 94.5 & 88.7 \\
InfoVQA & 80.3 & 79.5 & 80.1 & 80.2 & 80.3 & 69.2 \\
MMStar & 62.6 & 62.3 & 62.5 & 62.9 & 63.4 & 52.7 \\
RealWorldQA & 68.6 & 67.6 & 68.5 & 68.5 & 69.9 & 62.0 \\
ScienceQA & 76.7 & 76.6 & 76.4 & 76.1 & 76.2 & 82.3 \\
SeedBench & 74.1 & 73.7 & 74.0 & 74.1 & 74.1 & 67.8 \\
\midrule
\textbf{Average} & 77.9 & 77.6 & 77.8 & 77.9 & 78.2 & 71.6 \\
\bottomrule
\end{tabular}
\end{table}

\begin{table}[t]
\caption{Detailed performance of sequentially fine-tuning the \textbf{LLM} of \textbf{Qwen2.5-VL} on each target task.}
\label{tab:qwen2_5_vl_tune_parts_llm}
\centering
\begin{tabular}{l c c c c c c}
\toprule
\multicolumn{1}{c}{\multirow{2}{*}{Dataset}} & \multicolumn{1}{c}{Baseline} & \multicolumn{1}{c}{Stage 1} & \multicolumn{1}{c}{Stage 2} & \multicolumn{1}{c}{Stage 3} & \multicolumn{1}{c}{Stage 4} & \multicolumn{1}{c}{Stage 5} \\
& \multicolumn{1}{c}{--} & \multicolumn{1}{c}{CUB200} & \multicolumn{1}{c}{PixmoCount} & \multicolumn{1}{c}{PathVQA} & \multicolumn{1}{c}{TextVQA} & \multicolumn{1}{c}{TimeClock} \\
\midrule
\rowcolor{Gray}\multicolumn{7}{l}{\textbf{Target}}\\
CUB200 & 81.4 & 93.8 & 0.0 & 64.4 & 67.6 & 91.5 \\
PixmoCount & 58.6 & 55.8 & 47.0 & 50.6 & 41.0 & 49.1 \\
PathVQA & 29.2 & 4.9 & 0.0 & 63.0 & 59.7 & 60.2 \\
TextVQA & 83.0 & 47.9 & 11.6 & 73.3 & 82.1 & 61.8 \\
TimeClock & 8.2 & 0.0 & 0.0 & 0.0 & 4.6 & 58.5 \\
\midrule
\textbf{Average} & 52.1 & 40.5 & 11.7 & 50.3 & 51.0 & 64.2 \\
\midrule
\rowcolor{Gray}\multicolumn{7}{l}{\textbf{Held out}}\\
AI2D & 82.9 & 77.4 & 0.0 & 35.8 & 75.6 & 56.9 \\
ChartQA & 83.2 & 41.9 & 49.1 & 67.0 & 65.1 & 68.2 \\
DocVQA & 94.4 & 49.6 & 23.4 & 76.3 & 89.9 & 65.8 \\
InfoVQA & 80.3 & 41.6 & 28.5 & 60.8 & 74.5 & 49.3 \\
MMStar & 62.6 & 59.7 & 0.0 & 33.9 & 52.4 & 36.1 \\
RealWorldQA & 68.6 & 56.7 & 3.4 & 25.4 & 51.8 & 38.3 \\
ScienceQA & 76.7 & 77.6 & 0.0 & 39.5 & 69.9 & 59.3 \\
SeedBench & 74.1 & 71.7 & 0.0 & 21.9 & 61.8 & 51.8 \\
\midrule
\textbf{Average} & 77.9 & 59.5 & 13.1 & 45.1 & 67.6 & 53.2 \\
\bottomrule
\end{tabular}
\end{table}

\begin{table}[t]
\caption{Detailed performance of sequentially fine-tuning the \textbf{SA projection layers in the LLM} of \textbf{Qwen2.5-VL} on each target task.}
\label{tab:Qwen2.5-VL_SA_Proj_task_stage_performance_extracted}
\centering
\begin{tabular}{l c c c c c c}
\toprule
\multicolumn{1}{c}{\multirow{2}{*}{Dataset}} & \multicolumn{1}{c}{Baseline} & \multicolumn{1}{c}{Stage 1} & \multicolumn{1}{c}{Stage 2} & \multicolumn{1}{c}{Stage 3} & \multicolumn{1}{c}{Stage 4} & \multicolumn{1}{c}{Stage 5} \\
& \multicolumn{1}{c}{--} & \multicolumn{1}{c}{CUB200} & \multicolumn{1}{c}{PixmoCount} & \multicolumn{1}{c}{PathVQA} & \multicolumn{1}{c}{TextVQA} & \multicolumn{1}{c}{TimeClock} \\
\midrule
\rowcolor{Gray}\multicolumn{7}{l}{\textbf{Target}}\\
CUB200 & 81.4 & 93.7 & 93.6 & 93.4 & 93.4 & 93.2 \\
PixmoCount & 58.6 & 59.9 & 53.4 & 56.7 & 53.6 & 53.4 \\
PathVQA & 29.2 & 35.7 & 35.3 & 61.0 & 57.3 & 58.3 \\
TextVQA & 83.0 & 77.8 & 77.4 & 81.8 & 83.6 & 80.6 \\
TimeClock & 8.2 & 10.5 & 9.1 & 9.6 & 9.9 & 49.4 \\
\midrule
\textbf{Average} & 52.1 & 55.5 & 53.8 & 60.5 & 59.6 & 67.0 \\
\midrule
\rowcolor{Gray}\multicolumn{7}{l}{\textbf{Held out}}\\
AI2D & 82.9 & 83.3 & 82.8 & 83.3 & 82.4 & 82.2 \\
ChartQA & 83.2 & 84.4 & 79.9 & 86.8 & 86.2 & 84.5 \\
DocVQA & 94.4 & 85.6 & 92.6 & 92.9 & 93.8 & 92.7 \\
InfoVQA & 80.3 & 74.9 & 77.3 & 79.4 & 78.9 & 78.6 \\
MMStar & 62.6 & 63.6 & 63.7 & 62.4 & 61.4 & 61.8 \\
RealWorldQA & 68.6 & 70.1 & 68.1 & 67.8 & 69.2 & 68.4 \\
ScienceQA & 76.7 & 85.9 & 84.9 & 86.8 & 86.5 & 85.9 \\
SeedBench & 74.1 & 73.9 & 74.0 & 73.5 & 73.3 & 73.5 \\
\midrule
\textbf{Average} & 77.9 & 77.7 & 77.9 & 79.1 & 79.0 & 78.5 \\
\bottomrule
\end{tabular}
\end{table}

\begin{table}[t]
\caption{Detailed performance of sequentially fine-tuning the \textbf{MLP in the LLM} of \textbf{Qwen2.5-VL} on each target task.}
\label{tab:Qwen2.5-VL_MLP_task_stage_performance_extracted}
\centering
\begin{tabular}{l c c c c c c}
\toprule
\multicolumn{1}{c}{\multirow{2}{*}{Dataset}} & \multicolumn{1}{c}{Baseline} & \multicolumn{1}{c}{Stage 1} & \multicolumn{1}{c}{Stage 2} & \multicolumn{1}{c}{Stage 3} & \multicolumn{1}{c}{Stage 4} & \multicolumn{1}{c}{Stage 5} \\
& \multicolumn{1}{c}{--} & \multicolumn{1}{c}{CUB200} & \multicolumn{1}{c}{PixmoCount} & \multicolumn{1}{c}{PathVQA} & \multicolumn{1}{c}{TextVQA} & \multicolumn{1}{c}{TimeClock} \\
\midrule
\rowcolor{Gray}\multicolumn{7}{l}{\textbf{Target}}\\
CUB200 & 81.4 & 94.1 & 83.5 & 93.1 & 93.0 & 92.6 \\
PixmoCount & 58.6 & 56.6 & 50.0 & 53.4 & 31.8 & 52.2 \\
PathVQA & 29.2 & 32.4 & 2.8 & 61.5 & 60.6 & 60.4 \\
TextVQA & 83.0 & 76.7 & 8.2 & 78.9 & 83.4 & 64.4 \\
TimeClock & 8.2 & 7.5 & 4.8 & 6.0 & 3.3 & 60.2 \\
\midrule
\textbf{Average} & 52.1 & 53.5 & 29.9 & 58.6 & 54.4 & 66.0 \\
\midrule
\rowcolor{Gray}\multicolumn{7}{l}{\textbf{Held out}}\\
AI2D & 82.9 & 78.4 & 0.1 & 81.0 & 81.0 & 74.0 \\
ChartQA & 83.2 & 83.3 & 0.0 & 74.2 & 82.2 & 74.8 \\
DocVQA & 94.4 & 82.0 & 3.0 & 89.3 & 92.5 & 72.7 \\
InfoVQA & 80.3 & 71.4 & 2.4 & 75.4 & 78.3 & 59.4 \\
MMStar & 62.6 & 60.6 & 29.0 & 58.8 & 61.4 & 55.9 \\
RealWorldQA & 68.6 & 66.1 & 4.4 & 61.4 & 65.9 & 50.7 \\
ScienceQA & 76.7 & 79.6 & 0.1 & 80.3 & 82.2 & 77.0 \\
SeedBench & 74.1 & 73.2 & 20.5 & 72.0 & 72.6 & 70.9 \\
\midrule
\textbf{Average} & 77.9 & 74.3 & 7.4 & 74.0 & 77.0 & 66.9 \\
\bottomrule
\end{tabular}
\end{table}

\begin{table}[t]
\caption{Detailed performance of sequentially fine-tuning the \textbf{MLP (Gate \& Up) in the LLM} of \textbf{Qwen2.5-VL} on each target task.}
\label{tab:qwen2_5_vl_tune_parts_mlp_gate_up}
\centering
\begin{tabular}{l c c c c c c}
\toprule
\multicolumn{1}{c}{\multirow{2}{*}{Dataset}} & \multicolumn{1}{c}{Baseline} & \multicolumn{1}{c}{Stage 1} & \multicolumn{1}{c}{Stage 2} & \multicolumn{1}{c}{Stage 3} & \multicolumn{1}{c}{Stage 4} & \multicolumn{1}{c}{Stage 5} \\
& \multicolumn{1}{c}{--} & \multicolumn{1}{c}{CUB200} & \multicolumn{1}{c}{PixmoCount} & \multicolumn{1}{c}{PathVQA} & \multicolumn{1}{c}{TextVQA} & \multicolumn{1}{c}{TimeClock} \\
\midrule
\rowcolor{Gray}\multicolumn{7}{l}{\textbf{Target}}\\
CUB200 & 81.4 & 94.1 & 94.2 & 93.8 & 94.0 & 94.0 \\
PixmoCount & 58.6 & 59.7 & 49.6 & 50.0 & 50.9 & 54.7 \\
PathVQA & 29.2 & 36.0 & 22.0 & 61.5 & 61.7 & 62.3 \\
TextVQA & 83.0 & 75.9 & 74.7 & 81.1 & 83.8 & 79.6 \\
TimeClock & 8.2 & 8.1 & 6.5 & 7.6 & 5.1 & 55.5 \\
\midrule
\textbf{Average} & 52.1 & 54.8 & 49.4 & 58.8 & 59.1 & 69.2 \\
\midrule
\rowcolor{Gray}\multicolumn{7}{l}{\textbf{Held out}}\\
AI2D & 82.9 & 82.3 & 76.6 & 69.0 & 80.7 & 75.5 \\
ChartQA & 83.2 & 81.8 & 81.9 & 84.1 & 76.6 & 80.5 \\
DocVQA & 94.4 & 81.9 & 81.2 & 91.3 & 93.0 & 90.4 \\
InfoVQA & 80.3 & 72.3 & 68.3 & 78.4 & 79.5 & 78.4 \\
MMStar & 62.6 & 62.5 & 47.6 & 56.0 & 59.1 & 58.7 \\
RealWorldQA & 68.6 & 68.8 & 34.5 & 56.3 & 64.4 & 66.3 \\
ScienceQA & 76.7 & 83.4 & 78.7 & 56.5 & 72.6 & 65.3 \\
SeedBench & 74.1 & 73.8 & 60.1 & 68.2 & 71.7 & 71.1 \\
\midrule
\textbf{Average} & 77.9 & 75.8 & 66.1 & 70.0 & 74.7 & 73.3 \\
\bottomrule
\end{tabular}
\end{table}


\clearpage

\end{document}